\newcommand{\CLASSINPUTinnersidemargin}{10mm}
\documentclass[10pt,journal,compsoc]{IEEEtran}

\ifCLASSOPTIONcompsoc
  \usepackage[nocompress]{cite}
\else
  \usepackage{cite}
\fi

\usepackage[pdftex]{graphicx}
\graphicspath{{./figs/}}
\DeclareGraphicsExtensions{.pdf,.jpg,.png}
\usepackage{cite}
\usepackage[cmex10]{amsmath}
\usepackage{amssymb}
\usepackage{graphicx}
\usepackage{cases}
\usepackage{mdwmath}
\usepackage{mdwtab}
\usepackage{array}
\usepackage{url}
\usepackage[ruled,vlined]{algorithm2e}
\usepackage{booktabs}
\usepackage{threeparttable}
\usepackage{color}
\usepackage{caption}
\usepackage{pifont}
\usepackage{wasysym}
\usepackage{amssymb}
\usepackage{multirow}
\usepackage{makecell}
\usepackage[colorlinks,linkcolor=black]{hyperref}    
\usepackage{marvosym}
\usepackage{lipsum}
\usepackage[caption=false,font=footnotesize]{subfig}
\usepackage{amsmath}
\usepackage{xspace}
\usepackage{bm}
\usepackage{amsthm,amsmath,amssymb}
\usepackage{mathrsfs}
\usepackage{mathrsfs}
\usepackage{multicol} 
\usepackage{hyperref}
\hypersetup{
    colorlinks=true,             
    linkcolor=black,             
    filecolor=black,             
    urlcolor=black,              
    citecolor=black,             
    pdftitle={Overleaf Example},
    pdfpagemode=FullScreen,
}

\newcommand{\myPara}[1]{\vspace{0.02in}\noindent\textbf{#1}}
\newcommand{\tabincell}[2]{\begin{tabular}{@{}#1@{}}#2\end{tabular}}
\newcommand{\cmark}{\ding{51}}%
\newcommand{\xmark}{\ding{55}}%

\makeatletter
\newcommand{\rmnum}[1]{\romannumeral #1}
\newcommand{\Rmnum}[1]{\expandafter\@slowromancap\romannumeral #1@}
\makeatother
\usepackage{graphicx}
\usepackage{float}
\usepackage{diagbox}

\newcommand{\ie}{\textit{i}.\textit{e}., }
\newcommand{\eg}{\textit{e}.\textit{g}., }
\def\etal{{\em et al.}}
\newcommand{\etc}{\textit{e}\textit{t}\textit{c}}

\ifCLASSINFOpdf
\else
\fi
\begin{document}

\title{WebUAV-3M: A Benchmark for Unveiling the Power of Million-Scale Deep UAV Tracking}

\author{Chunhui~Zhang$^*$,
        Guanjie~Huang$^*$,
        Li~Liu$^{\dagger}$,\IEEEmembership{~Member,~IEEE},
        Shan~Huang,
        Yinan~Yang,
        Xiang~Wan,
        Shiming~Ge,\IEEEmembership{~Senior~Member,~IEEE},
        and~Dacheng~Tao,\IEEEmembership{~Fellow,~IEEE}

\IEEEcompsocitemizethanks{\IEEEcompsocthanksitem Chunhui Zhang is with the Cooperative Medianet Innovation Center, Shanghai Jiao Tong University, Shanghai 200240, China and also with the CloudWalk Technology Co., Ltd, 201203, China. Email: chunhui.zhang@sjtu.edu.cn.\protect

\IEEEcompsocthanksitem Guanjie Huang is with the Shenzhen Research Institute of Big Data, the Chinese University of Hong Kong, Shenzhen 518172, China. E-mail: rasel.laffel@live.com.\protect

\IEEEcompsocthanksitem 
Li Liu is with the Shenzhen Research Institute of Big Data, the Chinese University of Hong Kong, Shenzhen 518172, China. E-mail: liliu.math@gmail.com.\protect


\IEEEcompsocthanksitem Shan Huang and Yinan Yang are with the Shenzhen Research Institute of Big Data, the Chinese University of Hong Kong, Shenzhen 518172, China.\protect

\IEEEcompsocthanksitem Xiang Wan is with the Shenzhen Research Institute of Big Data, the Chinese University of Hong Kong, Shenzhen 518172, China. Email: wanxiang@sribd.cn.\protect

\IEEEcompsocthanksitem Shiming Ge is with the Institute of Information Engineering, Chinese Academy of Sciences, Beijing 100093, China. Email: geshiming@iie.ac.cn.\protect


\IEEEcompsocthanksitem Dacheng Tao is with the JD Explore Academy in JD.com, China and the University of Sydney, Australia. Email: dacheng.tao@gmail.com.\protect

}

\thanks{$^{*}$The first two authors contributed equally. This work was done during their internships at the SRIBD, the CUHK, Shenzhen.}
\thanks{$^{\dagger}$Corresponding author.}
}

%
%


\markboth{IEEE Transactions on Pattern Analysis and Machine Intelligence,~Vol.~X, No.~X, December~2022}%
{Chunhui Zhang \MakeLowercase{\textit{et al.}}: Bare Advanced Demo of IEEEtran.cls for IEEE Computer Society Journals}

%



\IEEEtitleabstractindextext{
\begin{abstract} 
Unmanned aerial vehicle (UAV) tracking is of great significance for a wide range of applications, such as delivery and agriculture. Previous benchmarks in this area mainly focused on small-scale tracking problems while ignoring the amounts of data, types of data modalities, diversities of target categories and scenarios, and evaluation protocols involved, greatly hiding the massive power of deep UAV tracking. In this work, we propose WebUAV-3M, the largest public UAV tracking benchmark to date, to facilitate both the development and evaluation of deep UAV trackers. WebUAV-3M contains over 3.3 million frames across 4,500 videos and offers 223 highly diverse target categories. Each video is densely annotated with bounding boxes by an efficient and scalable semi-automatic target annotation (SATA) pipeline. Importantly, to take advantage of the complementary superiority of language and audio, we enrich WebUAV-3M by innovatively providing both natural language specifications and audio descriptions. We believe that such additions will greatly boost future research in terms of exploring language features and audio cues for multi-modal UAV tracking. In addition, a fine-grained UAV tracking-under-scenario constraint (UTUSC) evaluation protocol and seven challenging scenario subtest sets are constructed to enable the community to develop, adapt and evaluate various types of advanced trackers. We provide extensive evaluations and detailed analyses of 43 representative trackers and envision future research directions in the field of deep UAV tracking and beyond. The dataset, toolkits, and baseline results are available at \textcolor{black}{\url{https://github.com/983632847/WebUAV-3M}.}

\end{abstract}

}

\maketitle

\IEEEdisplaynontitleabstractindextext

%
\IEEEpeerreviewmaketitle

\ifCLASSOPTIONcompsoc
\IEEEraisesectionheading{\section{Introduction}\label{sec:introduction}}
\else
\section{Introduction}
\label{sec:introduction}
\fi

%
%
%
%

\IEEEPARstart{U}{nmanned} aerial vehicle (UAV) tracking refers to the task of sequentially locating a moving target in a video captured from a low-altitude UAV without accessing prior knowledge about the target (\eg the target class and motion pattern) or its surrounding environment~\cite{mueller2016benchmark,zhang2020accurate,zhu2020vision,li2017visual}. This is one of the fundamental yet open problems in computer vision, and it has attracted increasing attention due to a wide range of real applications, such as transportation surveillance, aerial photography, marine search and rescue, delivery, and intelligent agriculture~\cite{li2021all,du2018unmanned}. Consequently, the automatic understanding of the visual data collected from UAVs has become highly demanding, and numerous algorithms have been proposed~\cite{zhu2020vision}. During this process, UAV tracking benchmarks have played a vital role in objectively evaluating and comparing different trackers. However, the further development and assessment of deep UAV tracking approaches are seriously limited by the following issues of existing benchmarks and datasets.

\begin{figure}[t]   
	\centering\centerline{\includegraphics[width=1.0\linewidth]{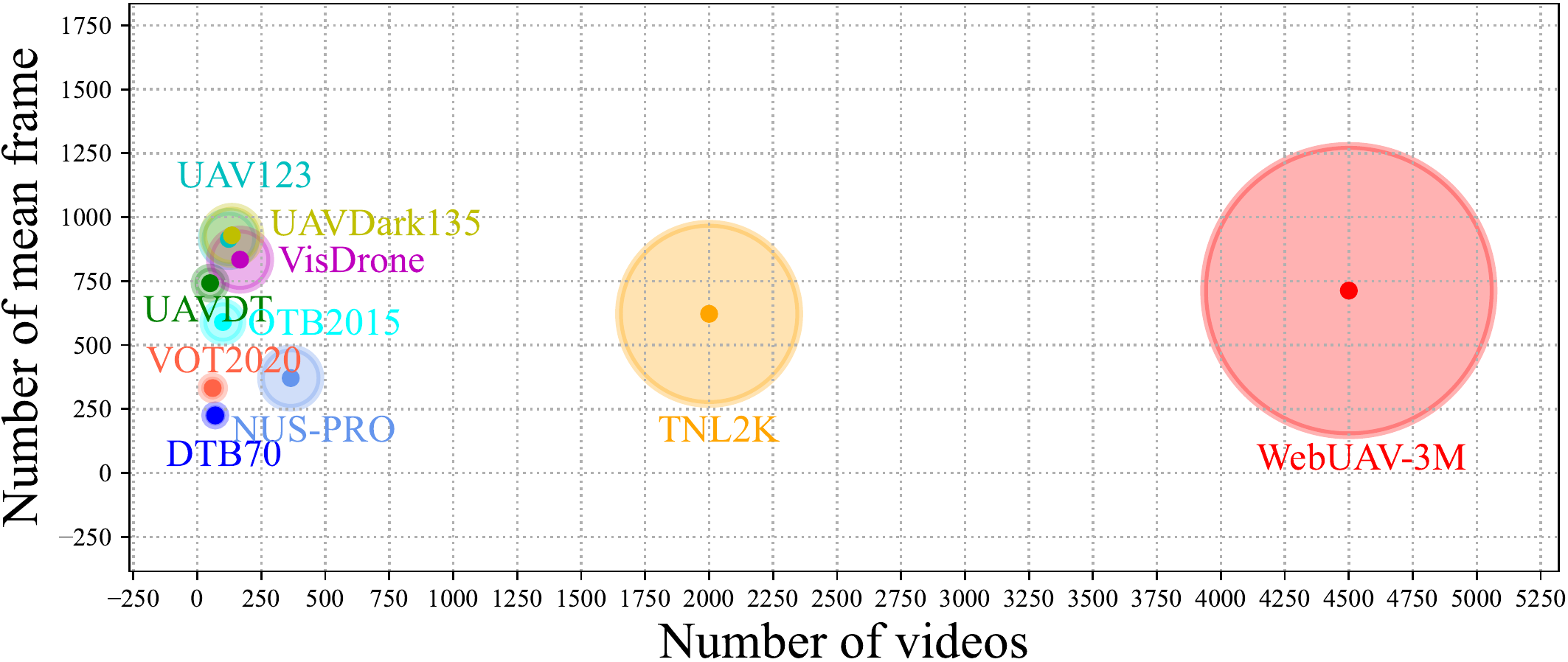}}
	\caption{Comparison of our WebUAV-3M with the public benchmarks, including UAV tracking benchmarks, \ie UAV123, DTB70, UAVDT, VisDrone, UAVDark135, and GOT benchmarks, \ie OTB2015, VOT2020, NUS-PRO, and TNL2K. The area of each circle is proportional to the number of total frames of the corresponding benchmark. Best viewed in color.}
	\label{fig:motivation}
\end{figure}

\myPara{(\rmnum{1}) Lack of large-scale benchmarks.} Deep UAV tracking models are challenging to train in cases involving the scarcity of labeled data due to their inherent data-hungry characteristics. Some recent efforts~\cite{DanelljanBKF19,HuangZH2020aaai,LiWWZXY19} have been devoted to pretraining models on large-scale general object tracking (GOT)\footnote{General object tracking aims to estimate the state of an arbitrary object in videos captured by common ground cameras when given only the location (\eg bounding box) in the first frame.} datasets~\cite{huang2019got,fan2019lasot} to achieve improved deep UAV tracking performance. Compared with videos collected by ground cameras, the video sequences captured by UAVs generally present more diverse yet unique viewpoints, more obvious motion blurs, and more varying target resolutions due to the rapid movements and continuously changing attitudes/heights of the UAVs during flight~\cite{zhang2020accurate,tianjiao2021cvpr}. These factors inevitably lead to suboptimal deep UAV tracking performance. In contrast, some works have continued to construct datasets captured by UAVs focusing on object detection or tracking~\cite{du2018unmanned,zhu2020vision,li2017visual}. However, these datasets are still limited in terms of their sizes and scenarios covered due to severe challenges in data collection and annotation~\cite{zhu2020vision}. As shown in Fig.~\ref{fig:motivation}, public UAV tracking benchmarks seldom have more than 500 video sequences. Thorough evaluations of existing or newly developed UAV tracking algorithms remain an open problem due to these small-scale benchmarks. A more general and comprehensive benchmark is desired to further boost video analysis research on deep UAV tracking. Furthermore, efficient labeling tools are necessary for large-scale and high-quality densely labeled benchmarks.

\begin{figure*}[t]   
	\centering\centerline{\includegraphics[width=1.0\linewidth]{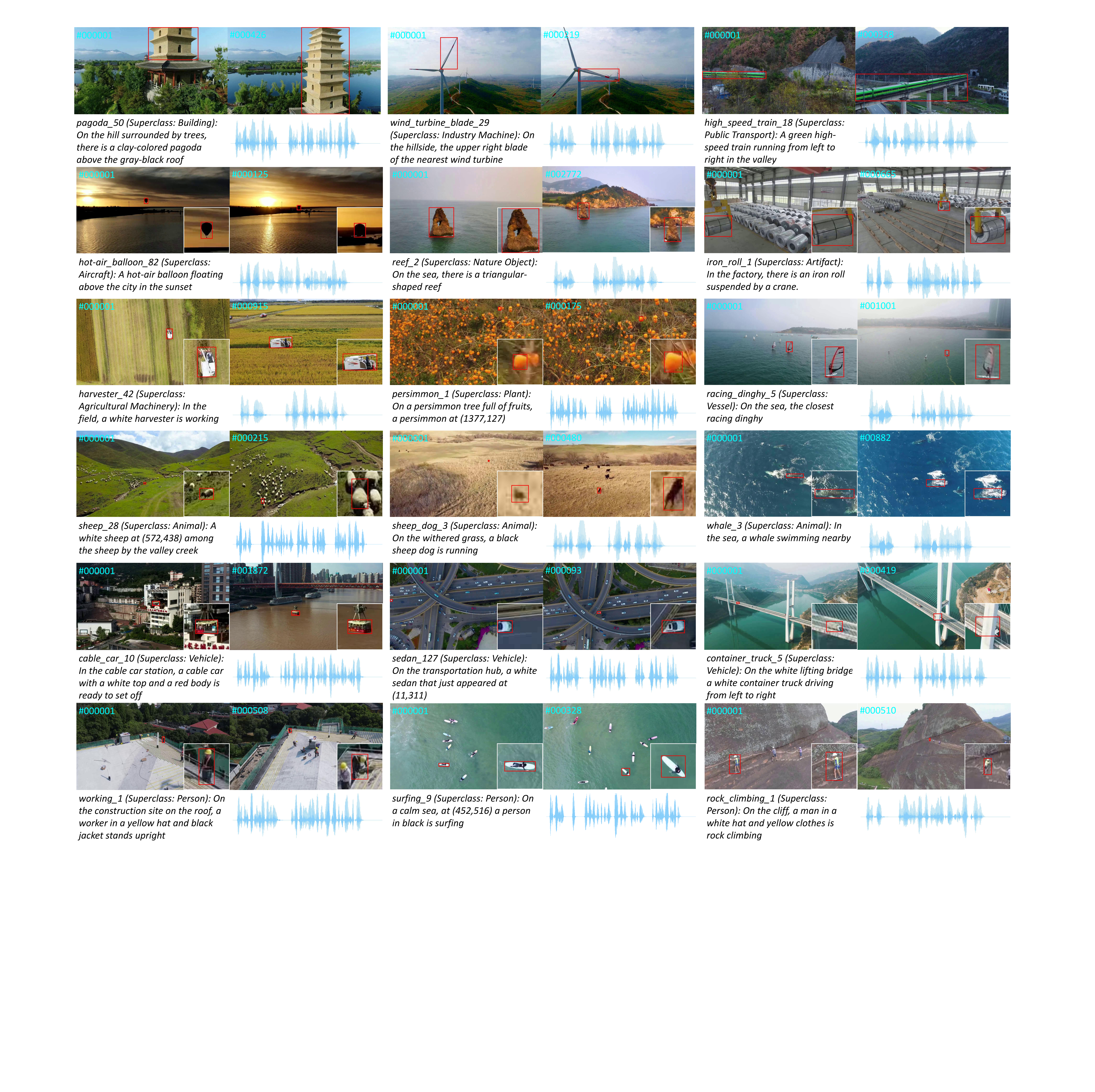}}
	\caption{A glance at some diverse video sequences and annotations in WebUAV-3M. All sequences are divided into 12 superclasses, including \emph{person}, \emph{building}, \emph{vehicle}, \emph{vessel}, \emph{public transport}, \emph{aircraft}, \emph{animal}, \emph{agricultural machinery}, \emph{industry machine}, \emph{plant}, \emph{artifact} and \emph{natural object}. Each sequence is attached to a semantic label: ``target class'' or ``motion class''. WebUAV-3M populates 223 target classes and 63 motion classes in total. In addition, we provide a natural language specification and two audio descriptions for each sequence. Best viewed in color.}
	\label{fig:Example_sequences}
\end{figure*}

\myPara{(\rmnum{2}) Lack of language and audio descriptions.} Natural language specifications and audio descriptions have recently proven to be beneficial for various vision tasks (\eg \cite{hu2016natural,zhao2019sound}) including object tracking~\cite{wang2021towards,feng2019robust,gan2019self}. First, compared with 2D visual bounding boxes, natural language specifications/audio descriptions (see Fig.~\ref{fig:Example_sequences}) can provide important auxiliary information. Serving as high-level semantic guidance, text information helps to alleviate the ambiguity issue of bounding boxes and the vast appearance variations exhibited by the target object~\cite{wang2021towards,gan2019self}. UAV tracking algorithms based on visual appearance may perform poorly when facing abrupt appearance changes, fast motion, or long-term occlusion. The appearance features initialized in the first frame are very different from those of the target during the tracking process. Second, initializing target objects with natural language specifications/audio descriptions is convenient and valuable in many real-world tracking applications (\eg human–machine interaction for the blind). Some recent works have been developed with natural language specifications to address model drift or simultaneous multiple-video tracking~\cite{wang2021towards,li2017tracking}, and with stereo sound and camera metadata to recover the coordinates of moving vehicles in reference frames~\cite{gan2019self}. However, to our best knowledge, natural language specification/audio description-based UAV tracking algorithms have seldom been explored in recent UAV tracking benchmarks~\cite{mueller2016benchmark,zhu2020vision,li2017visual,du2018unmanned}, which motivates the proposal of a new and large-scale benchmark for this task.

\myPara{(\rmnum{3}) Limited target classes.} Both UAV tracking and GOT are class-agnostic tasks~\cite{huang2019got}, which means that the tracker can be used for a wide range of target categories and should have robust tracking performance even for unseen target categories. However, most existing UAV tracking benchmarks consist of only limited categories relative to several recent GOT benchmarks. It is challenging to train robust deep UAV tracking algorithms for real-world application scenarios. For instance, one of the latest UAV tracking datasets, VisDrone~\cite{zhu2020vision}, includes only ten categories. Indeed, it covers a wide range of aspects, including locations (taken from 14 different cities separated by thousands of kilometers in China), environments (\ie urban and country), and densities (\ie sparse and crowded scenes). However, the small number of target categories inevitably limits deep UAV trackers' performance for a wide range of unseen categories~\cite{huang2019got}.

\myPara{(\rmnum{4}) Lack of a rigorous evaluation protocol.} 
Evaluation protocols and test sets play essential roles in analyzing deep UAV tracking systems~\cite{kristan2016novel}. Ideally, a robust evaluation protocol should encourage the given tracking system to have stable performance that is insensitive to various scenarios. This signifies that the developed tracker possesses generalization ability and can effectively handle special, challenging scenarios. However, most popular UAV tracking benchmarks, including UAV123~\cite{mueller2016benchmark}, DTB70~\cite{li2017visual}, UAVDT~\cite{du2018unmanned}, and VisDrone~\cite{zhu2020vision}, only provide global attribute-based evaluations in which the attributes occur anywhere in the sequences, and binary attribute annotations, which mainly target the overall tracking quality, are provided. For instance, UAVDT~\cite{du2018unmanned} conducts attribute-based evaluations in terms of eight global tracking attributes (\eg background clutter, object rotation, object blur, and scale variation). Unfortunately, global attribute-based evaluations may yield ambiguous results and cannot characterize the strengths and weaknesses of different trackers. Thus, fine-grained labeling is required to facilitate a more rigorous evaluation. Moreover, several stable test sets are desired to obtain more reliable evaluation results.

In the literature, many state-of-the-art datasets have been constructed to address the four issues mentioned above, \eg \cite{zhu2020vision,fan2019lasot,huang2019got} for large-scale benchmarks, ~\cite{wang2021towards} for tracking via natural language specification, \cite{huang2019got} for diverse object classes, and \cite{kristan2016novel,valmadre2018long} for rigorous evaluation protocols. Nevertheless, none of these datasets tackle all the issues, hiding the massive power of deep UAV tracking.

To this end, this work presents a sizeable unified UAV tracking benchmark, which solves the above problems by using a semi-automatic target annotation (SATA) tool to enable the tremendous WebUAV-3M dataset to be densely labeled (some representative videos are shown in Fig.~\ref{fig:Example_sequences}), and proposing a scenario constraint evaluation protocol to achieve fine-grained veritable evaluation. The contributions of this work can be summarized as follows.

\begin{figure*}[t]   
\centering\centerline{\includegraphics[width=1.0\linewidth]{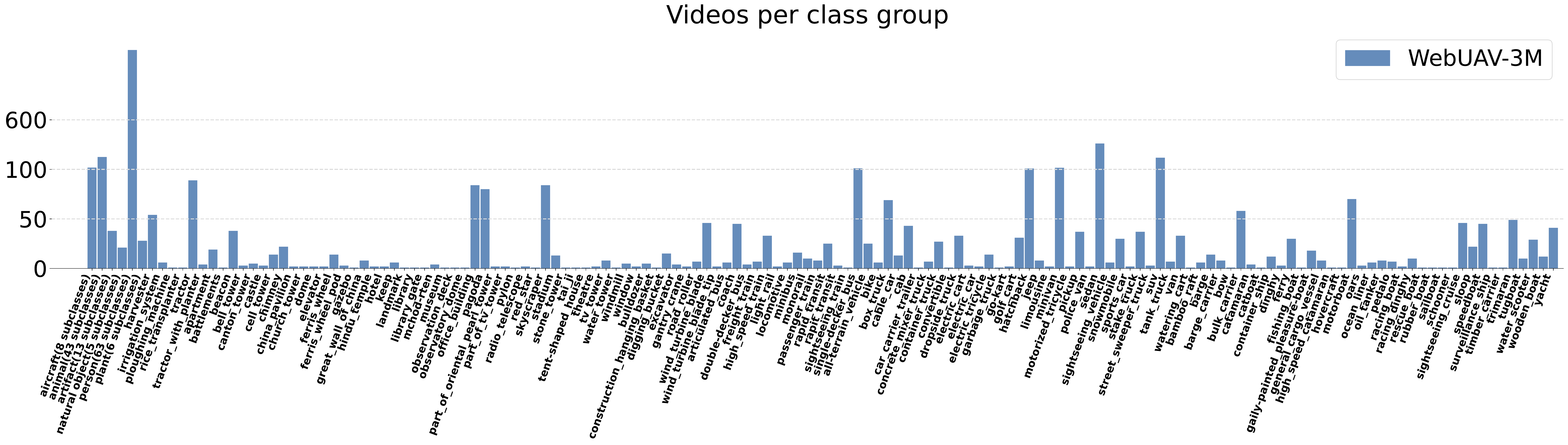}}
	\caption{The number of videos per group of object classes. Best viewed by zooming in.}
	\label{fig:Group_of_object_classes}
\end{figure*}

(1) We construct a new million-scale dataset called WebUAV-3M for deep UAV tracking, which collects 4,500 video sequences and densely annotates approximately 3.3 million high-quality bounding boxes (see Table~\ref{tab:Comp_WebUAV_3M}, Fig.~\ref{fig:Group_of_object_classes}). WebUAV-3M contains 223 target categories, offering coverage of real-world moving objects that is magnitudes wider than those of all existing UAV tracking benchmarks. By releasing WebUAV-3M, we aim to offer a dedicated platform for the unified training and comparison of UAV tracking algorithms.

(2) A general and scalable SATA pipeline, as shown in Fig.~\ref{fig:Pipeline}, is introduced to achieve accurate bounding box annotations and significantly reduce the required human labor. This pipeline makes it possible to label the tremendous WebUAV-3M dataset within three months completely.

(3) The UAV tracking-under-scenario constraint (UTUSC) evaluation protocol, as well as seven subtest sets with fine-grained and challenging scenarios, including low light, long-term occlusion, small targets, high-speed motion, target distortions, dual-dynamic disturbances, and adversarial examples, are constructed to facilitate the evaluation of deep UAV tracking algorithms for real-world applications. The UTUSC is a more rigorous and fine-grained protocol that yields more reliable evaluation quality than global attribute-based evaluation approaches.

(4) WebUAV-3M provides natural language specifications and audio descriptions in addition to 2D visual bounding box annotations, similar to existing UAV tracking datasets. For each video sequence, we label one sentence in English for the whole video and convert it into audio using a text-to-speech software. The aim is to encourage and facilitate explorations that integrate visual, lingual, and audio features for improving multi-modal deep UAV tracking.

(5) Based on the new benchmark, we perform million-scale deep UAV tracking experiments. We evaluate 43 representative trackers and analyze their performance using different evaluation metrics to provide extensive baselines for future comparisons on the WebUAV-3M dataset. The results indicate that substantial room for improvement remains regarding robust deep UAV tracking, as well as the fact that innovation in scenario-constrained deep UAV tracking is necessary.


\begin{table*}[!t]
	{\footnotesize
	\renewcommand\arraystretch{1.0}
	\caption{Comparison of WebUAV-3M with the popular GOT and UAV tracking benchmarks. WebUAV-3M is much larger than other UAV tracking benchmarks. It offers coverage of target classes and motion classes that is magnitudes broader than that of other benchmarks. All the sequences are densely labeled with a SATA method. Furthermore, WebUAV-3M provides natural language specifications and audio descriptions to facilitate the exploration of language features and audio cues for deep UAV tracking. ``n/a'' denotes ``Not Applicable''.}
	\label{tab:Comp_WebUAV_3M}
	\begin{center}
		\setlength{\tabcolsep}{0.65mm}{
			\begin{tabular}{|l||c|c|c|c|c|c|c|c|c|c|c|c|c|c|}
				\hline
				Dataset & Videos &   \tabincell{c}{ Min \\ frame} & \tabincell{c}{Mean\\ frame} & \tabincell{c}{Max\\frame} & \tabincell{c}{Total\\ frames} & \tabincell{c}{Frame \\rate} & \tabincell{c}{Total \\duration}  &  \tabincell{c}{Absent\\ labels} & Classes & Attributes & \tabincell{c}{Data \\partition}  &\tabincell{c}{ Lingual \\feature} & \tabincell{c}{Audio\\ cue}& \tabincell{c}{Annotation\\ method}\\
				\hline
			    \hline	
			    
			    \textbf{OTB2013}~\cite{wu2013online} & 51 & 71 & 578 & 3,872 & 29 K & 30 fps & 16.4 min & \xmark & 10 & 11 & Test & \xmark & \xmark & Manual\\
			    
				\textbf{OTB2015}~\cite{wu2015otb} & 100 & 71 &  590 &  3,872 & 59 K & 30 fps & 32.9 min  & \xmark & 16 & 11 & Test & \xmark & \xmark & Manual  \\ 
				
				\textbf{TC-128}~\cite{liang2015encoding} & 128 & 71 &  429 &  3,872 & 55 K & 30 fps & 30.7 min  & \xmark & 27 & 11 & Test & \xmark & \xmark & Manual  \\
				
				\textbf{VOT2014}~\cite{hadfield2014visual} & 25 & 164 & 409 & 1,210 & 10 K & 30 fps & 5.7 min & \xmark & 11 & n/a & Test & \xmark & \xmark & Manual \\ 
				
				\textbf{VOT2017}~\cite{kristan2017visual} & 60 & 41 & 356 & 1,500 & 21 K  & 30 fps & 11.9 min & \xmark & 24 & n/a & Test & \xmark & \xmark & Manual \\
				
				{\color{black}\textbf{VOT2020}}~\cite{kristan2020eighth} & 60 & 41 & 332 & 1,500 & 20 K  & 30 fps & 11.1 min & \xmark & $-$ & n/a & Test & \xmark & \xmark & Manual\\
				
				\textbf{NUS-PRO}~\cite{li2015nus} & 365 & 146 & 371 & 5,040 & 135 K  & 30 fps & 75.2 min & \xmark & 8 & n/a & Test & \xmark & \xmark & Manual \\
				
				\textbf{NfS}~\cite{kiani2017need} & 100 & 169 & 3,830 & 20,665 & 383 K  & 240 fps & 26.6 min & \xmark & 17 & 9 & Test & \xmark & \xmark & Manual \\
				
				{\color{black}\textbf{OxUvA}}~\cite{valmadre2018long} & 366 & 900 & 4,260 & 37,440 & 1.55 M  & 30 fps & 14 hours & \xmark & 22 & 6 & Test & \xmark & \xmark & Manual\\
				
	            {\color{black}\textbf{TrackingNet}}~\cite{muller2018trackingnet} & 30,643 & $-$ & 480 & $-$ & 14.43 M  & 30 fps & 140 hours & \xmark & 27 & 15 & Train/Test & \xmark & \xmark & Manual\\
				
				\textbf{GOT-10k}~\cite{huang2019got} & 10,000 & 29 & 149 & 1418 & 1.5 M  & 10 fps & $-$ & \cmark & 563 & 6 & Train/Test & \xmark & \xmark & Manual\\
				
				\textbf{LaSOT}~\cite{fan2019lasot} & 1,400 & 1,000 & 2,506 & 11,397 & 3.52 M  & 30 fps & 32.5 hours & \cmark & 70 & 14 & Train/Test & \cmark & \xmark & Manual\\
				
				\textbf{TNL2K}~\cite{wang2021towards} & 2,000 & 21 & 622 & 18,488 & 1.24 M  & 30 fps & $-$ & \cmark & $-$ & 17 & Train/Test & \cmark & \xmark & Manual\\
                \hline
				\hline
				
	        	\textbf{UAV123}~\cite{mueller2016benchmark} & 123 & 109 & 915 & 3,085 & 113 K  & 30 fps & 62.5 min & \xmark & 9 & 12 & Test & \xmark & \xmark & Manual \\
	        	
	        	\textbf{UAV20L}~\cite{mueller2016benchmark} & 20 & 1,717 & 2,934 & 5,527 & 59 K  & 30 fps & 32.6 min & \xmark & 5 & 12 & Test & \xmark & \xmark & Manual \\
	        	
	        	{\color{black}\textbf{DTB70}}~\cite{li2017visual} & 70 & 68 & 225 & 699 & 15.8 K  & 30 fps & 8.8 min & \xmark & $-$ & 11 & Test & \xmark & \xmark & Manual \\
	        	
	        	{\color{black}\textbf{UAVDT}}~\cite{du2018unmanned} & 50 & 82 & 742 & 2,969 & 37.1 K & 30 fps & 20.6 min & \xmark & 3 & 8 & Test & \xmark & \xmark & Semi-automatic\\
	        	
	        	{\color{black}\textbf{UAVDark135}}~\cite{li2021all} & 135 & 216 & 929 & 4,571 & 125.47 K  & 30 fps & 69.7 min & \xmark & 5 & 12 & Test & \xmark & \xmark & Manual \\

	        	{\color{black}\textbf{VisDrone}}~\cite{zhu2020vision} & 167 & 90 & 834 & 4,280 & 139.28 K  & 30 fps & 77.4 min & \xmark & 10 & 12 & Train/Val/Test & \xmark & \xmark & Manual \\	 
				\hline
				\hline
				
				{\color{black}\textbf{WebUAV-3M}} & 4,500 & 40 & 710 & 18,841 & 3.3 M  & 30 fps & 28.9 hours & \cmark & 223 & 17 & Train/Val/Test & \cmark  & \cmark & Semi-automatic\\
				\hline
			\end{tabular}
		}
	\end{center}
    }
\end{table*}

\section{Related Work}
\label{sec:relatedworks}

\subsection{Benchmarks for Tracking via Bounding Boxes}
According to different tasks, existing benchmarks for tracking via bounding boxes can be divided into two main categories, \ie GOT and UAV tracking benchmarks. In recent years, numerous datasets, \eg OTB~\cite{wu2013online,wu2015otb}, ALOV++~\cite{smeulders2013visual}, VOT~\cite{kristan2016novel}, TC-128~\cite{liang2015encoding}, NUS-PRO~\cite{li2015nus}, NfS~\cite{kiani2017need}, OxUvA~\cite{valmadre2018long}, TrackingNet~\cite{muller2018trackingnet}, and GOT-10k~\cite{huang2019got}, have been developed for GOT evaluation. OTB2013~\cite{wu2013online} and its extended version (OTB2015~\cite{wu2015otb}) are two widely used datasets with 50 and 100 sequences, respectively. NUS-PRO~\cite{li2015nus} focuses on tracking people and rigid objects. VOT~\cite{kristan2016novel} is an annual visual object tracking challenge that has been held alternately in ICCV and ECCV workshops since 2013. TC-128~\cite{liang2015encoding} is a color dataset, NfS~\cite{kiani2017need} is a high-frame-rate dataset, OxUvA~\cite{valmadre2018long} is a long-term dataset, and ALOV++~\cite{smeulders2013visual} is composed of 304 short sequences and 11 long sequences. TrackingNet~\cite{muller2018trackingnet} is a sparse (labeled every 30 frames) tracking dataset, while GOT-10k~\cite{huang2019got} is a large-scale one-shot tracking dataset. Inspired by these large-scale GOT datasets, we propose constructing a similar-scale UAV tracking dataset.

\begin{figure}[t]   
\vspace{-0.1cm}
\centering\centerline{\includegraphics[width=1.0\linewidth]{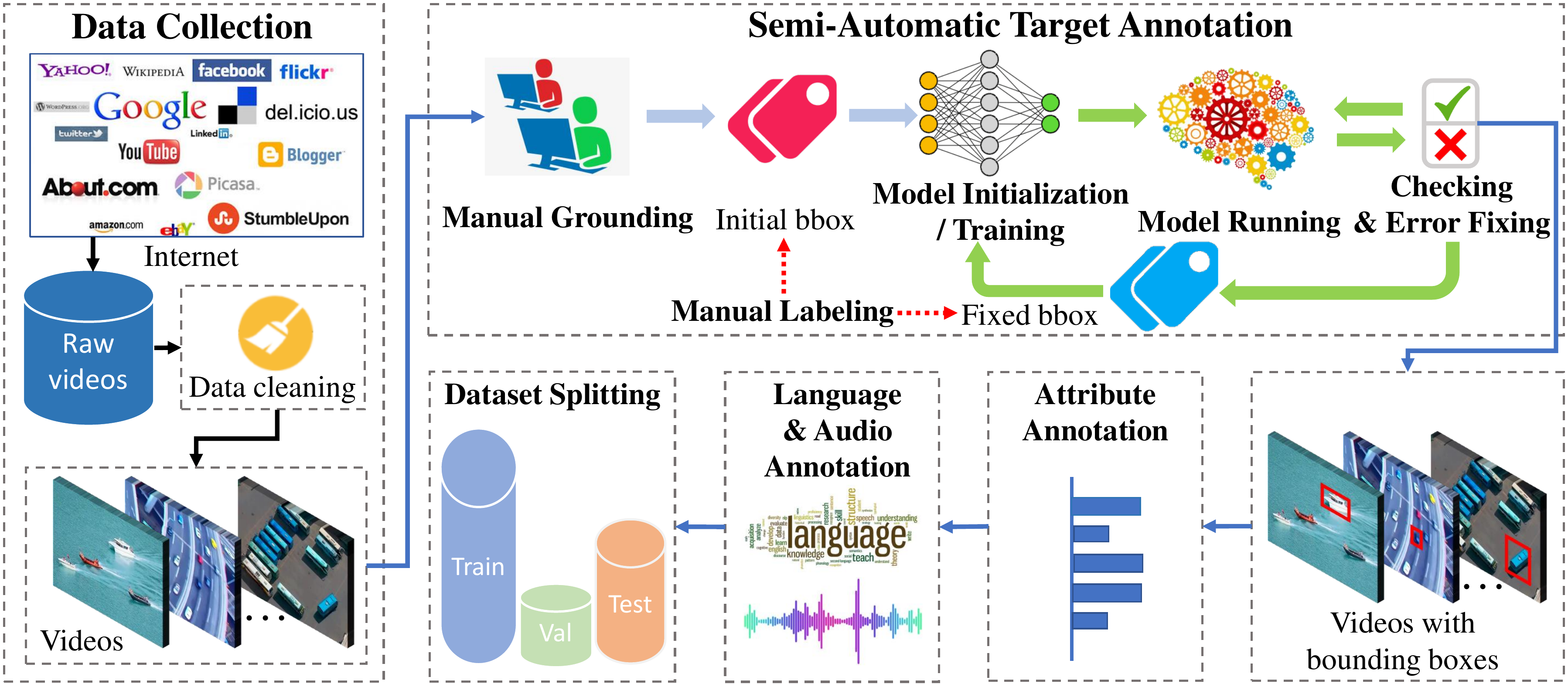}}
	\caption{An overview of the construction of WebUAV-3M. During the process of data collection, we first download videos from the internet to collect raw videos and then perform data cleaning (\ie video naming and cropping) to obtain cleaned videos for annotation. After that, a SATA pipeline is used to label the massive WebUAV-3M dataset. Finally, diverse attributes, natural language specifications, and audio descriptions are provided to enrich the dataset, which is further divided into a unified dataset, including training, verification, and test sets.} 
	\label{fig:Pipeline}
\end{figure}

UAV123 and UAV20L~\cite{mueller2016benchmark} are two datasets for UAV tracking, comprising 123 short and 20 long sequences, respectively. DTB70~\cite{li2017visual} provides 70 videos with manually annotated bounding boxes in all video frames. UAVDT~\cite{du2018unmanned} comprises 50 videos and eight kinds of attributes. UAVDark135~\cite{li2021all} is a UAV nighttime tracking benchmark comprising 135 sequences and more than 125k manually annotated frames. VisDrone~\cite{zhu2020vision} is an annual visual object tracking challenge held every year in conjunction with the ECCV and ICCV workshops since 2018.  

To date, WebUAV-3M is the largest UAV tracking benchmark with high-quality, dense bounding box annotations produced. Compared with others, WebUAV-3M is the most extensive benchmark, including 223 target categories, 63 motion classes, and 17 challenging global attributes for reliable UAV tracking evaluation. In addition, WebUAV-3M provides extra lingual and audio descriptions for each video for the first time. The detailed comparisons of WebUAV-3M with existing UAV tracking benchmarks and GOT benchmarks are demonstrated in Table~\ref{tab:Comp_WebUAV_3M}.

\subsection{Benchmarks for Tracking via Language/Audio}
Only a few benchmarks have been developed for performing tracking via natural language/audio with respect to this new rising topic. Some GOT benchmarks, \eg LaSOT~\cite{fan2019lasot} and TNL2K~\cite{wang2021towards}, provide both visual bounding box annotations and natural language specifications. The lingual OTB99~\cite{li2017tracking} dataset was constructed by adding a natural language description of each target to the videos in OTB2015. In addition, \cite{beal2003graphical}, AVDIAR~\cite{gebru2018audio}, and AVOT~\cite{wilson2020avot} are audio-video datasets for tracking moving objects in videos. The Auditory Vehicle Tracking~\cite{gan2019self} dataset includes 3,243 short video clips, which can be used to track the moving vehicles in the reference frame purely from stereo sound and camera metadata without any visual inputs.

The aforementioned tracking benchmarks were all mainly designed for GOT. One issue with the existing benchmarks is that their videos do not contain serious viewpoint changes, continuous camera motions, and dark night scenarios from the UAV viewpoint. This limits the application of existing trackers in the real world. In addition, these benchmarks ignore adversarial examples, which restricts the development of adversarial learning-based trackers~\cite{jia2020robust,liang2020efficient,wang2021towards}. In contrast, the proposed WebUAV-3M dataset is specifically designed for UAV tracking in various challenging scenarios, including low light, long-term occlusion, small targets, high-speed motion, target distortions, dual-dynamic disturbances, and adversarial examples. Scenario-specific subtest sets and baseline results are also provided, which will benefit future research.

\section{Construction of WebUAV-3M}
In this section, we introduce the motivations for building such a large-scale labeled dataset and the details of the construction process of WebUAV-3M. To ensure high-quality annotations for all the data, we conduct a quality control process for dataset construction (see Table~\ref{tab:quality_control}).

\begin{table}[!t]
	{\scriptsize
	\renewcommand\arraystretch{1.0}
	\caption{Quality control process for dataset construction. Data annotation, attribute annotation, and language and audio annotation (marked with *) are conducted by a professional data annotation team (approximately ten people).  ``$\times $N'' denotes that the associated stage is performed N times.}
	\label{tab:quality_control}
	\begin{center}
		\setlength{\tabcolsep}{2.25mm}{
			\begin{tabular}{|c|c|c|c|}
				\hline
			    Stage & Description & Executor  & Proportion \\
				\hline
			    \hline	
			    
				1 & Data collection & Collectors & $100\%$  \\
				\hline
				
				2 & Data verification & The authors & $100\%$  \\
				\hline
				
				3* & Data annotation & Annotation team & $100\%$  \\
				\hline
				
				4 & Annotations verification  & Verification team  &  $75\%$  \\
				\hline
				
				5 & Annotations verification $\times 3$ & The authors & $100\%$  \\
				\hline
				
				6* & Attribute annotation  & Annotation team & $100\%$  \\
				\hline
				
				7 & Attribute verification  $\times 3$ & The authors & $100\%$  \\
				\hline
				
				8* &  Language and audio annotation  & Annotation team & $100\%$  \\
				\hline
				
				9 &  Language and audio verification $\times 3$ & The authors & $100\%$  \\
				\hline
				
				10 &  Dataset acceptance & The authors & $100\%$  \\
				\hline
			\end{tabular}
		}
	\end{center}
    }
\end{table}

\subsection{Motivations for Dataset Construction}
In the era of deep learning, large-scale and precisely labeled datasets have played important roles in various computer vision tasks~\cite{russakovsky2015imagenet}. This has also profoundly affected the field of deep UAV tracking, which started to train models from large-scale annotated video data, rather than just single-frame static images. This change is very natural and reasonable since UAV tracking is a model-free task that requires the learning of discriminative features from videos to perform accurate moving object localization~\cite{mueller2016benchmark,zhu2020vision}. Nevertheless, existing large-scale image datasets, such as the most popular image classification dataset (ImageNet~\cite{russakovsky2015imagenet}), contain more than 10 M images that enable the learning of low-level visual features from static images but are not suitable for training deep tracking models~\cite{fan2019lasot}. Thus, this initially inspired us to construct a large-scale annotated video dataset for deep UAV tracking. Until very recently, some video object detection datasets, \eg ImageNet VID~\cite{russakovsky2015imagenet} and YouTube-BB~\cite{real2017youtube}, have been widely used to improve the performance of deep tracking models. Despite their large scales, these datasets are not ideally suitable for UAV tracking. First, in many videos, the target is almost stationary throughout the video, which is not conducive to learning motion information. Second, the initial frames of some videos contain incomplete targets, making them less optimal for deep UAV tracking~\cite{huang2019got}. Another observation that inspired us to propose WebUAV-3M is that we noticed that large-scale densely annotated GOT datasets (\eg GOT-10k~\cite{huang2019got}, LaSOT~\cite{fan2019lasot}) had brought great recent advances in GOT. This inspired us to construct a similar-scale UAV tracking dataset with diverse target categories and multi-modal annotations to facilitate research in this area. Finally, constructing such a large-scale annotated dataset will not only facilitate the development of UAV tracking approaches in the near future but also promote the development of other computer vision tasks from a broader perspective. Furthermore, although recent self-supervised and weakly supervised learning methods have made great progress in tasks such as image classification and detection, learning generalizable features for recognizing objects or understanding language is still challenging~\cite{wang2021towards,li2017tracking}. In particular, considering the complexity (\eg tiny targets, low light, and occlusion) of UAV tracking tasks~\cite{zhu2020vision}, large-scale supervised learning with labeled data is still the domain paradigm for the foreseeable future.

\subsection{Data Collection}
Our benchmark contains 4,500 video sequences with more than 3.3 million frames. Most of them are downloaded and clipped from YouTube under Creative Commons licenses\footnote{https://creativecommons.org/licenses/}. First, we collect an initial pool of raw videos from the internet (see Fig.~\ref{fig:Pipeline}). In this process, keywords such as \textit{aerial photography}, \textit{aerial video}, \textit{drone}, and \textit{UAV} are used to search and download more than 28k videos from YouTube. Note that only videos longer than 10 seconds are kept. In addition, defective videos that contain repeated scenes (\ie those with similar appearances and motion patterns), long-term stationary targets, or noisy segments (\eg nondrone shooting, incomplete trajectories, or massive targets that fill over half the screen) are manually removed. As a result, we obtain 3,617 videos with at least one potential target object. In addition, to promote the diversity and heterogeneity of the dataset, we also borrow 616 and 267 videos from two aerial-view video datasets, Stanford Drone~\cite{robicquet2016learning} and Okutama-Action~\cite{barekatain2017okutama}, respectively. These videos are reannotated via the SATA pipeline by our professional annotation team. Finally, we obtain 4,500 videos for the final pool of raw videos.

We perform data cleaning on these videos for further annotation. Specifically, we ask the collectors to manually crop out the segment of the moving target from each entire video. During this process, the collectors randomly select moving targets in the videos to ensure the diversity of the dataset. After that, the collectors rename the cropped videos based on their corresponding targets. All videos are divided into 12 superclasses with reference to Wordnet~\cite{miller1995wordnet}, including \emph{person}, \emph{building}, \emph{vehicle}, \emph{vessel}, \emph{public transport}, \emph{aircraft}, \emph{animal}, \emph{agricultural machinery}, \emph{industry machine}, \emph{plant}, \emph{artifact}, and \emph{natural object}. Each video in the \emph{person} superclass has two-dimensional labels: the target class (person) and motion class (\eg acting, biking, boating, bungee jumping, chute rafting, dancing, running, hiking, swimming, walking, skiing, and surfing). In comparison, the videos in the \emph{nonperson} superclasses are labeled with over 220 different target classes (\eg harvester, tractor, balloon, kite, horse, sheep, hat, beacon, chimney, excavator, wind turbine blade, reef, coach, light rail, box truck, minivan, barge, and motorboat). Afterward, the authors of this work check all collected videos and perform the last screening step (determining whether to accept each video or not). All rejected videos are replaced with new videos belonging to the same category. This data verification process ensures that the accepted videos are of high quality and are suitable for UAV tracking tasks. 

Eventually, we compile a large-scale dataset by gathering 4,500 cropped video sequences with over 3.3 million frames, 12 superclasses, 223 target classes, and 63 motion classes. The statistics of WebUAV-3M are illustrated in Table~\ref{tab:Comp_WebUAV_3M}. We significantly expand the existing UAV tracking datasets; the total number of frames is increased by 23 times, and the number of target classes is increased by 22 times. The average video length of WebUAV-3M is 710 frames (\ie 23.7 seconds at 30 frames per second (FPS)). The shortest video contains 40 frames (\ie 1.3 seconds), while the longest video consists of 18,841 frames (\ie 628.0 seconds). Screenshots of some representative videos collected in WebUAV-3M are shown in Fig.~\ref{fig:Example_sequences}.

\begin{table}[t]
	{\scriptsize
	\renewcommand\arraystretch{1.0}
	\caption{Comparison between the times spent per bounding box by our SATA method and other annotation tools.}
	\label{tab:time}
	\begin{center}
		\setlength{\tabcolsep}{2.35mm}{
			\begin{tabular}{|l|c|c|c|c|c|}
				\hline
				\multirow{2}*{Annotation tool} &\multirow{2}*{Type}  & \multicolumn{4}{c|}{Time per bounding box} \\
				\cline{3-6}
				& & Easy & Medium & Hard & Mean \\
				\hline
			    \hline
			    
			    Labelme~\cite{russell2008labelme} & Manual & 13.23 s  & 16.08 s & 18.11 s  & 15.81 s\\
				\hline 
				
				VoTT\footnotemark& Manual  & 6.87 s  & 8.61 s & 8.67 s  & 8.05 s\\
				\hline
				
				ViTBAT~\cite{biresaw2016vitbat} & Semi-automatic  & 2.50 s  & 5.60 s & 5.96 s  & 4.68 s\\
				\hline
				
				CVAT$^{\ref{CVAT}}$ & Semi-automatic  & 1.71 s  & 3.76 s & 6.12 s  & 3.86 s\\
				\hline
				
				\textbf{SATA (Ours)} & Semi-automatic   & \textbf{0.11 s}  & \textbf{2.94 s} & \textbf{5.91} s  & \textbf{2.99 s} \\
				\hline
			\end{tabular}
		}
	\end{center}
    }
\end{table}
\footnotetext{https://github.com/microsoft/VoTT}

\begin{figure}[t]   
	\centering\centerline{\includegraphics[width=1.0\linewidth]{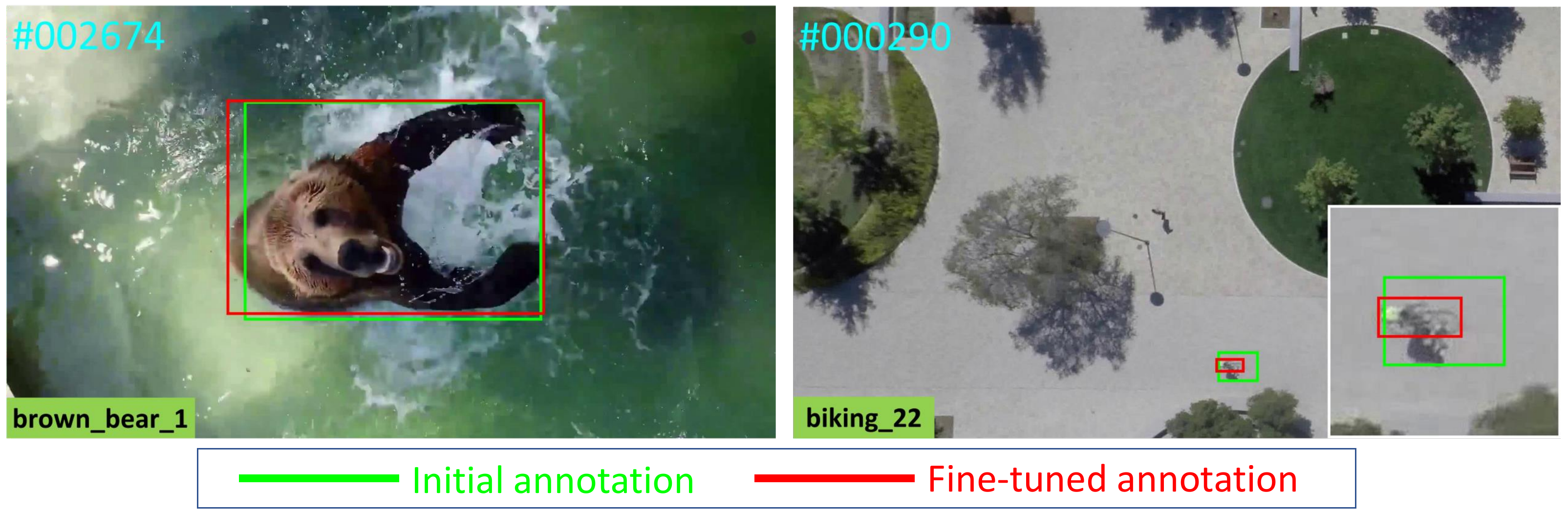}}
	\caption{Examples of accurate annotations generated by SATA.}
	\label{fig:accurate_annotations}
\end{figure}

\subsection{Semi-Automatic Target Annotation}
The newly proposed SATA pipeline is shown in Fig.~\ref{fig:Pipeline}. Once the cleaned videos are ready to be fed into the annotation process, the SATA pipeline can be divided into five steps.

\textbf{Step 1: Manual Grounding.} After each video is cleaned, the tracking target of interest in the first frame is manually selected randomly, and the ground-truth object bounding box, \ie $[x_1, y_1, w, h]$ is drawn, where $(x_1, y_1)$, $w$, and $h$ are the left corner point, width, and height of the target, respectively.
 
\textbf{Step 2: Manual Labeling}. The human-annotated object bounding box of the current frame is obtained. The object bounding box can either be acquired from the Manual Grounding step (for the first frame) or the Checking $\&$ Error Fixing step (for subsequent frames in which the model running step generates incorrect or poor-quality labels).

\textbf{Step 3: Model Initialization/Training}. Except for the initial bounding box (bbox) obtained from the first frame that initializes the pretrained tracking model, the other bounding boxes (fixed bbox) are ready for training the tracking model $\mathcal{M}$. $\mathcal{M}$ can execute the optimization process based on the loss between the ground truth (from the last Step 5) and the predicted bounding box (from the last Step 4). Additionally, the training step can be manually skipped when the model does not need further training. As mentioned in~\cite{muller2018trackingnet}, dense annotation can be assisted with state-of-the-art single object trackers. In this work, we adopt an off-the-shelf deep tracking model~\cite{LiWWZXY19} for semi-automatic annotation due to its efficiency and high performance.

\textbf{Step 4: Model Running}. The tracking model $\mathcal{M}$ generates a tracking prediction. Then, the frame indicator moves to the next frame. Specifically, the tracking model $\mathcal{M}$ takes the ground-truth bounding box parameter of the last frame and the image of the current frame as inputs and outputs the bounding box parameter of the tracking prediction for the current frame.

\textbf{Step 5: Checking $\&$ Error Fixing}.
{\color{black}The predicted bounding box is manually checked, refined, and fed into the next annotation loop until the video ends in a real-time interactive manner.} The annotation tool shows the bounding box obtained based on the prediction parameter generated from Step 4 with the current frame image. Annotators manually adjust the bounding box or keep the prediction based on the tracking quality.

In the above SATA steps, we extend the standard in object detection~\cite{su2012crowdsourcing}, the labeling of objects’ tight bounding boxes to form our own video annotation rules. The detailed descriptions of our annotation rules can be found in the~\textbf{supplemental material}.

\begin{figure}[t]   
	\centering\centerline{\includegraphics[width=1.0\linewidth]{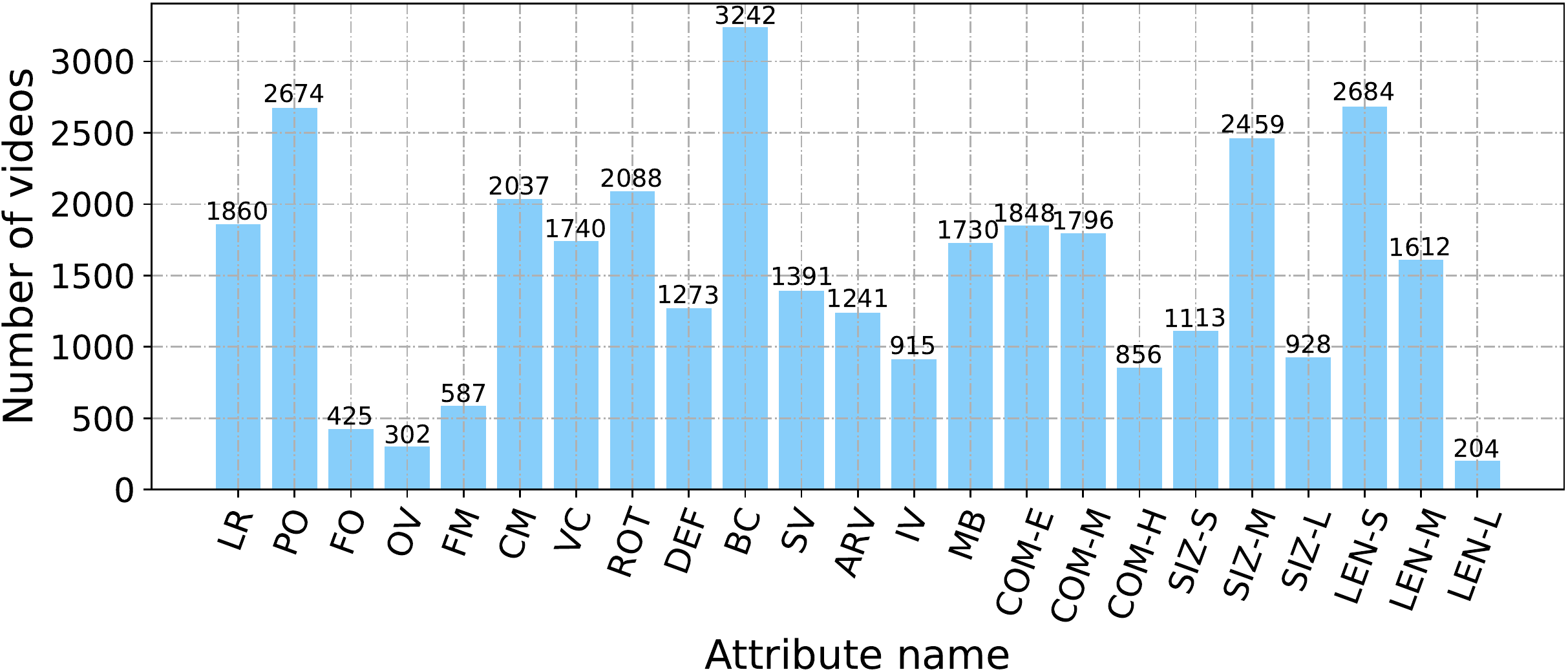}}
	\caption{Distribution of videos for each attribute.}
	\label{fig:distribution_of_videos}
\end{figure}

Considering the vast appearance variations, occlusions, rapid motions, and frequent disappearances of targets, most tracking datasets adopt semi-automatic (\ie tracking algorithms that generate preliminary annotations and manually correct failed frames)~\cite{biresaw2016vitbat,dai2021video} or manual~\cite{russell2008labelme} annotation methods to ensure labeling quality. In general, advanced semi-automatic tools can greatly improve labeling efficiency while delivering highly accurate bounding box annotations. Although many mature semi-automatic annotation tools are available in the community, such as the video tracking and behavior annotation tool (ViTBAT)~\cite{biresaw2016vitbat} and CVAT\footnote{https://github.com/openvinotoolkit/cvat\label{CVAT}}, each of them has its own drawbacks when facing specific tasks. The problems mainly concern the insufficient accuracy of predictive annotations and the lack of special functions. SATA uses the interactive iterative updating method with a high-performance tracking model to improve the overall annotation process. As a result, SATA attains a faster annotation speed and more accurate bounding boxes than those of popular semi-automatic/manual tools for the image or video annotation. In Table~\ref{tab:time}, we compare the times spent per bounding box by our SATA method and other annotation tools on easy, medium, and hard videos. Our SATA approach has a significant advantage in terms of average time consumption; it requires 2.99 seconds, for each bounding box annotation. Furthermore, our semi-automatic tool, SATA, can manually fine-tune the bounding boxes acquired from tracking algorithms to obtain more accurate annotations (left) and effectively fix incorrect annotations in an existing dataset~\cite{robicquet2016learning} (right), as shown in Fig.~\ref{fig:accurate_annotations}. {\color{black}In summary, SATA has three benefits: generating accurate bounding boxes in short segments by an advanced tracking model, real-time manual checking and simultaneous error fixing, and greatly reducing the total annotation time and human labor.} With all the above efforts, we finally reach a benchmark with high-quality, dense annotations; some examples are shown in Fig.~\ref{fig:Example_sequences}.

\subsection{Attribute Annotation}
Following popular tracking benchmarks~\cite{zhu2020vision,wu2015otb,huang2019got}, we further label each video sequence with multiple attributes to better analyze the performance of tracking algorithms. The proposed WebUAV-3M dataset has 17 attributes, including low resolution (LR), partial occlusion (PO), full occlusion (FO), out-of-view (OV), fast motion (FM), camera motion (CM), viewpoint changes (VC), rotation (ROT), deformation (DEF), background clutter (BC), scale variations (SV), aspect ratio variations (ARV), illumination variations (IV), motion blur (MB), complexity (COM), size (SIZ), and length (LEN). It is worth noting that these attributes not only contain specific challenging factors at the target level, such as LR (the target box is smaller than 400 pixels) and MB (the target region is blurred due to target or camera motion) but also indicate the challenging video-level factors, \ie COM (the complexity of the current video: easy (COM-E), medium (COM-M), or hard (COM-H)), SIZ (the size of the current video: small (SIZ-S), medium (SIZ-M), or large (SIZ-L)), and LEN (the length of the current video: short (LEN-S), medium (LEN-M), or long (LEN-L)). However, most of the existing benchmarks~\cite{zhu2020vision,huang2019got} do not explicitly provide these video-level attributes, which may seriously affect the obtained tracking results. For example, trackers are more likely to accumulate errors in long videos and cause model drift and tracking failures. Please refer to the~\textbf{supplemental material} for more details about the 17 attributes.

\begin{figure}[t]   
	\centering\centerline{\includegraphics[width=1.0\linewidth]{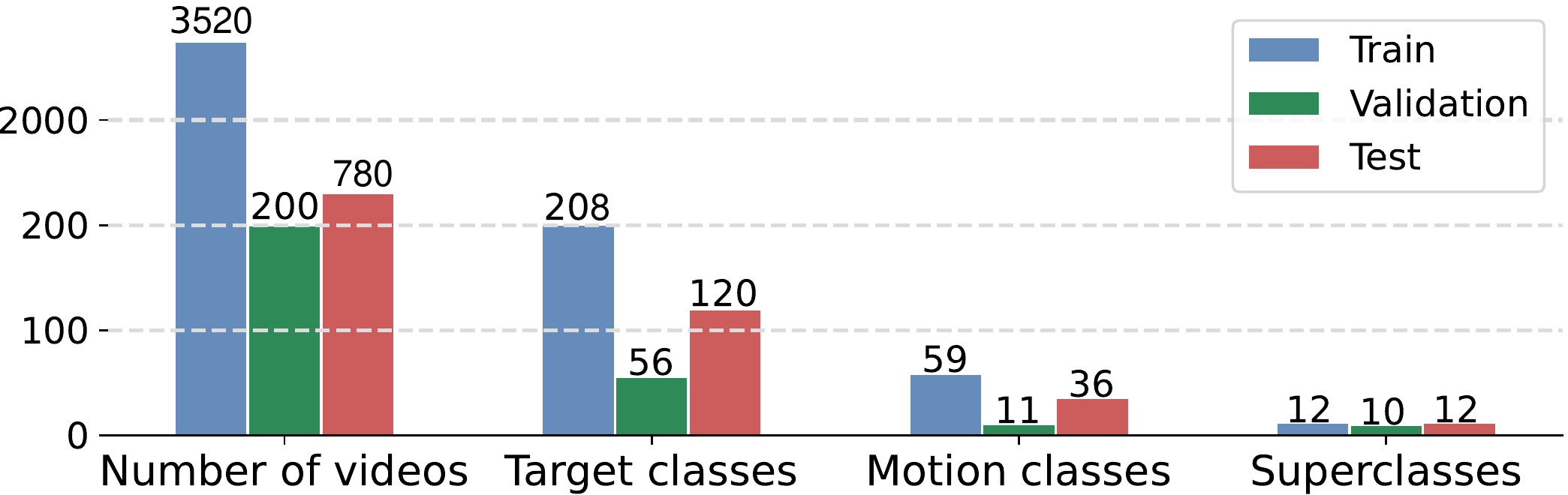}}
	\caption{Dataset splits of WebUAV-3M.}
	\label{fig:splits}
\end{figure}

The distribution of the sequences of each attribute in WebUAV-3M is shown in Fig.~\ref{fig:distribution_of_videos}. We can observe that the most common challenge factors in WebUAV-3M dataset are BC, occlusion (PO and FO), ROT, CM, and LR, which are well-known challenges for UAV tracking in real-world applications. In addition, WebUAV-3M is a moderately difficult dataset, which is mainly composed of medium-sized videos (with sizes between $\sqrt{640\!\times\!480}$ pixels and $\sqrt{1280\!\times\!720}$ pixels) and short- and medium-length videos (with lengths less than 1800 frames over 60 seconds for 30 FPS). Videos with diverse, challenging factors will provide a good platform for evaluating deep trackers. We also provide the co-occurrence distribution of the 17 attributes in the~\textbf{supplemental material}.

\subsection{Language and Audio Annotation}
As mentioned earlier, we provide natural language specifications and audio descriptions {\color{black}(see Fig.~\ref{fig:Example_sequences})} in WebUAV-3M to facilitate multi-modal UAV tracking. {\color{black}Specifically, WebUAV-3M contains approximately 800 English words and focuses on expressing the target’s class name, position (relative location), attribute, behavior, and surroundings via one English sentence for the whole video sequence~\cite{fan2019lasot,chen2021pix2seq}.} For WebUAV-3M, the annotation team labels 4,500 sentences for all videos. The natural language specifications can provide auxiliary information to achieve accurate tracking. Language can assist in reducing uncertainty as a kind of global semantic information when the target’s appearance changes significantly or similar distractors are present. {\color{black}For these sentences, we ask the annotation team to provide both female and male audio descriptions using a text-to-speech software (Balabolka)\footnote{http://balabolka.site/balabolka.htm}.} Note that the audio descriptions can also provide auxiliary help for tracking. In total, we obtain 9,000 audio descriptions. {\color{black}More details about the language and audio annotation are shown in the~\textbf{supplemental material}.}

\begin{figure*}[t]
\vspace{-0.2cm}
\begin{minipage}[t]{0.56\linewidth}
\centering
\includegraphics[width=0.32\linewidth]{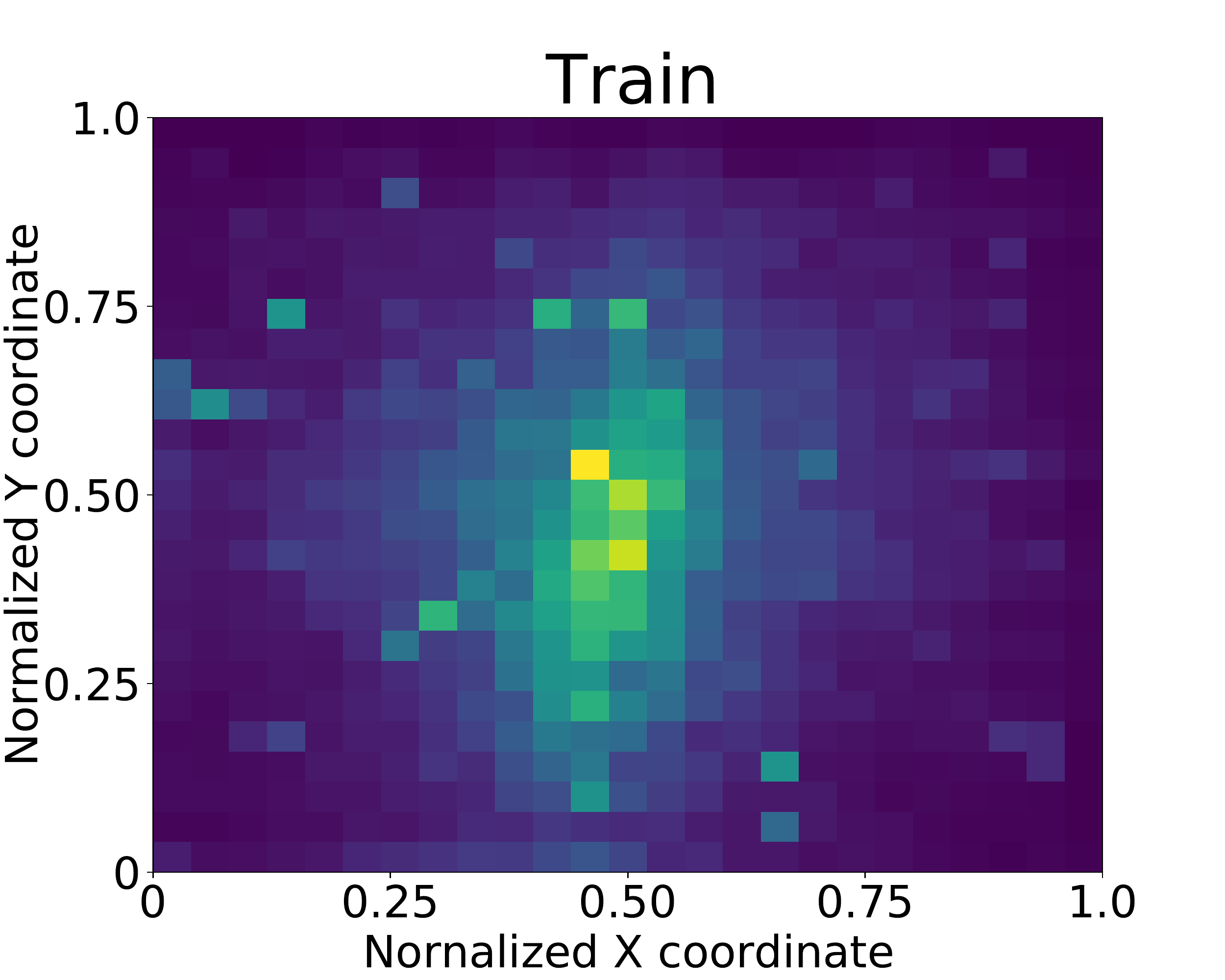}
\includegraphics[width=0.32\linewidth]{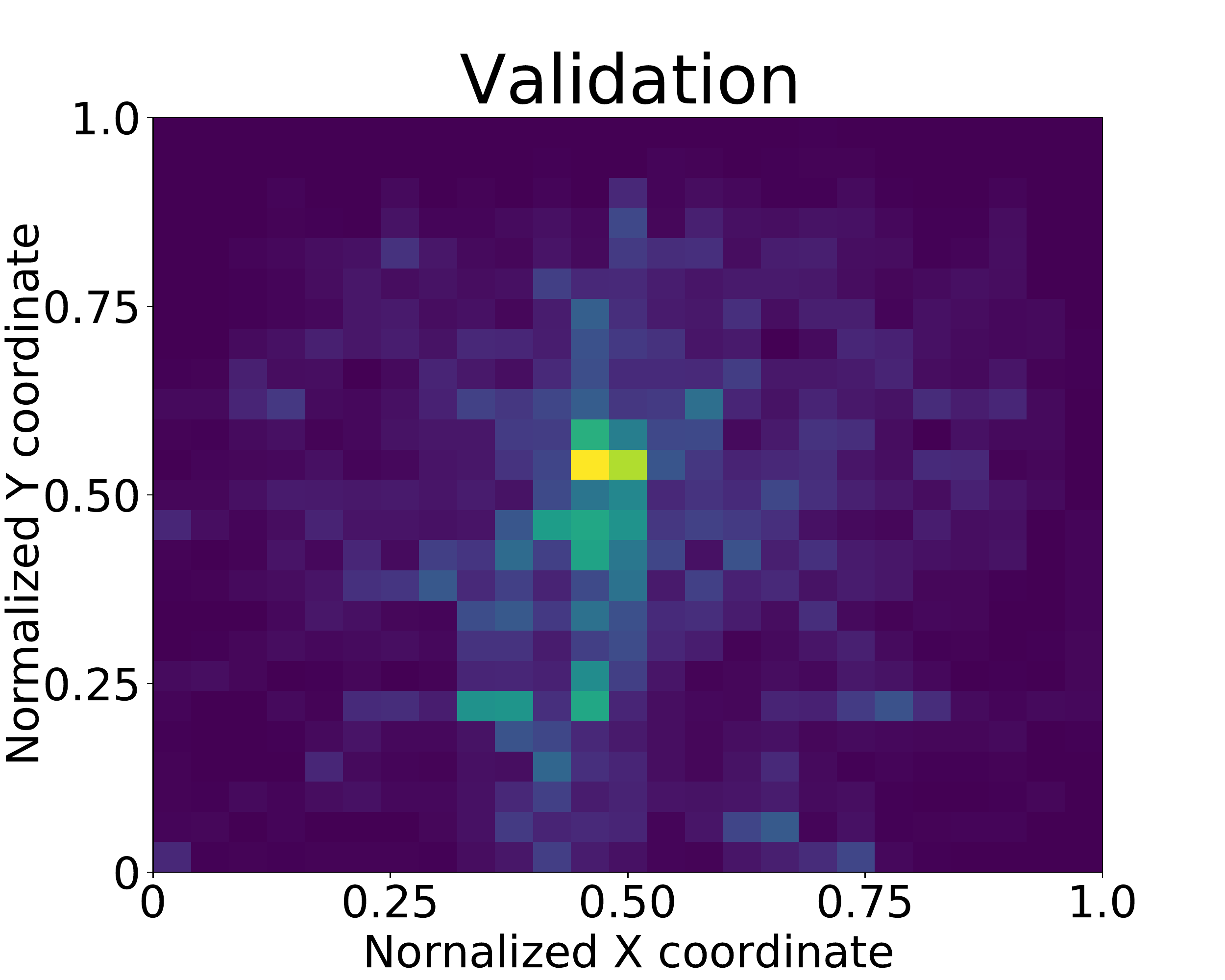}
\includegraphics[width=0.32\linewidth]{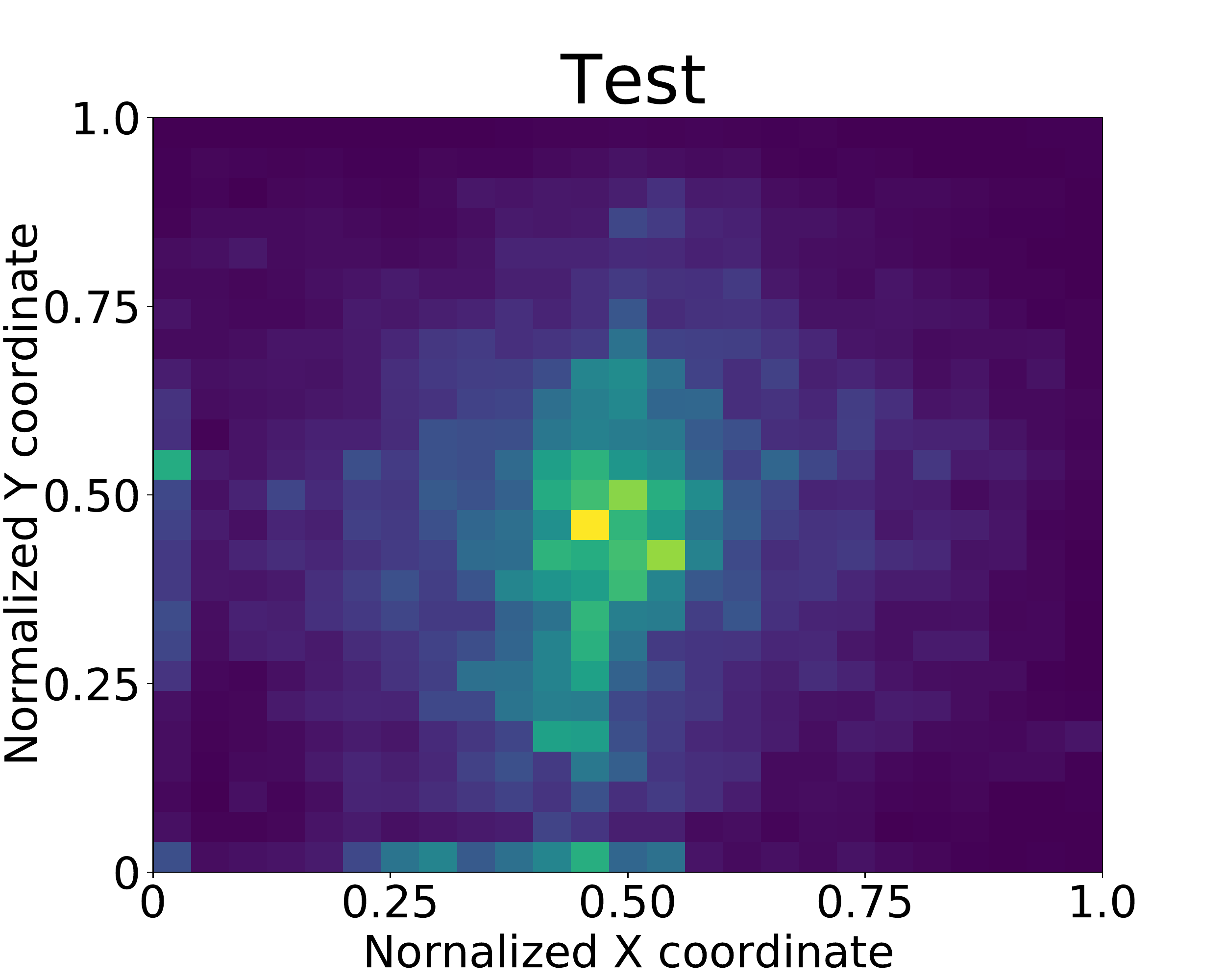}
\caption*{\color{black}(a) Position distribution}
\label{fig:position}
\end{minipage}
\begin{minipage}[t]{0.215\linewidth}
\centering
\includegraphics[width=1.0\linewidth]{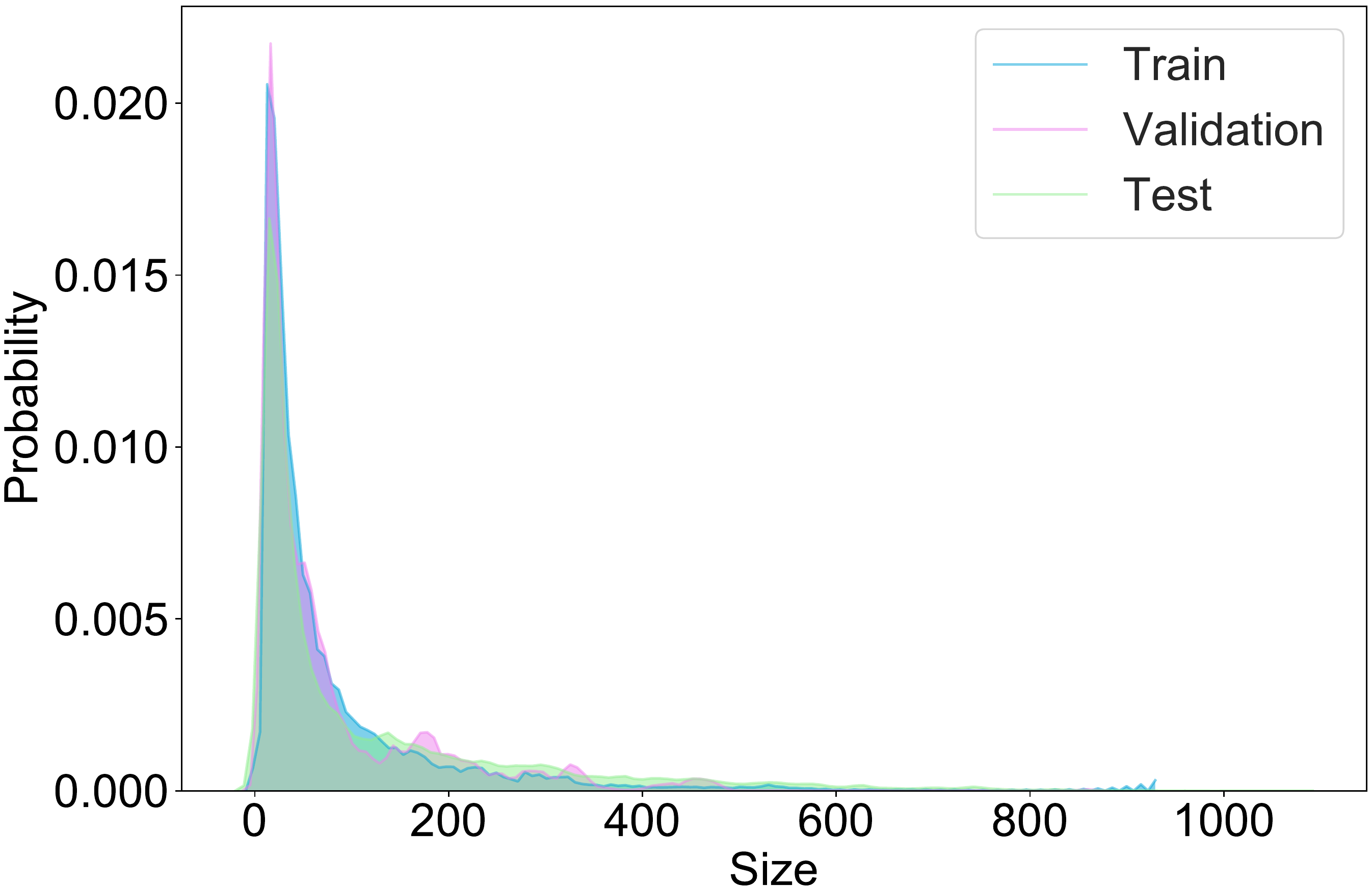}
\caption*{(b) Size distribution}
\label{fig:scale}
\end{minipage}
\begin{minipage}[t]{0.215\linewidth}
\centering
\includegraphics[width=1.0\linewidth]{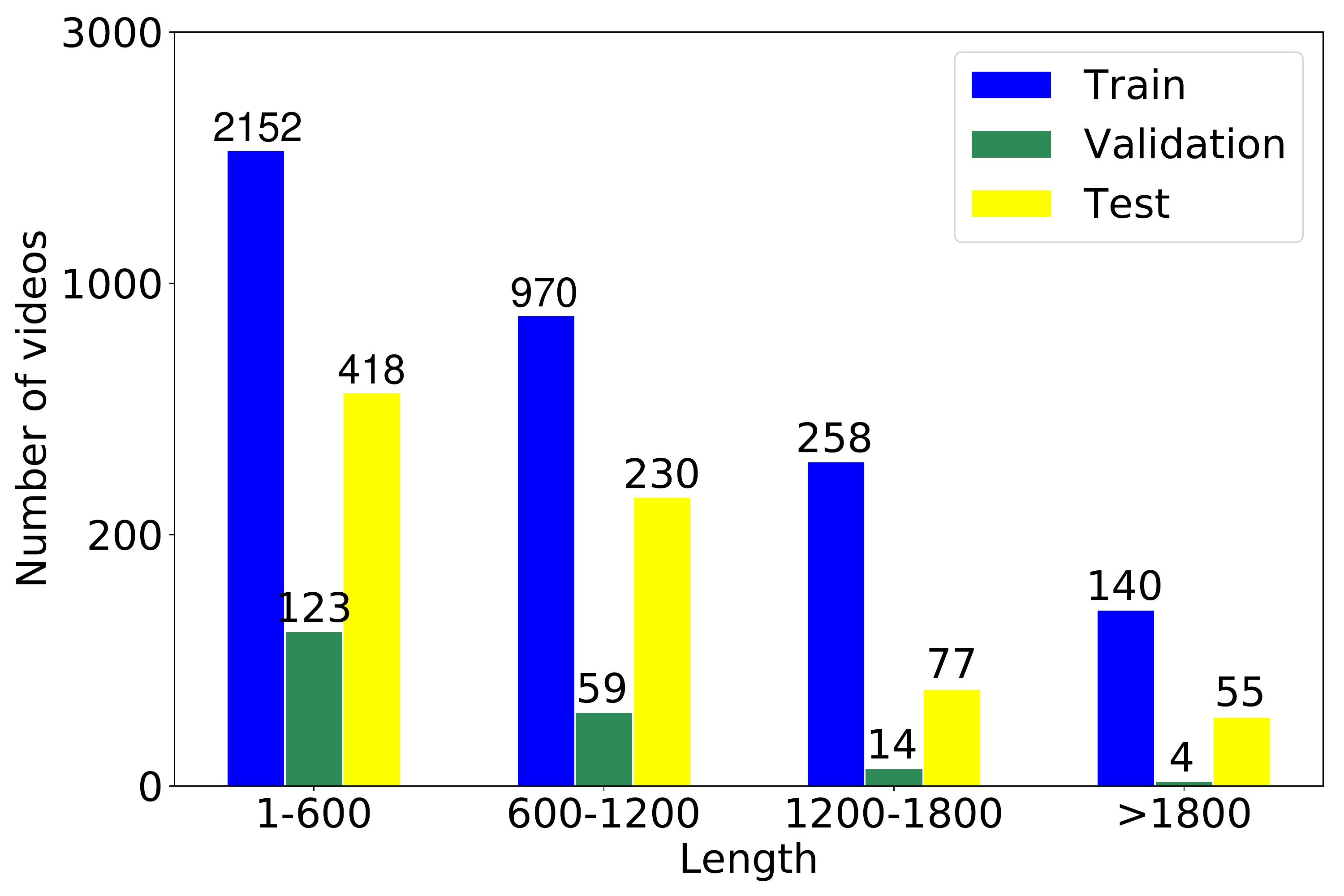}
\caption*{(c) Length distribution}
\label{fig:length}
\end{minipage}
\caption{Target position, size, and video length distributions in WebUAV-3M. Best viewed by zooming in.}
\label{fig:statistical_analysis}
\end{figure*}

\subsection{Dataset Splitting}
\label{sec:Dataset Splitting}
We split WebUAV-3M into training, validation, and test sets to provide a unified evaluation benchmark for deep UAV tracking. We aim to design a compact and informative test set containing medium-difficulty and multicategory videos. The evaluation results reliably reflect the generalization abilities of different algorithms for various seen and unseen target classes and motion classes. To achieve this, we provide the detailed splitting strategy in the~\textbf{supplemental material}.

The final splits of our WebUAV-3M dataset are shown in Fig.~\ref{fig:splits}. The training set contains 3520 videos, 208 target classes, 59 motion classes, and 12 superclasses. The test set includes 780 videos, 120 target classes, 36 motion classes, and 12 superclasses. We find that the number of instances in our test set is comparable to the number of videos in the test sets of TNL2K (700 videos) and GOT-10k (420 videos), which are both large-scale datasets with millions of labeled frames (see Table~\ref{tab:Comp_WebUAV_3M}). We argue that a sufficient number of videos and a compact test set can alleviate the problem of saturation/overfitting faced by small-scale datasets~\cite{wu2013online,wu2015otb}. Moreover, we construct an informative test set that covers various target classes, motion patterns, and most of the typical challenging scenarios to facilitate the assessment of deep UAV tracking algorithms.

\subsection{Statistical Analysis}
\myPara{Long-tail property.} As shown in Fig.~\ref{fig:Group_of_object_classes}, we can observe that the number of videos in each group of object classes in WebUAV-3M exhibits a long-tail distribution, which reflects the true distribution of the targets in the videos collected by the UAVs. For example, 1,305 and 362 videos are contained in the person and sedan class groups, respectively, while only {\color{black}4} and 2 videos are included in the {\color{black}balloon} and radio telescope categories. These long-tailed distributions of object classes pose a significant challenge when building accurate and robust models for the practical world.

\myPara{Position distribution.} {\color{black}The distribution of the normalized target center positions in WebUAV-3M is shown in Fig.~\ref{fig:statistical_analysis}(a). It shows that the targets in the training, validation, and test sets have similar position distributions, concentrated (\ie highlighted) in the central region of the images. Compared with those of the other two sets, the position distribution of the test set is presented as a center mean Gaussian, indicating the high quality of the test set.}

\myPara{Size distribution.}
The target size distribution is demonstrated in Fig.~\ref{fig:statistical_analysis}(b). The sizes of the targets vary widely in the entire dataset, from 10 to 1000 pixels. We find that the training, validation, and test sets have similar target size distributions. The average size of the target is approximately 50 pixels, which indicates that the UAV tracking task usually faces serious small target challenges.

\myPara{Length distribution.}
The video length distribution is shown in Fig.~\ref{fig:statistical_analysis}(c). We find that WebUAV-3M contains 2693, 1259, 349, and 199 videos with segments containing 1-600, 600-1200, 1200-1800, and more than 1800 frames, respectively. The four segments in the training and test sets comprise [2152, 970, 258, 140] and [418, 230, 77, 55] videos, respectively. The various video lengths in the test set are useful for evaluating short-term tracking algorithms (less than 600 frames for each video) and are suitable for evaluating long-term tracking algorithms (more than 1800 frames for each video).

\begin{figure*}[t]   
\vspace{-0.2cm}
\centering\centerline{\includegraphics[width=1.0\linewidth]{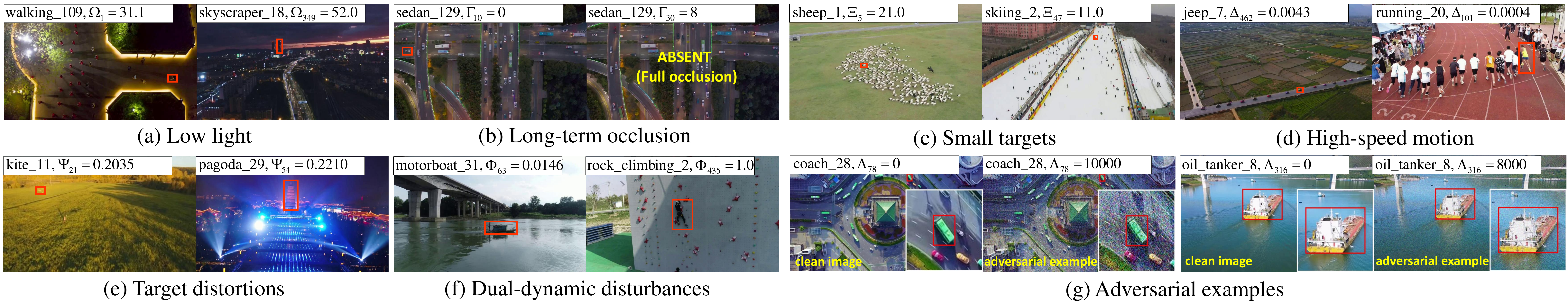}}
	\caption{Screenshots of some videos obtained from seven scenarios in WebUAV-3M. The video name and scenario indicator value of the current frame are presented at the top of the corresponding image.}
	\label{fig:Scenarios}
\end{figure*}

\section{UTUSC Protocol}
\label{sec:UTUSC}

\subsection{Evaluation Protocols}
Popular evaluation protocols for UAV tracking mainly target the pursuit of overall performance. For example, DTB70~\cite{li2017visual}, UAVDT~\cite{du2018unmanned}, UAVDark135~\cite{li2021all} and VisDrone~\cite{zhu2020vision} evaluate overall performance and attribute-based performance using tracking success and precision, as in~\cite{wu2015otb}. UAV123~\cite{mueller2016benchmark} not only applies an attribute-based evaluation but also provides an online evaluation to measure tracker performance in terms of different aspects (\eg the impact of a dynamic frame rate or the trajectory error between the target and the UAV motion) based on the Unreal Engine 4 simulator\footnote{https://www.unrealengine.com/en-US/}. Although the result of a global evaluation on a large set of videos is an important indicator of a tracker’s overall performance, such an aggregate measurement hides many subtleties that differentiate trackers and thus cannot reflect the weaknesses and strengths of different algorithms~\cite{valmadre2018long}. 

\begin{figure*}[t]
    \centering
    \vspace{-0.5cm}
    \subfloat[Low light]{\includegraphics[width =0.28\columnwidth]{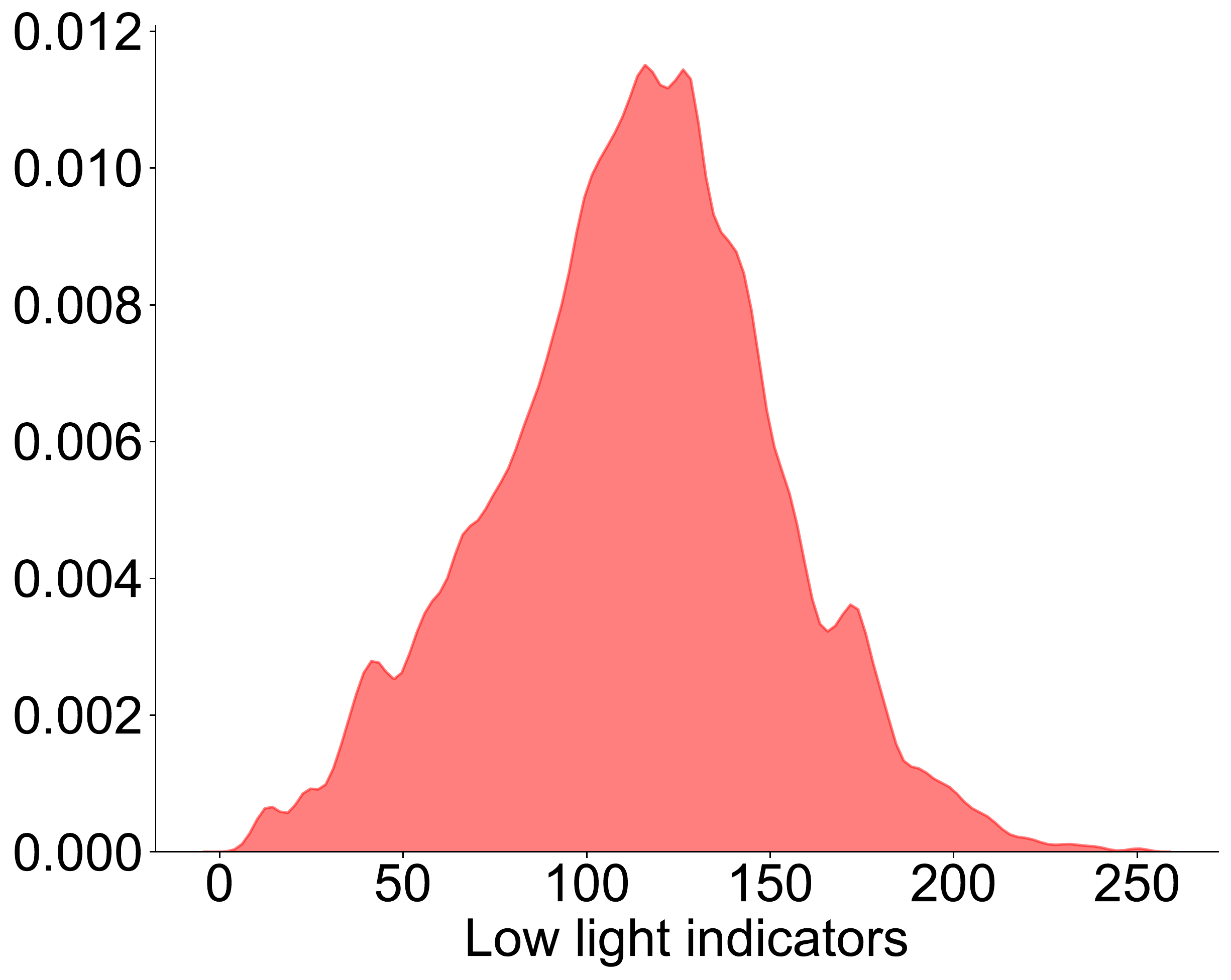}}
    ~
    \subfloat[Long-term occlusion]{\includegraphics[width =0.28\columnwidth]{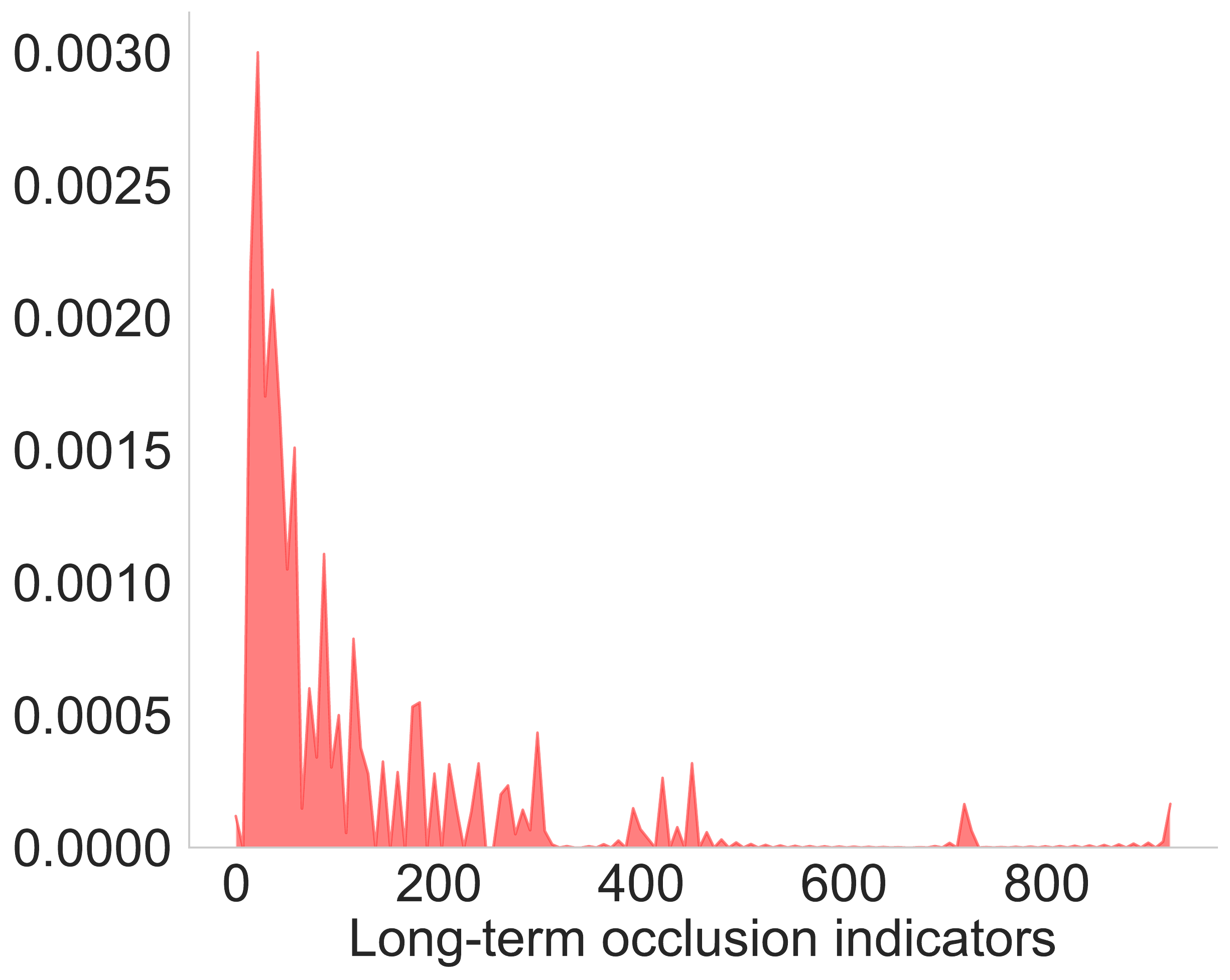}}
    ~
    \subfloat[Small targets]{\includegraphics[width =0.28\columnwidth]{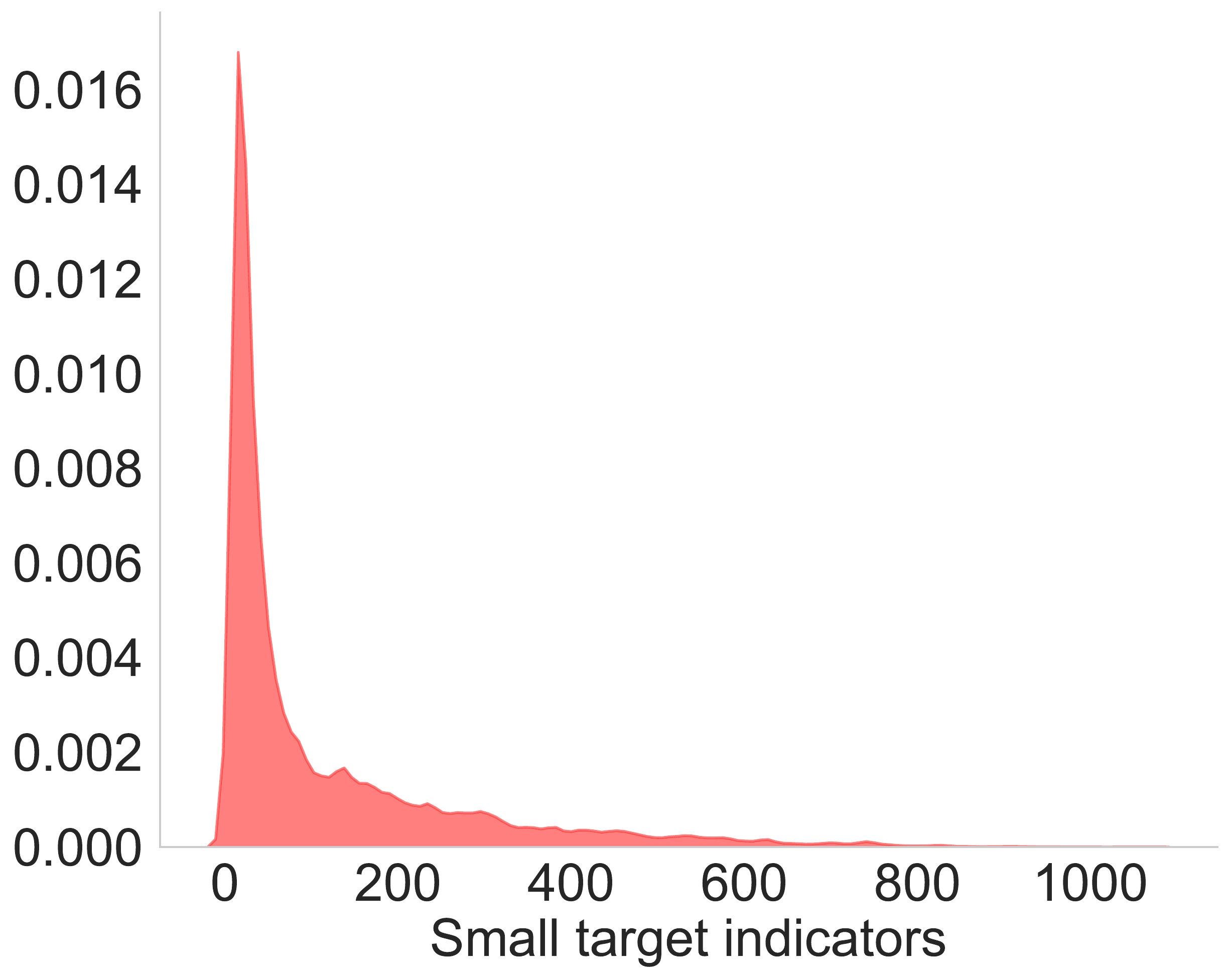}}
    ~
    \subfloat[High-speed motion]{\includegraphics[width =0.28\columnwidth]{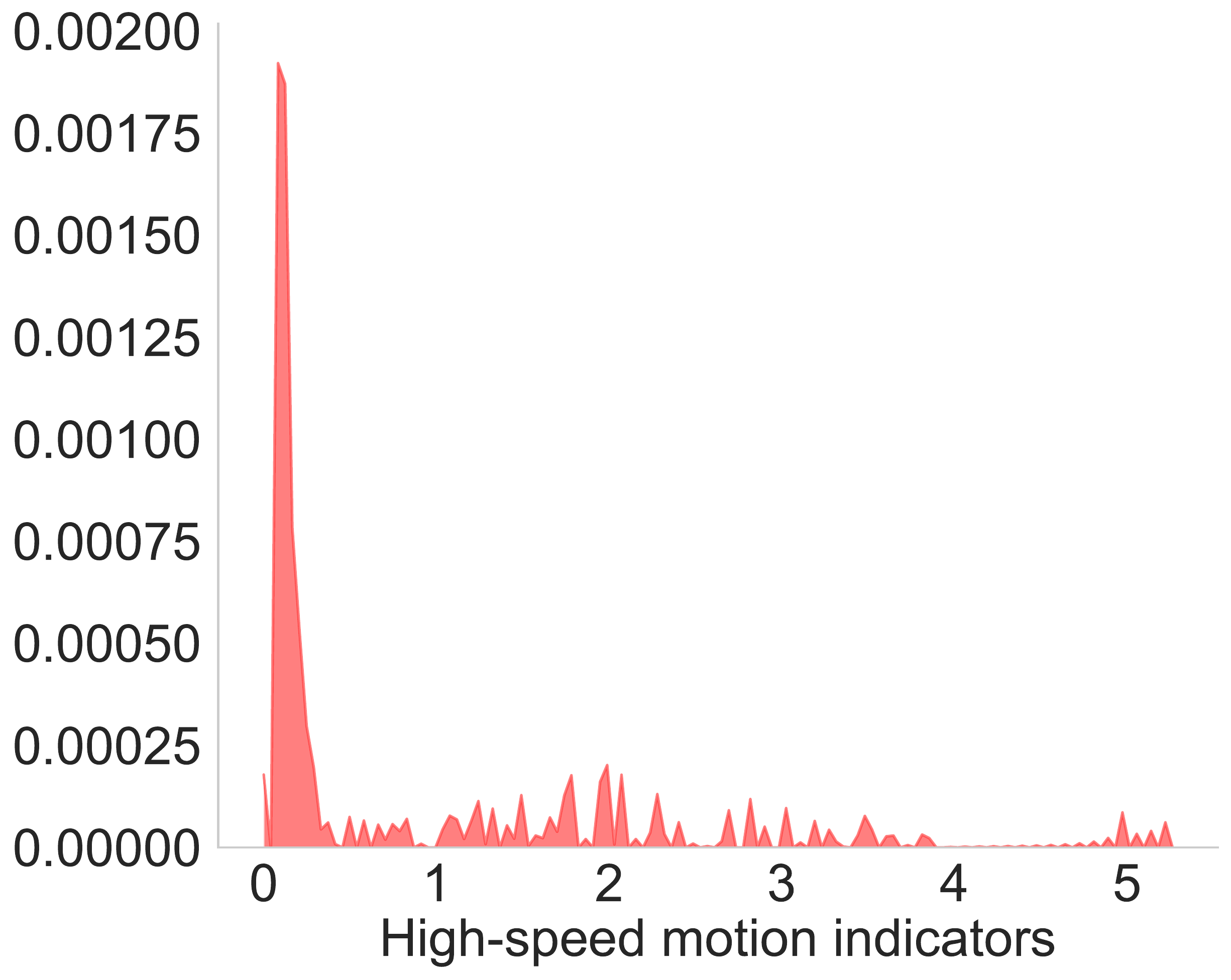}}
    ~
    \subfloat[Target distortions]{\includegraphics[width =0.28\columnwidth]{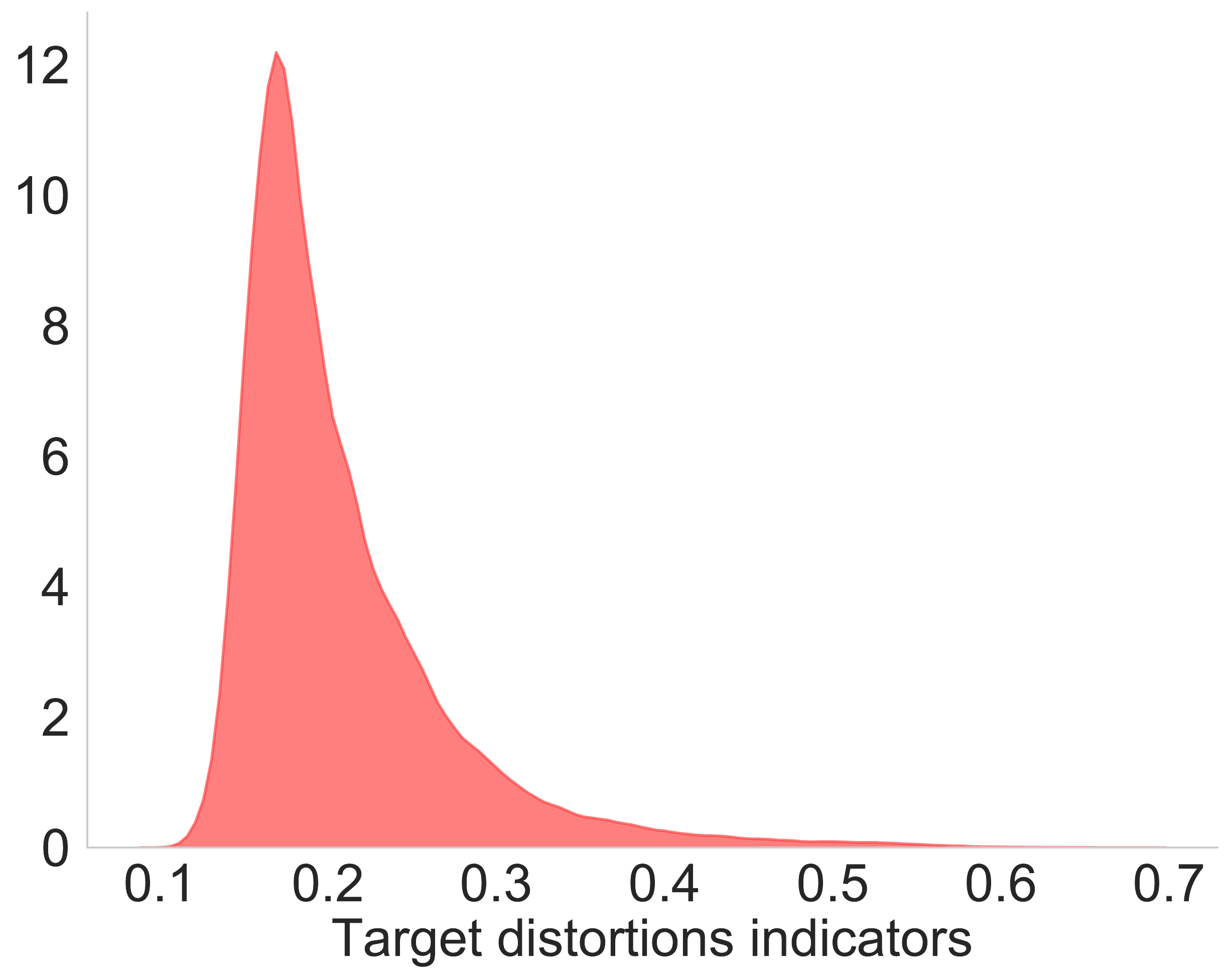}}
    ~
    \subfloat[Dual-dynamic disturbances]{\includegraphics[width =0.28\columnwidth]{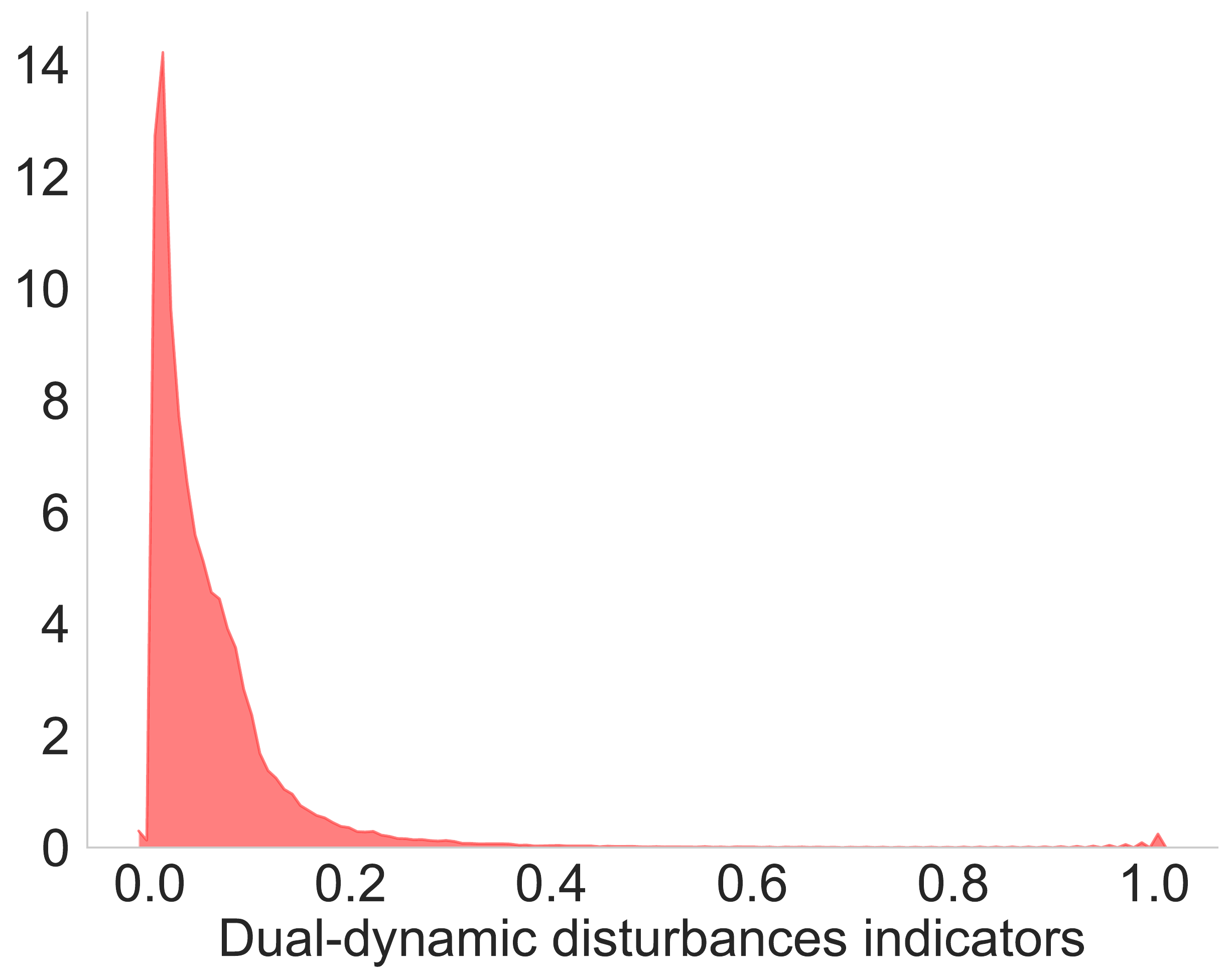}}
    ~
    \subfloat[Adversarial examples]{\includegraphics[width =0.28\columnwidth]{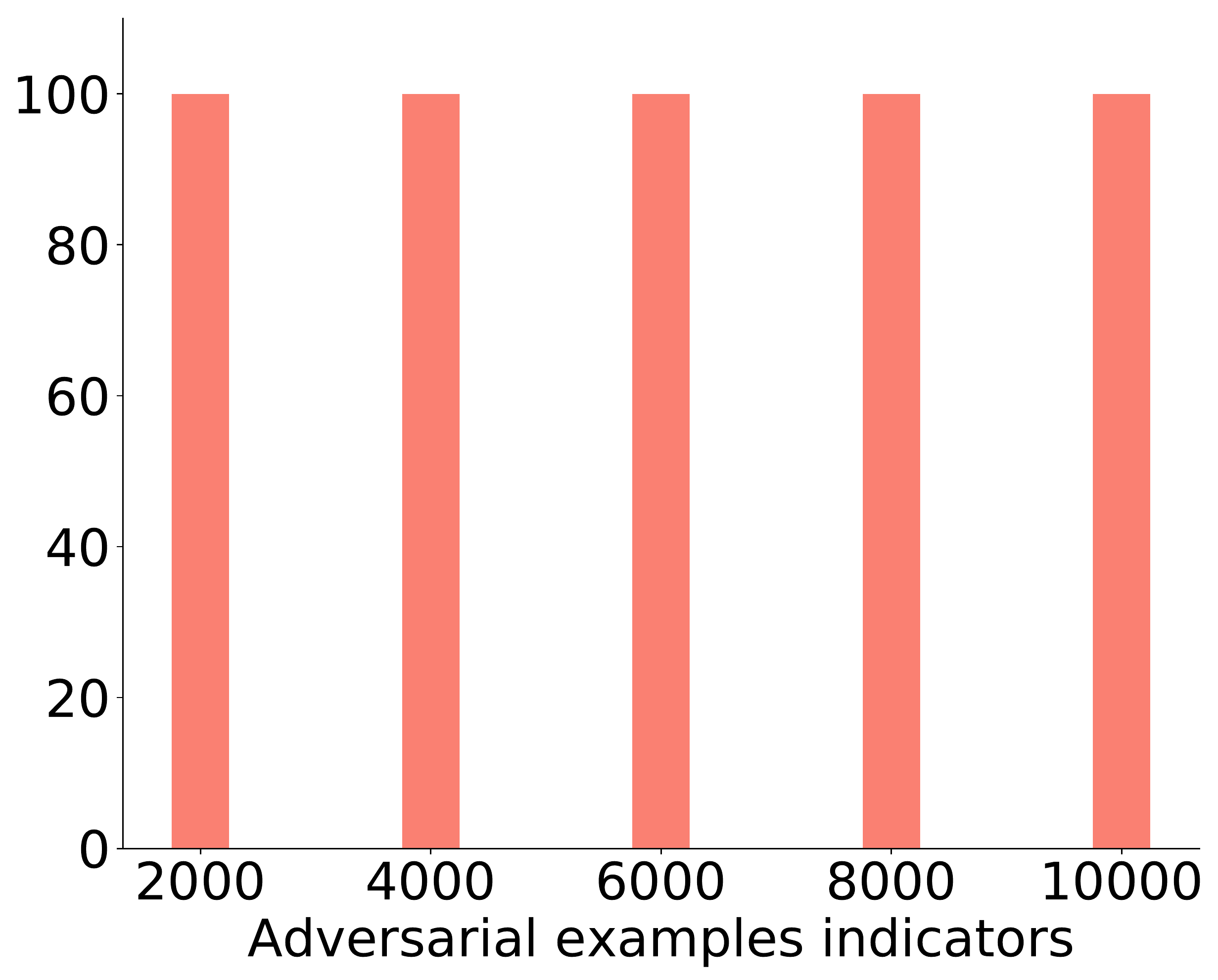}}
\caption{Scenario indicator distributions of the test set. The framewise difficulty indicators represent the degrees of challenge for current tracking algorithms. Best viewed by zooming in.}
\label{fig:scenario_distribution}
\end{figure*}

\begin{table*}[t]
	{\scriptsize
	\renewcommand\arraystretch{1.0}
	\caption{The details of the seven scenario subtest sets. $\#$ denotes the number of the corresponding item.}
	\label{tab:scenario_test_sets}
	\begin{center}
	\setlength{\tabcolsep}{1.2mm}{
    \begin{tabular}{|l|c|c|c|c|c|c|c|} 
    \hline
    \textbf{Scenario}  & \textbf{Low light}	&\textbf{Long-term occlusion}	&\textbf{Small targets}	&\textbf{High-speed motion}	&\textbf{Target distortions}	&\textbf{Dual-dynamic disturbances}	& \textbf{Adversarial examples} \\
    \hline 
    \hline 
    
    \textbf{$\#$Superclass} &11 & 10 & 10  & 11   & 12   & 11   & 10 \\
    
    \textbf{$\#$Motion class}  & 11 & 13  &  6   & 13  & 9  & 14   & 10 \\
    
    \textbf{$\#$Target class}  & 44 & 39  & 46   & 49  & 47  & 40   & 41 \\
    
    \textbf{$\#$Videos}  & 100 & 100  & 100   & 100   & 100  & 100   & 100 \\
    \hline 
	\end{tabular}
	}
	\end{center}
    }
\end{table*}

In this work, we design the UTUSC protocol, which enables researchers to comprehensively evaluate their UAV tracking algorithms in different application scenarios (see Fig.~\ref{fig:Scenarios}). Instead of using the highly subjective manual labeling approach for the global attributes in each frame or video sequence~\cite{zhu2020vision,mueller2016benchmark,fan2019lasot}, we decide to evaluate trackers based on more objective and scalable annotations. These indicators can be calculated directly from ground-truth bounding boxes and metadata (\eg video sequences). Concretely, we define a set of fine-grained and continuous indicators for each video frame. The UTUSC focuses on the following seven scenarios.

\myPara{Low light.} Videos captured in low light conditions often exhibit low visibility, which is likely to harm the performance of tracking algorithms designed primarily for high-visibility inputs. The degree of low light in frame t can be defined as the mean image intensity ${\Omega}_t= \frac{1}{3}\sum_{c=1}^{3}I_{c}$ (with four times the extension region centered on the target location), where ${I}_{c}$ denotes the image intensity of the RGB channels~\cite{richards1982lightness}. Low light scenarios aim to evaluate UAV tracking systems deployed in arbitrary light conditions (\eg searching for missing persons or rescuing survivors on dark nights).

\myPara{Long-term occlusion.} As demonstrated in many tracking benchmarks~\cite{zhu2020vision,huang2019got,fan2019lasot}, frequent occlusion (\ie long-term occlusion) is one of the most challenging factors that can easily cause model drifting and tracking failure~\cite{zhu2020vision}. Different from existing benchmarks (\eg \cite{huang2019got}) that provide manual annotations of object visibility ratios to facilitate occlusion-aware tracking methods, we explicitly define the duration that occlusion lasts in each frame through the number of continuous occlusion frames. Assuming that occlusion begins at the $t$th frame and ends at $t+k$th frame, we measure the occlusion duration between frames $t$ to $t+k$ frames as ${\Gamma}=\{0, 1, 2, k-1\}$. Long-term occlusion scenarios aim to evaluate robust UAV tracking algorithms deployed in environments with frequent occlusion (\eg traffic monitoring, security, and surveillance situations).

\myPara{Small targets.} A small target occupies a small proportion of an image, has a relatively low resolution, and can extract very few effective visual features, so the existing tracking algorithms cannot accurately predict the locations of such targets. We measure the target size in the $t$th frame as ${\Xi}_t={\Xi}(w_t, h_t)=\sqrt{w_t h_t}$, where $(w_t, h_t)$ denotes the ground-truth bounding box dimensions. Small target scenarios aim to compare different UAV algorithms in various real-world applications with low resolutions and restricted information due to small objects (\eg aerial photography and intelligent agriculture scenarios).

\myPara{High-speed motion.} Most previously developed tracking methods~\cite{li2018cvpr,DanelljanBKF19} heavily rely on the preset anchor boxes to regress the bounding boxes of a target in consecutive frames, so they cannot adapt to high-speed target motion. We measure the target speed in the $t$th frame relative to its size as follows~\cite{valmadre2018long},
\begin{equation}
\Delta_{t} = \frac{1}{\sqrt{s_{t-1}s_{t}}}\frac{||p_{t}-p_{t-1}||_2}{T_{t}-T_{t-1}},
\label{eq:highspeed}
\end{equation}
where $s_t=\sqrt{w_t h_t}$ denotes the target size, $p_{t}=(x_t, y_t)$ represents the center coordinates of the target, and $T_t$ denotes the time instant of frame $t$. High-speed motion scenarios aim to evaluate real-time UAV tracking algorithms deployed in tracking applications with fast motion, such as vehicle and athlete tracking.

\myPara{Target distortions.} Various target distortions (\eg brightening, white noise, and motion blur) easily result in target appearance distortion, making it difficult to extract representative features, thus decreasing the discriminative abilities of tracking algorithms. We measure target distortions via a meta-learning-based no-reference image quality assessment (IQA) model (MetaIQA) as follows~\cite{zhu2020metaiqa},
\begin{equation}
{\Psi}_t=f_{\theta}(x_t;\theta),
\label{eq:target_distortions}
\end{equation}
where $x_t$ is the input image (with four times the extension region centered on the target location) of frame $t$, ${\Psi}_t$ denotes the predicted target distortion score, and $f_{\theta}$ represents the model parameters. Target distortion scenarios can serve as distortion robustness evaluations for different methods deployed in situations with various image distortion levels (\eg security threat forecasting and emergency monitoring).

\myPara{Dual-dynamic disturbances.} UAV tracking is usually challenged by the dual-dynamic disturbances~\cite{zhang2020accurate} that arise from not only diverse moving targets but also motion cameras, leading to a more severe model drift issue than that encountered in traditional visual tracking scenarios. Dual-dynamic disturbances can be characterized by the motion of the ground truth and the abrupt motion of the camera in each frame. We define the degree of dual-dynamic disturbances in frame $t$ as follows:  
\begin{equation}
\Phi_{t} = \left\{ \begin{array}{l}
1, \textnormal{~if~} d(p_t, p_{t-1}) >= \sqrt{s_{t-1}} \textnormal{~and~}c_{t}=1, \\
\frac{d (p_t, p_{t-1})}{\sqrt{s_{t-1}}}, \textnormal{~if~} d (p_t, p_{t-1})\!<\!\sqrt{s_{t-1}} \textnormal{~and~} c_{t}\!=\!1, \\
0, \textnormal{~otherwise}.
\end{array} \right.
\label{eq:dual_dynamic}
\end{equation}
where $d (\cdot)$ denotes the Euclidean distance, $c_t$ equals 1 when frame $t$ has abrupt camera motion and 0 otherwise. Dual-dynamic disturbance scenarios can evaluate different tracking algorithms employed under challenging terrains and harsh conditions (\eg marine search and rescue).

\myPara{Adversarial examples.} Extensive works have proven that deep convolutional neural networks (CNNs) are vulnerable to adversarial attacks that add visually imperceptible noises to original images~\cite{jia2021iou,liang2020efficient}. To provide a good platform for the study of deep UAV tracking algorithms that are robust against adversarial attacks, we generate {\color{black}general} adversarial examples on top of some randomly selected video sequences in our WebUAV-3M dataset by leveraging an iterative orthogonal {\color{black}IoU} attack toolkit~\cite{jia2021iou}. Formally, an adversarial example is written as
\begin{equation}
I_{k+1}^{j}=\varsigma(I_k, \eta^{j} )+\epsilon \cdot \psi (I_k, \eta^{j}), ~~s.t.~~\rho (I_{k+1}^{j}-I_{0})<M,
\label{eq:adversarial}
\end{equation}
where $k$ is the iteration index, $\epsilon$ is a parameter, $\{\eta^{j}\}_{j=1}^{n}$ denotes $n$ random perturbations, $\varsigma(I_k, \eta^{j} )$ is the tangential perturbation and $\psi (I_k, \eta^{j})$ is the normal perturbation. $I_{0}$ is the original image of the $t$th frame, and $\rho (I_{k+1}^{j}-I_{0})$ is the ${l}_{2}$ norm. Please refer to~\cite{jia2021iou} concerning the process of adversarial example generation. Considering that perturbations are measured by the $l_{2}$ norm, we define the adversarial example degree of the $t$th frame as $\Lambda_t=M$. We hope that adversarial example scenarios can reveal the potential threats of adversarial attacks against trackers in many critical real-world applications (\eg intelligent shipping and delivery) and could work as a new way to evaluate the robustness of trackers.

\subsection{Scenario Subtest Sets}
To facilitate the UTUSC protocol, we select one hundred representative video sequences for each scenario subtest set from the test set in Section~\ref{sec:Dataset Splitting}. Importantly, the selected video sequences need to cover all scenario indicators for evaluating the detailed performance of the different algorithms in various scenarios. To this end, we first visualize the distributions of the framewise scenario indicators, including \emph{low light}, \emph{long-term occlusion}, \emph{small targets}, \emph{high-speed motion}, \emph{target distortions}, \emph{dual-dynamic disturbances} and \emph{adversarial examples}, on the test set in Fig.~\ref{fig:scenario_distribution}. We can observe that these scenario indicators present a normal distribution (low light), a long-tailed distribution (long-term occlusion, small targets, high-speed motion, target distortions, and dual-dynamic disturbances), or a uniform distribution (adversarial examples). The wide distributions of different scenario indicators make WebUAV-3M a challenging benchmark for deep UAV tracking under the scenario constraint protocol.

After all the framewise difficulty indicators of the test set are acquired, we calculate the average scenario indicator for each video, divide the videos into several discrete intervals, and sample one hundred videos uniformly from all intervals. As shown in Table~\ref{tab:scenario_test_sets}, each scenario subtest set has similar numbers of superclasses, motion classes, and target classes. The average number of superclasses is eleven. The small targets scenario has only six motion classes, which indicates that the challenging small targets factors are only distributed in a small part of the person superclass (only people labeled with the motion class) in WebUAV-3M. The challenging long-term occlusion factor is distributed in the smallest number of target classes (\ie 39), while high-speed motion is distributed in the broadest number of target classes (\ie 49).

\section{Experiments}

\subsection{Baseline Methods}
To provide extensive baselines for future research, we evaluate 43 representative tracking methods. We mainly consider deep tracking algorithms, comprising correlation filter (CF)-based trackers (\eg CF trackers with deep features (CF2~\cite{ma2015hierarchical}, DeepSRDCF~\cite{danelljan2015convolutional}, CCOT~\cite{danelljan2016beyond}, ECO~\cite{danelljan2017eco}, STRCF~\cite{feng2018cvpr},  LADCF~\cite{XuFWK19tip}, AutoTrack~\cite{Li0DHL20}) and deep discriminative correlation filter (DCF) trackers (ATOM~\cite{DanelljanBKF19}, DiMP~\cite{BhatDGT19iccv}, PrDiMP~\cite{danelljan2020probabilistic}, UTrack~\cite{zhang2020accurate} and KeepTrack~\cite{mayer2021learning})), Siamese network-based trackers (\eg SiamFC~\cite{bertinetto2016fully} and its variants DSiam~\cite{guo2017iccv}, SiamRPN~\cite{li2018cvpr}, DaSiamRPN~\cite{Zhu_2018_ECCV}, SiamMask~\cite{Wang0BHT19}, SiamPRN++~\cite{LiWWZXY19}, SiamDW~\cite{ZhangP19}, UpdateNet~\cite{zhang2019learning}, UDT~\cite{WangS0ZLL19}, SiamFC++~\cite{XuWLYY20}, SiamCAR~\cite{guo2020siamcar}, SiamBAN~\cite{ChenZLZJ20}, Ocean~\cite{zhang2020ocean},
RPT~\cite{ma2020rpt}, SiamGAT~\cite{guo2021graph} and AutoMatch~\cite{zhang2021learn}), and other deep trackers (\eg the multidomain CNN (MDNet)~\cite{nam2016learning}, the GOTURN deep regression network~\cite{held2016learning}, the VITAL adversarial learning tracker~\cite{song2018vital}, the meta-learning tracker (MetaTracker)~\cite{meta2018eccv}, the ACT reinforcement learning tracker~\cite{ChenWLWL18}, the AlphaRefine refinement module~\cite{yan2021alpha}, the LightTrack neural architecture search tracker~\cite{yan2021lighttrack}, and transformer-based trackers~\cite{vaswani2017attention} (\eg TransT~\cite{chen2021transformer}, TrDiMP~\cite{wang2021transformer} and HiFT~\cite{cao2021hift})), since they have demonstrated state-of-the-art performance in recent tracking benchmarks~\cite{wu2015otb,huang2019got,fan2019lasot} and challenges~\cite{zhu2020vision,kristan2016novel}. We also evaluate some traditional (\ie handcrafted feature-based) trackers (\eg KCF~\cite{henriques2014high}, CACF~\cite{mueller2017context}, BACF~\cite{kiani2017learning}, MCCT~\cite{wangnin2018cvpr}, and ARCF~\cite{Huang0LLL19}) for completeness (see Table~\ref{tab:Overall_results}).

\subsection{Implementation Details}
Our evaluation only includes algorithms if their original codes are publicly available. The parameters of each tracker are fixed for all the considered video sequences. Since no parameter adjustment is performed on our benchmark, the evaluation results in this work can be viewed as the lower bound of the tracking performance. We also apply the default network weights provided in the source codes except during the retraining experiment. {\color{black}Some trackers have several different variants (\eg DiMP18 and DiMP50) and optimized codebases (\eg pytracking\footnote{https://github.com/visionml/pytracking}). For fair comparisons, we choose the codes from original papers and select the variants with the highest performance (\eg PrDiMP refers to PrDiMP50 employing ResNet50 as the backbone in Table~\ref{tab:Overall_results}). The experiments are implemented using Python 3.6, PyTorch 1.7/1.2, or MATLAB R2017b with an Intel (R) Xeon (R) Gold 6230R CPU @ 2.10 GHz, three NVIDIA RTX A5000 GPUs and a Dell 64G Memory on an Ubuntu 18.04 server.}

\subsection{Evaluation Metrics}
\label{sec:Evaluation_Metrics}
In this work, we perform a one-pass evaluation (OPE)~\cite{wu2015otb,huang2019got,fan2019lasot} and adopt four popular metrics, \ie the \emph{precision plot} (Pre), \emph{normalized precision plot} (nPre), \emph{success plot} (AUC) and \emph{mean accuracy}~\cite{jiang2021anti} (mAcc) measures, and a newly proposed metric, the \emph{complete success plot} (cAUC), to assess the performance of different tracking algorithms.

The \emph{precision plot} is adopted to measure the percentage of frames whose center location errors are within the pre-defined threshold. The trackers are ranked in terms of this metric by one representative precision score (\eg the score obtained when the threshold=$20$ pixels). Since the \emph{precision plot} is sensitive to the target size and image resolution, the \emph{normalized precision plot}, \ie a plot normalizing each precision score over the size of the ground-truth bounding box, was introduced in~\cite{muller2018trackingnet}. The \emph{success plot} indicates the percentage of frames whose overlap scores are higher than a given threshold. Different trackers are ranked in terms of this metric using the area under the curve (between 0 to 1) of each success plot. The \emph{accuracy} metric was introduced in~\cite{jiang2021anti}; it encourages the trackers to provide invisible predictions of the given target when the target disappears. 

\begin{table*}[t]
	{\small
	\renewcommand\arraystretch{1.0}
	\caption{Overall tracking results of the 43 baseline trackers on WebUAV-3M test set. The trackers are ranked by publication time. The top two results of the performance are indicated in bold and underlined. The ``properties column'' denotes the attributes of different trackers: correlation filter (yes/no), Siamese network (yes/no), deep learning (yes/no), feature representation (Transformer, CNN - convolutional neural networks, HOG - histogram of gradients, CN – color names, CH - color histogram and Gray - grayscale features), and pretraining (yes/no).}
	\label{tab:Overall_results}
	\begin{center}
		\setlength{\tabcolsep}{1.0mm}{
			\begin{tabular}{|l|c|cccccc|ccccc|}
			   \hline
			   \multirow{2}*{\textbf{Tracker}} & \multirow{2}*{\textbf{Publication}} & \multicolumn{6}{c|}{\textbf{Performance}} & \multicolumn{5}{c|}{\textbf{Properties}}\\
			   \cline{3-13} & & \textbf{Pre} & \textbf{nPre} & \textbf{AUC} & \textbf{cAUC} & \textbf{mAcc} & \textbf{FPS} & \textbf{CF} &  \textbf{Siamese} & \textbf{DL} & \textbf{Feature} & \textbf{Pretraining}\\
				
			   \hline
			   \hline	
               01. KCF~\cite{henriques2014high} & TPAMI-2015 & 0.457 & 0.294 & 0.272 & 0.245 & 0.271 & \underline{\color{black}132.9@CPU} & \cmark & \xmark & \xmark & HOG & \xmark \\
               
               02. CF2~\cite{ma2015hierarchical} & ICCV-2015 & 0.538 & 0.356 & 0.343 & 0.311 & 0.343 & 4.3@CPU & \cmark & \xmark & \cmark & CNN &  \xmark \\
               
               03. DeepSRDCF~\cite{danelljan2015convolutional} & ICCVW-2015 & 0.586 & 0.467 & 0.402 & 0.365  & 0.405 & 3.7@CPU & \cmark & \xmark & \cmark & CNN, HOG, CN &  \xmark \\

               04. MDNet~\cite{nam2016learning} & CVPR-2016 & 0.561 & 0.446 & 0.390 & 0.347 & 0.392 & 1.6@GPU & \xmark & \xmark & \cmark & CNN & \cmark\\
               
               05. SiamFC~\cite{bertinetto2016fully} & ECCVW-2016  & 0.534 & 0.393 & 0.351 & 0.317 & 0.352 & 78.2@GPU & \xmark & \cmark & \cmark & CNN & \cmark\\
               
               06. GOTURN~\cite{held2016learning} & ECCV-2016 & 0.375 & 0.238 & 0.215 & 0.178 & 0.212 & 86.9@GPU & \xmark & \xmark & \cmark & CNN & \cmark\\

               07. CCOT~\cite{danelljan2016beyond} & ECCV-2016 & 0.614 & 0.488 & 0.408 & 0.365 & 0.411 & 2.6@GPU & \cmark & \xmark & \cmark & CNN & \xmark\\              
              
               08. ECO~\cite{danelljan2017eco} & CVPR-2017 & {\color{black}0.657} & 0.515 & 0.450 & 0.412 & 0.454 & 5.8@GPU & \cmark & \xmark & \cmark & CNN, HOG, CN &  \xmark \\
               
               09. CACF~\cite{mueller2017context} & CVPR-2017 & 0.601 & 0.479 & 0.412 & 0.375 & 0.414 & 34.8@CPU & \cmark & \xmark & \xmark & HOG, CH & \xmark \\
               
               10. BACF~\cite{kiani2017learning} & ICCV-2017 & 0.612 & 0.497 & 0.422 & 0.382  & 0.426 & 38.5@CPU & \cmark & \xmark & \xmark  & HOG & \xmark\\

               11. DSiam~\cite{guo2017iccv} & ICCV-2017 & 0.614 & 0.489 & 0.437 & 0.401 & 0.440 & 25.3@CPU & \xmark & \cmark & \cmark & CNN & \cmark \\
               
               12. VITAL~\cite{song2018vital} & CVPR-2018 & 0.519  & 0.417 & 0.361  & 0.323 & 0.364 & 3.2@GPU & \xmark & \xmark & \cmark & CNN & \cmark\\

               13. SiamRPN~\cite{li2018cvpr} & CVPR-2018 & 0.519 & 0.367 & 0.345 &  0.317 & 0.346 & \textbf{\color{black}142.8@GPU} & \xmark & \cmark & \cmark & CNN & \cmark\\
               
               14. STRCF~\cite{feng2018cvpr} & CVPR-2018 & 0.631 & 0.510 & 0.447 &  0.404 & 0.451 & 8.3@GPU & \cmark & \xmark & \cmark & CNN, HOG, CN & \xmark\\
               
               15. MCCT~\cite{wangnin2018cvpr} & CVPR-2018 & 0.623 & 0.498 & 0.440  & 0.404 & 0.444 & 44.1@CPU & \cmark & \xmark  & \xmark & HOG, CN & \xmark \\
               
               16. DaSiamRPN~\cite{Zhu_2018_ECCV} & ECCV-2018 & 0.489 & 0.347 & 0.310 & 0.289 & 0.310 & {\color{black}98.4@GPU} & \xmark & \cmark & \cmark & CNN & \cmark\\
               
               17. MetaTracker~\cite{meta2018eccv} & ECCV-2018 & 0.605 & 0.404 &  0.376 & 0.336 & 0.376 & 3.6@GPU & \xmark & \xmark & \cmark & CNN & \cmark\\
               
               18. ACT~\cite{ChenWLWL18} & ECCV-2018 & 0.518 & 0.388 & 0.343 & 0.305 & 0.344  & 28.7@GPU & \xmark & \xmark & \cmark & CNN & \cmark\\

               19. SiamMask~\cite{Wang0BHT19} & CVPR-2019 & 0.615 & 0.497 & 0.435  & 0.396 & 0.438 & 50.6@GPU & \xmark & \cmark & \cmark & CNN & \cmark\\

               20. SiamPRN++~\cite{LiWWZXY19} & CVPR-2019 & 0.607 & 0.482 & 0.433 & 0.401 & 0.437 & 32.5@GPU & \xmark & \cmark & \cmark & CNN & \cmark\\
               
               21. ATOM~\cite{DanelljanBKF19} & CVPR-2019 & {\color{black}0.439} & {\color{black}0.351} & {\color{black}0.291} & {\color{black}0.261} & {\color{black}0.292} & {\color{black}28.6@GPU} & \cmark & \xmark & \cmark & CNN & \cmark\\
               
               22. SiamDW~\cite{ZhangP19} & CVPR-2019 & 0.559 & 0.447 & 0.373 & 0.327 & 0.374 & 39.5@GPU & \xmark & \cmark & \cmark & CNN & \cmark\\

               23. UDT~\cite{WangS0ZLL19} & CVPR-2019 & 0.572 & 0.465 & 0.416 & 0.379 & 0.419 & 68.2@GPU & \xmark & \cmark & \cmark & CNN & \cmark\\
               
               24. ARCF~\cite{Huang0LLL19} & ICCV-2019 & 0.613 & 0.490 & 0.405 & 0.363 & 0.407 & 57.6@CPU & \cmark & \xmark & \xmark & HOG, CN, Gray & \xmark\\
               
               25. LADCF~\cite{XuFWK19tip} & TIP-2019 & 0.470 & 0.375 & 0.339 & 0.307  & 0.342 & 8.6@GPU & \cmark & \xmark & \cmark & CNN, HOG, CN & \xmark\\ 
               
               26. UpdateNet~\cite{zhang2019learning} & ICCV-2019 & 0.536 & 0.380 & 0.360 & 0.330 & 0.360 & 76.3@GPU & \xmark & \cmark & \cmark & CNN & \cmark\\
               
               27. DiMP~\cite{BhatDGT19iccv} & ICCV-2019  & {\color{black}0.544} & {\color{black}0.441} & {\color{black}0.364} & {\color{black}0.336} & {\color{black}0.367} & {\color{black}37.6@GPU} & \cmark & \xmark & \cmark & CNN & \cmark\\

               28. SiamFC++~\cite{XuWLYY20} & AAAI-2020 & 0.554 & 0.431 & 0.388 & 0.352 & 0.391 & 86.2@GPU & \xmark & \cmark & \cmark & CNN & \cmark\\  
               
               29. SiamCAR~\cite{guo2020siamcar} & CVPR-2020 & 0.642 & 0.512 & 0.412 & 0.378 & 0.415 & 46.7@GPU & \xmark & \cmark & \cmark & CNN & \cmark\\
               
               30. PrDiMP~\cite{danelljan2020probabilistic} & CVPR-2020 & {\color{black}0.674} & {\color{black}0.575} & {\color{black}0.514}  & {\color{black}0.484} & {\color{black}0.521} & {\color{black}32.4@GPU} & \cmark & \xmark & \cmark & CNN & \cmark\\   
               
               31. AutoTrack~\cite{Li0DHL20} & CVPR-2020 & 0.618 & 0.495 & 0.446 & 0.412 & 0.450 & 55.8@CPU & \cmark & \xmark & \xmark & HOG, CN, Gray & \xmark\\  
               
               32. SiamBAN~\cite{ChenZLZJ20} & CVPR-2020 & 0.615 & 0.498 & 0.438 & 0.401 & 0.442 & 37.1@GPU & \xmark & \cmark & \cmark & CNN & \cmark\\  
               
               33. UTrack~\cite{zhang2020accurate} & ACM MM-2020 & 0.540 & 0.438 & 0.362 & 0.333  & 0.364 & 21.6@GPU & \cmark & \xmark & \cmark & CNN & \cmark\\ 
               
               34. Ocean~\cite{zhang2020ocean} & ECCV-2020 & 0.505 & 0.409 & 0.369 & 0.325 & 0.366 & 65.3@GPU & \xmark & \cmark & \cmark & CNN & \cmark\\
               
               35. RPT~\cite{ma2020rpt} & ECCVW-2020 & 0.660 & 0.546 & {\color{black}0.495} & {\color{black}0.462} & {\color{black}0.501} & 33.2@GPU & \xmark & \cmark & \cmark & CNN & \cmark\\  
               
               36. SiamGAT~\cite{guo2021graph} & CVPR-2021 & 0.573 & 0.474 & 0.393 & 0.349 & 0.395 & {\color{black}88.5@GPU} & \xmark & \cmark & \cmark & CNN & \cmark\\ 
               
               37. LightTrack~\cite{yan2021lighttrack} & CVPR-2021 & 0.602 & 0.494 & 0.458 & 0.415 & 0.459 & 64.9@GPU & \xmark & \xmark & \cmark & CNN & \cmark\\ 
               
               38. TrDiMP~\cite{wang2021transformer} & CVPR-2021 & 0.600 & 0.476 & 0.399 & 0.370 & 0.402 & 23.0@GPU & \cmark & \xmark & \cmark & {\color{black}CNN, Transformer} & \cmark\\

               39. TransT~\cite{chen2021transformer} & CVPR-2021 & {\color{black}0.618} & {\color{black}0.509} & 0.448 & {\color{black}0.422} & 0.453 & 53.5@GPU & \xmark & \cmark & \cmark & {\color{black}CNN, Transformer} & \cmark\\           

               40. AlphaRefine~\cite{yan2021alpha} & CVPR-2021 & \textbf{\color{black}0.753} & \textbf{\color{black}0.643} & \textbf{\color{black}0.593} & \textbf{\color{black}0.562} & \textbf{\color{black}0.602} & 42.3@GPU & \cmark & \xmark & \cmark & CNN & \cmark\\               
             
               41. HiFT~\cite{cao2021hift} & ICCV-2021 & 0.515 & 0.413 & 0.360 & 0.314 & 0.358 & {\color{black}122.6@GPU} & \xmark & \cmark & \cmark & {\color{black}CNN, Transformer} & \cmark\\
               
               42. AutoMatch~\cite{zhang2021learn} & ICCV-2021 & 0.618 & 0.496 & 0.454 & 0.416 & 0.458 & 63.1@GPU & \xmark & \cmark & \cmark & CNN & \cmark\\  
               
               43. KeepTrack~\cite{mayer2021learning} & ICCV-2021 & \underline{\color{black}0.710} & \underline{\color{black}0.603} & \underline{\color{black}0.543} & \underline{\color{black}0.512} & \underline{\color{black}0.550} & 33.9@GPU & \cmark & \xmark & \cmark & CNN & \cmark\\ 
				\hline	
			\end{tabular}
		}
	\end{center}
    }
\end{table*}

A good metric for predicted box evaluation should consider three critical geometric factors, \ie the central point distance, overlap area, and aspect ratio~\cite{zheng2020distance,wu2015otb}. However, the above four metrics only measure the central point distance or overlap area and do not reflect the aspect ratio of the target object. To that end, we propose the \emph{complete success plot} evaluation metric. {\color{black}Based on the overlap score, we first introduce the complete overlap score $S_c$ by imposing the normalized distance and the consistency of the aspect ratio, $S_{c}\!=\!\frac{|B_G \cap B_P|}{|B_G \cup B_P|}\!-\!\frac{d^{2}(\bm{b}^{G},\bm{b}^{P})}{c^{2}}\!-\!\alpha v$, where $\bm{b}^{G}$ and $\bm{b}^{P}$ are the central points of $B^{G}$ and $B^{P}$, respectively, $d(\cdot)$ is the Euclidean distance, and $c$ is the diagonal length of the smallest closed box that covers the ground-truth bounding box and the predicted box.} $\alpha>0$ is a balance parameter, and $v=\frac{4}{\pi^{2}}(arctan\frac{w^{G}}{h^{G}}-arctan\frac{w^{P}}{h^{P}})^{2}$ measures the consistency of the aspect ratio, as in~\cite{zheng2020distance}. Similar to the \emph{success plot}, the \emph{complete success plot} is defined as the percentage of frames where the complete overlap score $S_c$ is higher than a given threshold. {\color{black}Please refer to our evaluation toolkits for more technical details.}

\begin{figure*}[t]
    \centering
    \vspace{-0.4cm}
    \subfloat[Low light]{\includegraphics[width =0.66\columnwidth]{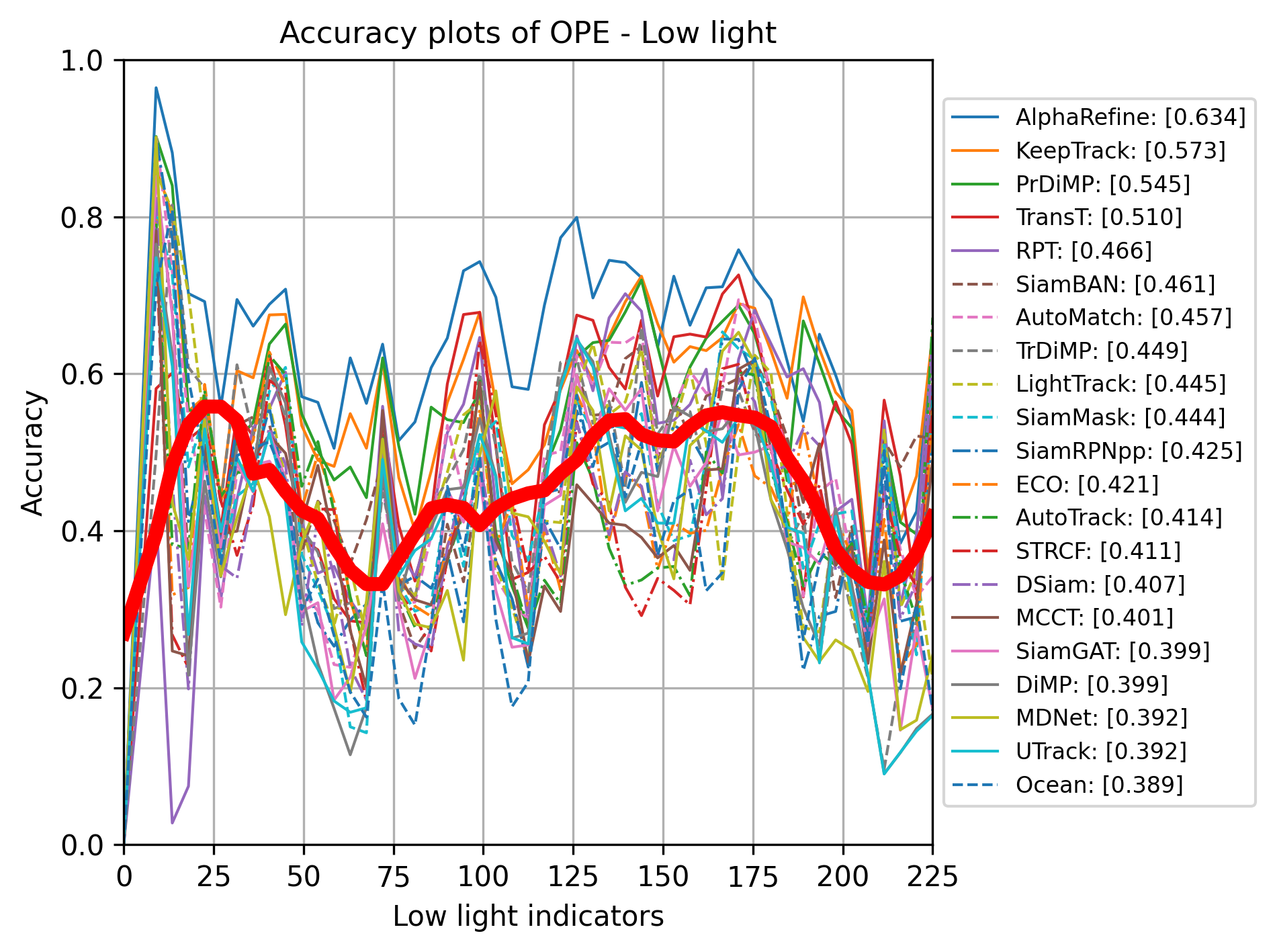}}
    ~
    \subfloat[Long-term occlusion]{\includegraphics[width =0.66\columnwidth]{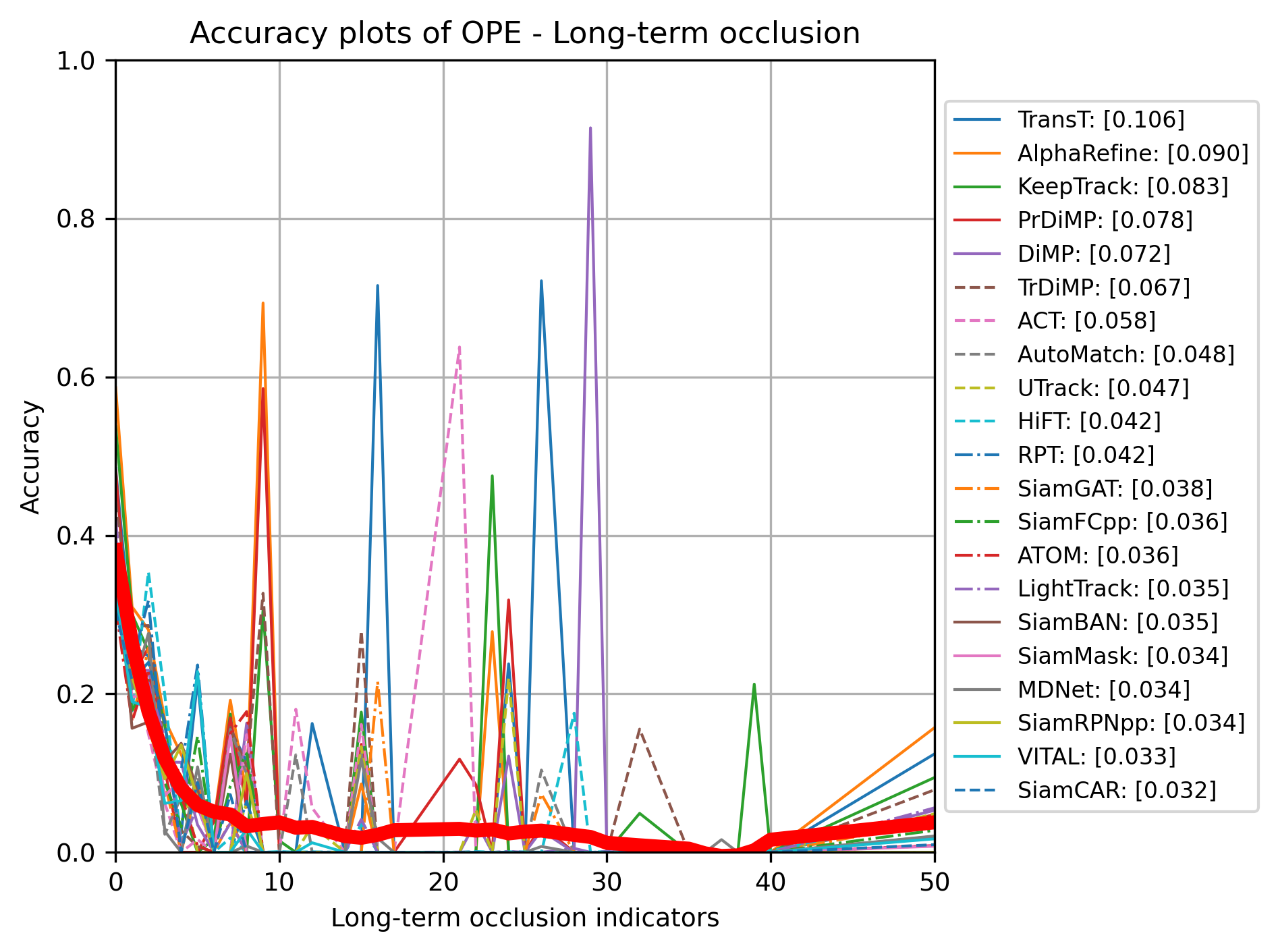}}
    ~
    \subfloat[Small targets]{\includegraphics[width =0.66\columnwidth]{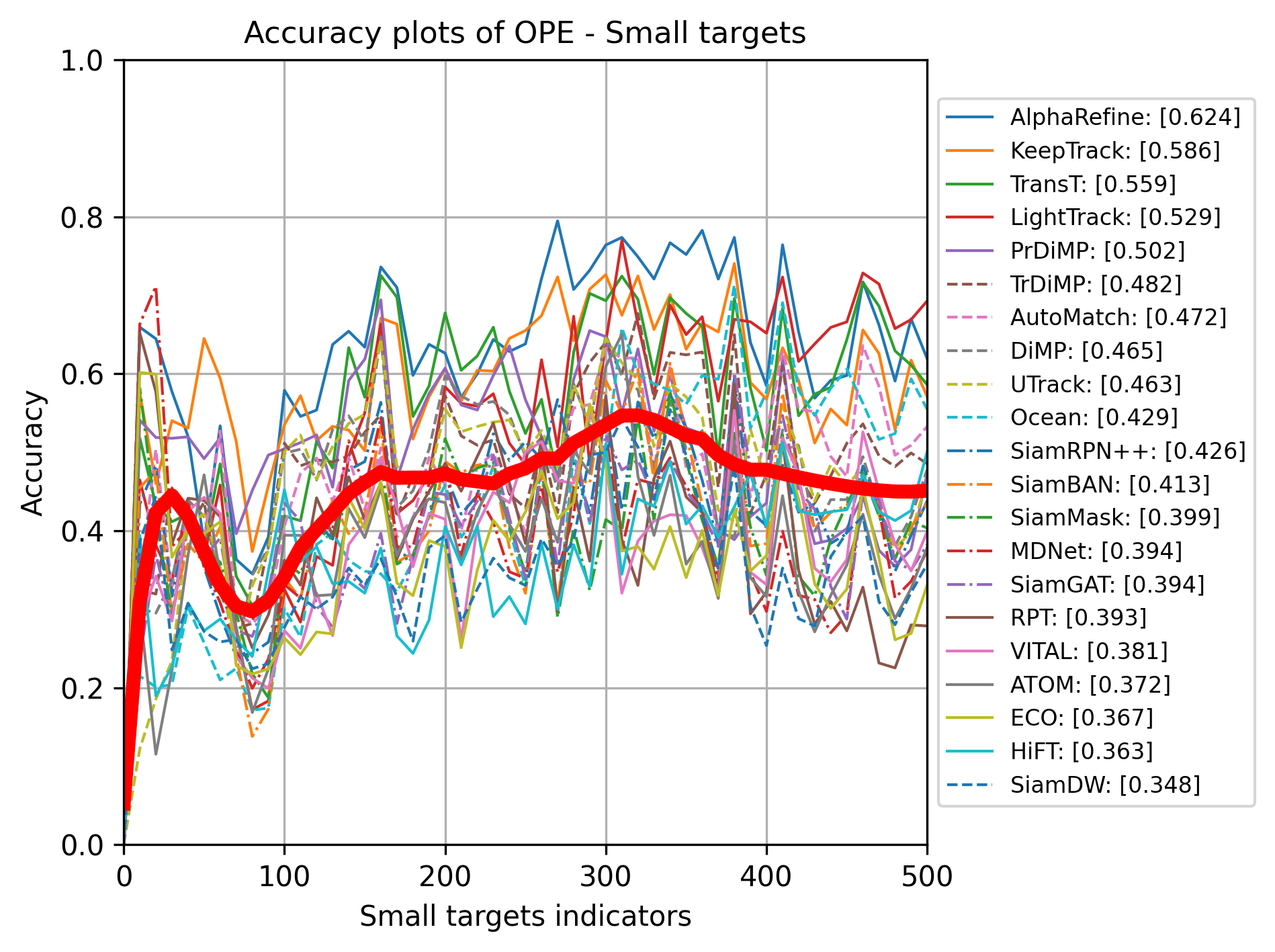}} \\
    
    \subfloat[High-speed motion]{\includegraphics[width =0.66\columnwidth]{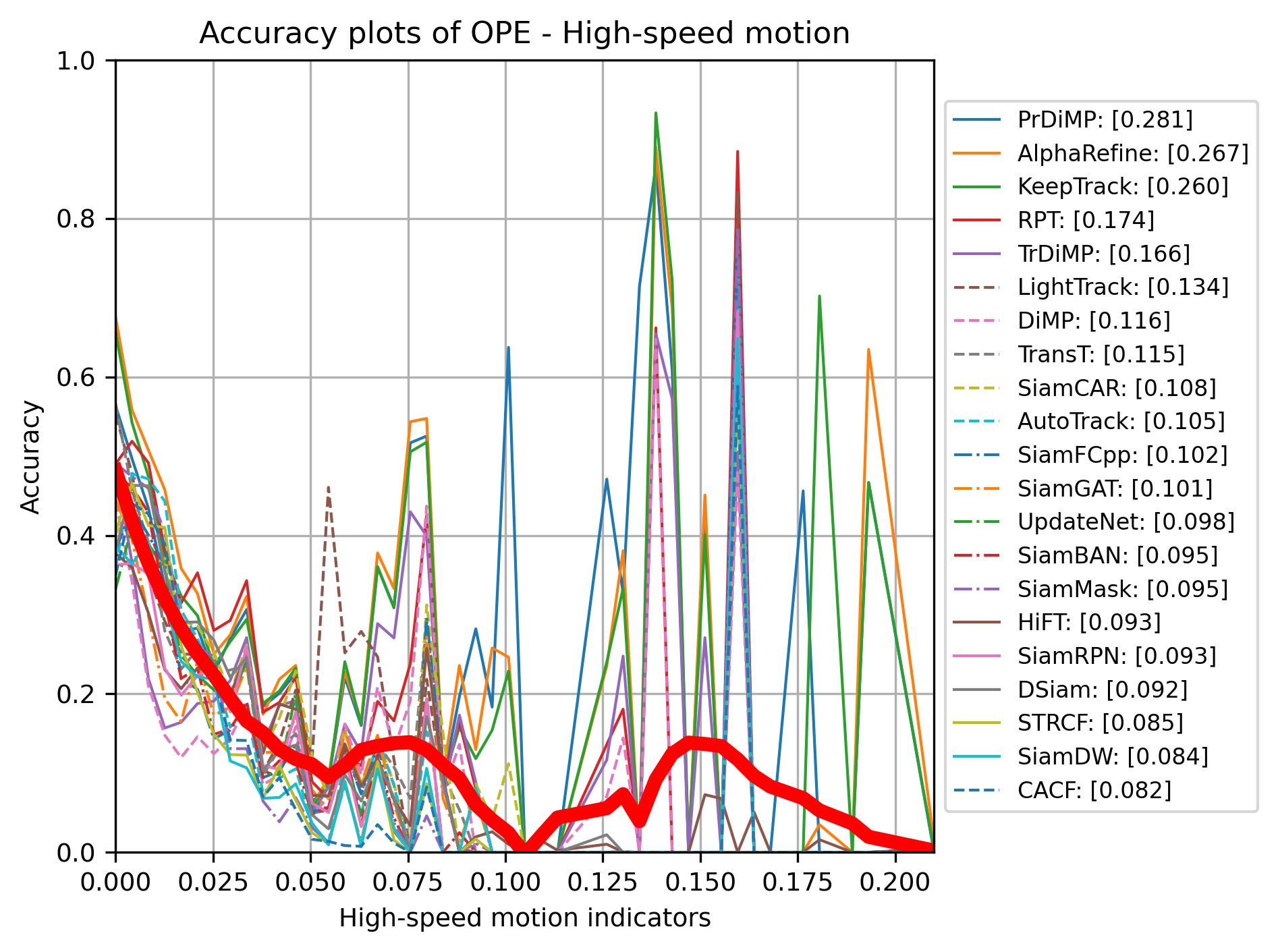}}
    ~
    \subfloat[Target distortions]{\includegraphics[width =0.66\columnwidth]{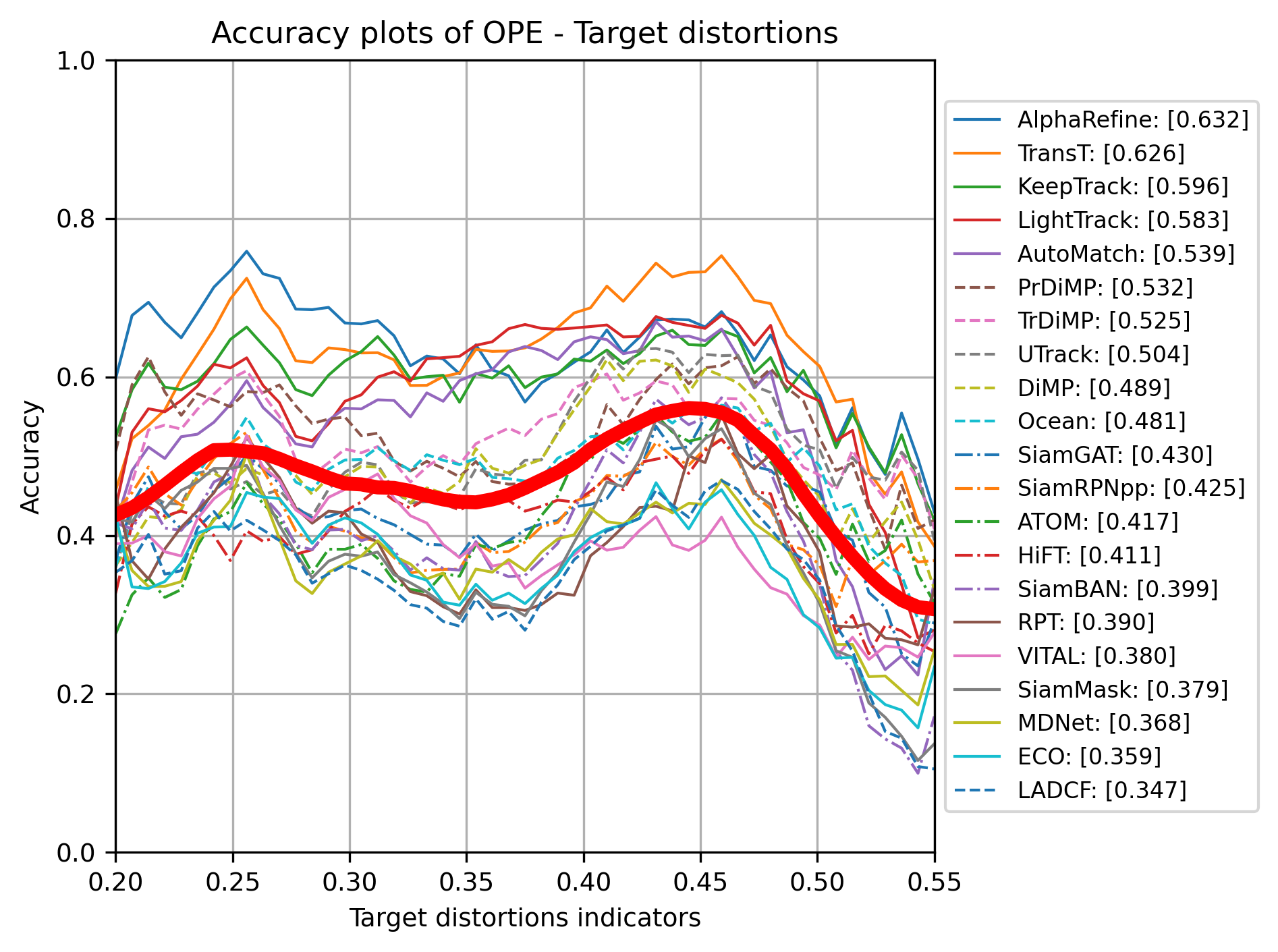}}
    ~
    \subfloat[Dual-dynamic disturbances]{\includegraphics[width =0.66\columnwidth]{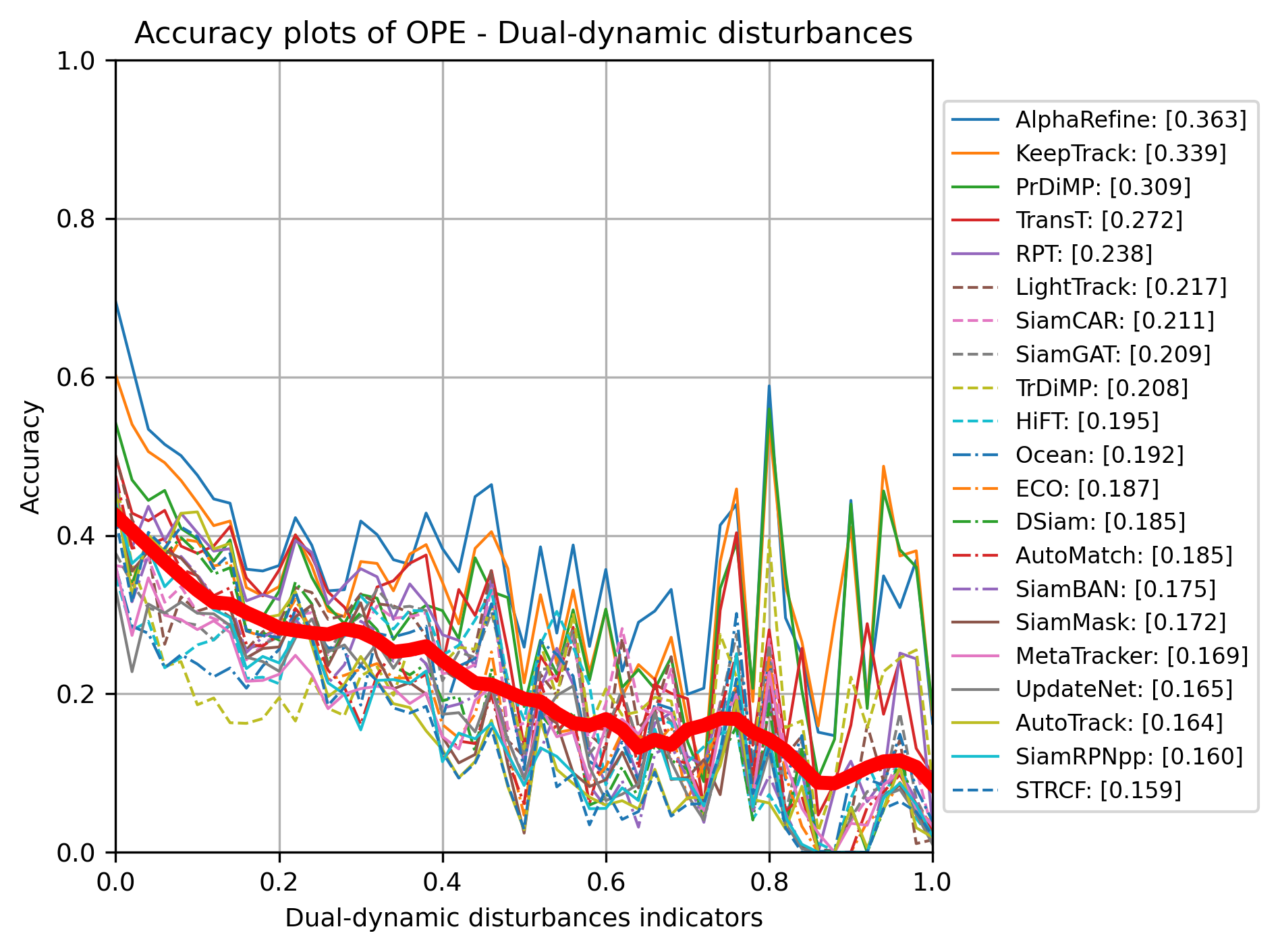}}
\caption{Evaluation results obtained on the six scenario subtest sets of WebUAV-3M under the UTUSC protocol using the mAcc score. For clarity, only the top 21 trackers are shown. The thick red curves represent the average performance across all baseline trackers.}
\label{fig:scenario_evaluation}
\end{figure*}

\begin{figure}[t]   
	\centering\centerline{\includegraphics[width=1.0\linewidth]{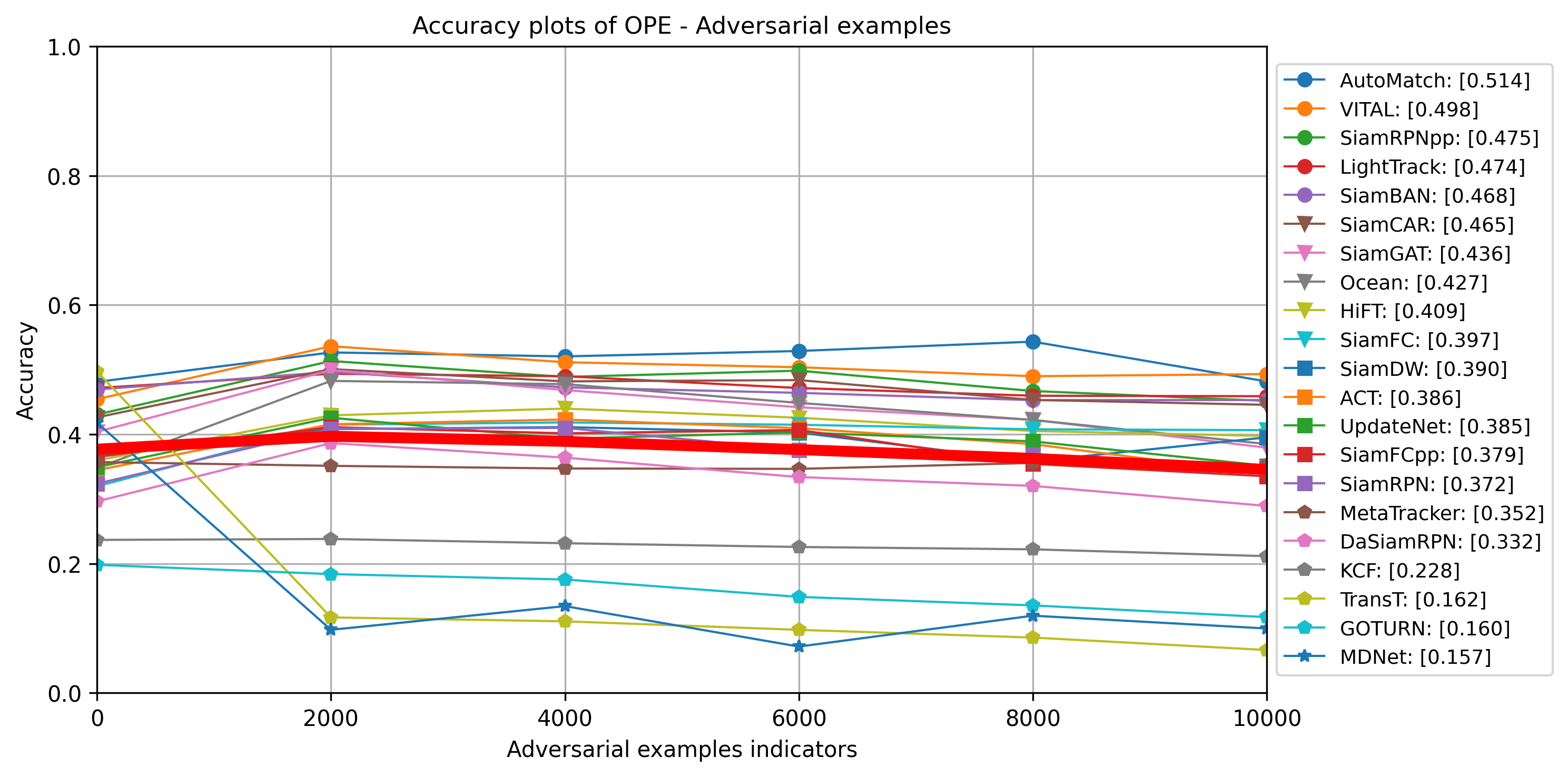}}
	\caption{Evaluation results obtained on the adversarial example subtest set of WebUAV-3M using the mAcc score.}
	\label{fig:accuracy_plots_Adversarial_examples}
\end{figure}

\subsection{Overall Performance}
We report the overall performance achieved by 43 baseline trackers on the WebUAV-3M test set. The evaluation results are summarized in Table~\ref{tab:Overall_results}. The top three trackers are AlphaRefine, KeepTrack, and PrDiMP. All of these trackers are built upon deep CNN features with end-to-end feature learning. KeepTrack and PrDiMP are deep DCF trackers, while AlphaRefine adopts a flexible and accurate refinement module to strengthen DiMP. These advanced trackers show the superiority of the end-to-end learnable architectures in the deep DCF paradigm~\cite{yan2021alpha,mayer2021learning,jiang2018acquisition}.

Specifically, AlphaRefine achieves the best performance, and it outperforms the second-placed tracker by $4.3\%$, $4.0\%$, $5.0\%$, $5.0\%$ and $5.2\%$ in terms of the Pre, nPre, AUC, cAUC and mAcc scores, respectively. Compared with other trackers, KeepTrack and AlphaRefine provide leading Pre scores, \ie $71.0\%$ and $75.3\%$, while the Pre scores of other trackers are lower than $70.0\%$. The next three trackers are PrDiMP, RPT, and ECO. PrDiMP is a deep DCF-based method, RPT is a Siamese network-based method with point set representation learning, and ECO is a CF-based method with a deep and handcrafted features fusion mechanism. Regarding the performance of the pioneering deep tracking frameworks, \ie TransT, TrDiMP, and HiFT, we can observe that these trackers achieve compelling performance compared with that of state-of-the-art transformerless deep trackers, which are expected to have a substantial impact on the tracking community in the coming years. Among the traditional methods using handcrafted features, MCCT, AutoTrack, ARCF, BACF, and CACF obtain the top five evaluation results. {\color{black}By using only handcrafted features, MCCT and AutoTrack achieve results that are comparable to or even better than those of some deep trackers, such as STRCF, CCOT, SiamMask, SiamBAN, LightTrack, and VITAL. Interestingly, we find that some classic trackers (\eg CF2, and ECO) are even better than recent methods (\eg ATOM, and Ocean), showing that the formers have excellent generalization ability without bells and whistles.}

Considering the requirements of many real-time applications, we also report the tracking speeds of different methods using FPS, as shown in Table~\ref{tab:Overall_results}. Among all the baseline trackers, the three fastest methods, \ie SiamRPN, KCF, and HiFT, exceed 100 FPS. SiamRPN benefits from its straightforward feature extraction network (\ie AlexNet~\cite{krizhevsky2012imagenet}) and efficient RPN subnetwork~\cite{ren2015faster}, while HiFT possesses a lightweight encoder-decoder architecture that achieves high-speed deep tracking. Although only a CPU is used, KCF achieves the second-highest speed of 132.0 FPS due to the fast Fourier transform (FFT) employed when learning and applying the correlation filter.

\subsection{Attribute-Based Performance}
For completeness, we also report the global attribute-based evaluation results to analyze the performance of the 43 baseline trackers for each challenge factor. The evaluation results demonstrate that the most challenging attributes include FO, OV, FM, DEF, IV, COM-H, SIZ-S, and LEN-L. The performance rankings of different algorithms on each attribute are similar to the overall rankings on the WebUAV-3M test set, which shows that WebUAV-3M provides reliable evaluations. More detailed descriptions can be found in the~\textbf{supplemental material}.

\subsection{Baselines Under the UTUSC Protocol}
Although the overall performance evaluation and global attribute-based evaluation bring to light the general quality of trackers and the performance characteristics of different algorithms when facing various attributes, respectively, they cannot differentiate trackers according to continuous, objective, and framewise difficulty indicators and thus cannot deeply reveal the weaknesses and strengths of different algorithms. To this end, the rigorous and dedicated UTUSC protocol is introduced for comprehensively evaluating deep UAV tracking algorithms, as described in Section~\ref{sec:UTUSC}. The evaluation results obtained by 43 baseline trackers on seven scenario subtest sets in WebUAV-3M are shown in Fig.~\ref{fig:scenario_evaluation} and Fig.~\ref{fig:accuracy_plots_Adversarial_examples}. We present a detailed analysis of the baseline trackers under the UTUSC protocol as follows.

\myPara{Low light.} We only consider frames with mean image intensities in [0, 225] for the low light evaluation scenario. Unsurprisingly, all baseline trackers exhibit sharp performance drops when the low-light indicator falls (\ie below 25), suggesting the difficulty of low-light conditions for tracking algorithms designed primarily for high-visibility inputs. We also find that the performance of the tracking algorithms fluctuates wildly when the low light indicator is between 25 and 225.

\myPara{Long-term occlusion.} The tracking results of subsequent frames can be used to measure an algorithm’s ability to resist long-term occlusion when occlusion occurs. To ensure the accuracy of a long-term occlusion indicator, we consider the next K (\eg K=5) frames after the current frame. We observe that the trackers show consistent performance decreases when the long-term occlusion indicators become larger, indicating that occlusion is still challenging for current deep trackers. Among all baseline trackers, the tracker that is least affected by long-term occlusion is TransT, which introduces transformer-style ego-context augmentation and cross-feature augmentation modules to establish dependence between long-distance features and aggregate global information ~\cite{chen2021transformer}.

\myPara{Small targets.}
We consider targets with sizes in [0, 500] for the evaluation of the small target. The mAcc scores do not change much when targets’ sizes are relatively large (\ie above 150) but drop rapidly when their sizes become very low (\ie below 30). At 320, the average performance across all baseline trackers peaks.

\myPara{High-speed motion.} 
All baseline trackers exhibit degraded performance when the target moves more rapidly. Some trackers (\eg PrDiMP, AlphaRefine, KeepTrack, TrDiMP, \etc.) show multiple peaks at relatively large high-speed motion indications (\ie above 0.05), indicating that they can track some fast targets. PrDiMP ranks 1st in the high-speed motion scenario by using a probabilistic regression formulation to estimate the uncertainty of the target state.

\myPara{Target distortions.} Low-quality images present larger target distortion indicators, as in~\cite{zhu2020metaiqa}, indicating severe distortion of target appearances (\eg brightening, white noise, and motion blur). Most trackers seem to be well adapted to the high-quality images in the dataset but drop rapidly when the image quality deteriorates (\ie when the indicator exceeds 0.45).

\myPara{Dual-dynamic disturbances.} Dual-dynamic disturbances affect all tracking algorithms, which exhibit significant performance drops when the indicators increase from 0 to 1.0. AlphaRefine ranks 1st in terms of the mAcc score in this challenging scenario; this is consistent with the results obtained on the entire test set.

\myPara{Adversarial examples.} We consider the degree of adversarial examples in [0, 2000, 4000, 6000, 8000, 10000]. ``0'' indicates no image perturbation. As the degree increases, the adversarial noise becomes more severe. From Fig.~\ref{fig:accuracy_plots_Adversarial_examples}, we find that the tracking algorithms do not consistently degrade as much as one would expect as the number of iterations increases. {\color{black}Most of the tracking algorithms present only slight performance degradations, indicating the limitations of general adversarial examples.} However, adding adversarial examples significantly degrades the performance of the state-of-the-art transformer-based tracker (TransT) and CNN-based tracker (MDNet). This reveals that both the transformer-based and CNN-based trackers are vulnerable to adversarial attacks. Therefore, the design of more powerful and robust deep UAV tracking algorithms is still an open problem.

\subsection{Data Quality Validation}
\label{sec:data_quality_validation}

\myPara{Retraining experiment on WebUAV-3M.}
We first retrain five deep trackers on the WebUAV-3M training set, including ATOM, GOTURN, MDNet, SiamFC, and SiamRPN, to verify the quality of the data annotations. We apply the same network architectures and hyperparameters as those used by the original authors throughout our experiments. The results obtained on the WebUAV-3M test set by our retraining models and the deep models trained on tracking datasets (\ie ALOV++~\cite{smeulders2013visual}, LaSOT~\cite{fan2019lasot}, GOT-10k~\cite{huang2019got}, TrackingNet~\cite{muller2018trackingnet}), video object detection datasets (\ie ImageNet VID~\cite{russakovsky2015imagenet}, YouTube-BB~\cite{real2017youtube}), and object detection datasets (\ie ImageNet DET~\cite{russakovsky2015imagenet}, COCO~\cite{lin2014microsoft}), are summarized in Table~\ref{tab:retraining_results}. From Table~\ref{tab:retraining_results}, we can observe that consistent performance gains are achieved by five retraining-based deep trackers (\ie SiamFC, SiamRPN, ATOM, GOTURN, and MDNet) in terms of their Pre, nPre, AUC, cAUC, and mAcc scores. These results demonstrate that our WebUAV-3M dataset can provide a high-quality platform for training deep trackers and assessing most existing deep UAV tracking methods. 

\begin{table}[t]
	{
	\scriptsize
	\renewcommand\arraystretch{1.0}
	\caption{The results obtained after retraining five deep trackers on WebUAV-3M.}
	\label{tab:retraining_results}
	\begin{center}
		\setlength{\tabcolsep}{0.80mm}{
		\begin{tabular}{|l|c|c|c|c|c|c|}
				\hline
				\textbf{Tracker} & \textbf{Training data} & \textbf{Pre} & \textbf{nPre} & \textbf{AUC} & \textbf{cAUC} & \textbf{mAcc} \\
				\hline
			    \hline
			    GOTURN & ALOV++, ImageNet DET & {\color{black}0.375}  & {\color{black}0.238} & {\color{black}0.215}  & {\color{black}0.178} & {\color{black}0.212}\\
				
				MDNet & ImageNet-VID & {\color{black}0.561} & {\color{black}0.446} & {\color{black}0.390} & {\color{black}0.347} & {\color{black}0.392}\\
				
                ATOM & LaSOT, GOT-10k, TrackingNet, COCO  & {\color{black}0.439} & {\color{black}0.351} & {\color{black}0.291}  & {\color{black}0.261} & {\color{black}0.292}\\
				
				SiamFC  & ImageNet VID & {\color{black}0.534}  & {\color{black}0.393} & {\color{black}0.351}  & {\color{black}0.317} & {\color{black}0.352}\\
				
				SiamRPN & ImageNet VID, YouTube-BB & {\color{black}0.519}  & {\color{black}0.367} & {\color{black}0.345}  & {\color{black}0.317} & {\color{black}0.346}\\
				\hline
				
				GOTURN & \multirow{5}*{WebUAV-3M train set}   & 0.499  & 0.355 & 0.318  & 0.280 & 0.318\\
				
				MDNet &   & 0.603 & 0.506 & 0.445 & 0.409 & 0.450\\
		
				ATOM &  & 0.580 & 0.435 & 0.390  & 0.344 & 0.387\\
		
				SiamFC  &  & 0.629  & 0.521 & 0.454  & 0.417 & 0.458\\

				SiamRPN &   & 0.617  & 0.514 & 0.453  & 0.415 & 0.457\\
				\hline
			\end{tabular}
		}
	\end{center}
    }
\end{table}

\begin{figure}[t]   
	\centering\centerline{\includegraphics[width=1.0\linewidth]{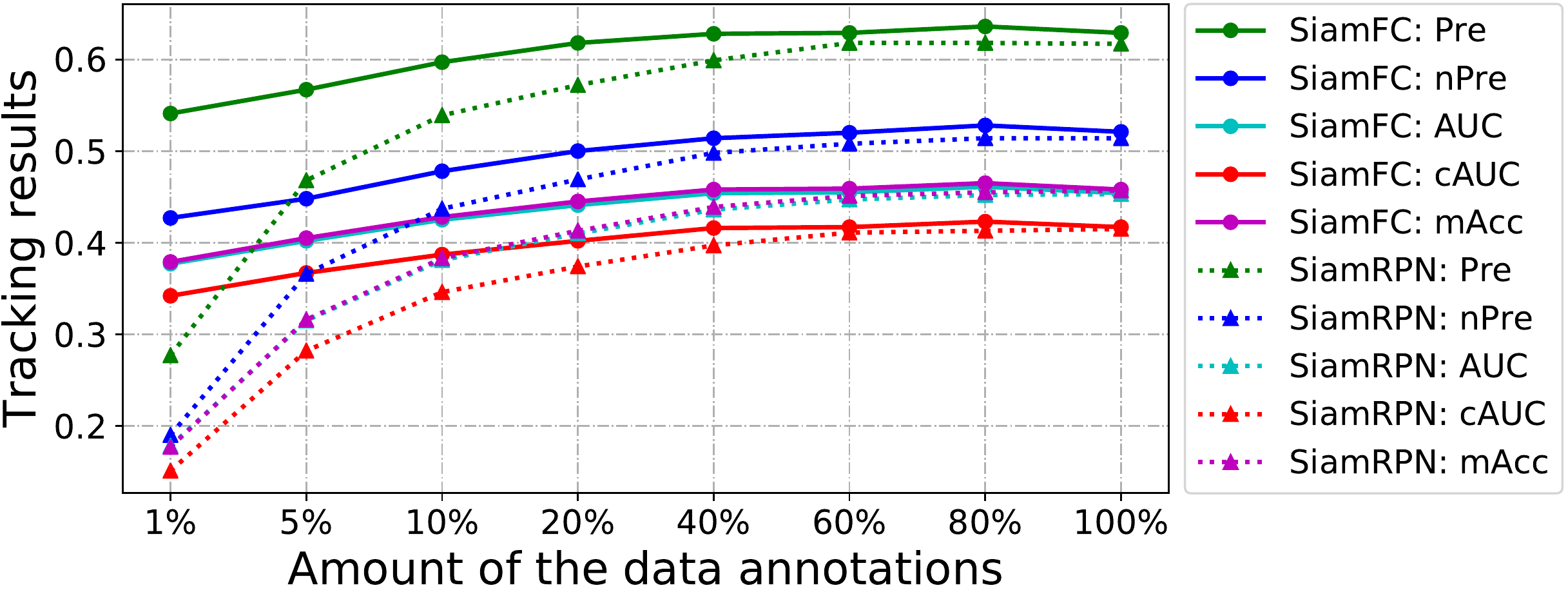}}
	\caption{Impacts of different numbers of data annotations.}
	\label{fig:Impact_of_data_annotations}
\end{figure}

\begin{table}[t]
	{
	\scriptsize
	\renewcommand\arraystretch{1.0}
	\caption{Intraclass domain generalization results (cAUC) of SiamFC on \emph{vehicle} superclass. Sedan, SUV, minivan, hatchback, and box truck are the five largest target classes in \emph{vehicle} superclass. Sedan, SUV, minivan, hatchback, and box truck are the five largest target classes in \emph{vehicle}. The top two results are in bold and underlined.}
	\label{tab:Intra_class_vehicle}
	\begin{center}
		\setlength{\tabcolsep}{2.1mm}{
		\begin{tabular}{|l|c|c|c|c|c|c|}
				\hline
				\diagbox [width=5em,trim=l] {\textbf{Test}}{\textbf{Train}} & \textbf{box truck} & \textbf{hatchback} & \textbf{minivan}  & \textbf{SUV} & \textbf{sedan} & \textbf{Average}\\
				\hline
			    \hline
			    
			   \textbf{box truck} &  0.507   & 0.552 & 0.558   & \underline{0.570}  & \textbf{0.573} & 0.552\\
				
		       \textbf{hatchback} &  0.439   & 0.486 & \textbf{0.528}   & 0.518  & \underline{0.526} & 0.499\\
				
               \textbf{minivan} &  0.483   & 0.493 & 0.481   & \textbf{0.509}  & \underline{0.496} & 0.492\\
				
			   \textbf{SUV} &  0.484   & \underline{0.535} & 0.524   & 0.511  & \textbf{0.546}  & 0.520\\
			
			   \textbf{sedan} &  0.385   & 0.434 & 0.419   & \underline{0.462}  & \textbf{0.469} & 0.434\\
				\hline

			\end{tabular}
		}
	\end{center}
    }
\end{table}

\begin{table}[t]
	{
	\scriptsize
	\renewcommand\arraystretch{1.0}
	\caption{Intraclass domain generalization results (cAUC) of SiamFC on the \emph{person} superclass. Walking, biking, standing, riding an electric bicycle (\ie riding) and sitting are the five largest motion classes in \emph{person}.}
	\label{tab:Intra_class_person}
	\begin{center}
		\setlength{\tabcolsep}{2.4mm}{
		\begin{tabular}{|l|c|c|c|c|c|c|}
				\hline
				\diagbox [width=5em,trim=l] {\textbf{Test}}{\textbf{Train}} & \textbf{sitting} & \textbf{riding} & \textbf{standing}  & \textbf{biking} & \textbf{walking}  & \textbf{Average}\\
				\hline
			    \hline
			   \textbf{sitting} &  0.290  & 0.268 & 0.276   & \underline{0.322}  & \textbf{0.376} & 0.306 \\
		       \textbf{riding} & 0.479   & \underline{0.601} & 0.536   & \textbf{0.606}  & 0.592  & 0.563\\
				
               \textbf{standing} & 0.348   & 0.268 & 0.320   & \underline{0.362}  & \textbf{0.401} & 0.340 \\
				
			   \textbf{biking} & 0.217   & 0.212 & 0.257   & \textbf{0.345}  & \underline{0.316}  & 0.269\\
			
			   \textbf{walking} &  0.307   & 0.305 & 0.322   & \underline{0.364}  & \textbf{0.393}  & 0.338\\
				\hline

			\end{tabular}
		}
	\end{center}
    }
\end{table}

\myPara{Impact of the number of data annotations.}
After verifying the high quality of our data annotations, we further explore the impact of different numbers of data annotations on two widely used deep models (\ie SiamFC and SiamRPN). We randomly select 1$\%$, 5$\%$, 10$\%$, 20$\%$, 40$\%$, 60$\%$, 80$\%$ and 100$\%$ of the videos from the WebUAV-3M training set for training and use the WebUAV-3M test set for testing. The results are presented in Fig.~\ref{fig:Impact_of_data_annotations}. We find some significant differences regarding the impacts of data annotations on different deep models. First, when using a small number of data annotations (\ie 1$\%$, or 35 videos), SiamFC achieves good results, but SiamRPN achieves poor performance. This is because SiamFC is a vanilla Siamese network that only uses the compact AlexNet~\cite{krizhevsky2012imagenet}, while SiamRPN adopts a more complex region proposal network~\cite{ren2015faster}. Second, the performance of SiamRPN significantly improves with the use of larger amounts of data (from 1$\%$ to 10$\%$, or 35 videos to 350 videos), while SiamFC presents a moderately increasing performance trend. Third, when more training data are used, the performance of SiamFC converges or even slightly degrades when all training data are used, indicating overfitting. In contrast, SiamRPN gradually converges as more data are used, and the best performance is obtained when all data are used. Nevertheless, both SiamFC and SiamRPN imply that using larger amounts of training data and more diverse data benefits the model in learning better feature representations.

\begin{table*}[t]
	{
	\scriptsize
	\renewcommand\arraystretch{1.0}
	\caption{Interclass domain generalization results (cAUC) of SiamFC on the two largest subclasses of each of the six largest superclasses, \ie \emph{person}, \emph{vehicle}, \emph{vessel}, \emph{building}, \emph{public transport} and \emph{animal}. The top two results are in bold and underlined.}
	\label{tab:Inter_class}
	\begin{center}
		\setlength{\tabcolsep}{1.415mm}{
		\begin{tabular}{|ll|cc|cc|cc|cc|cc|cc|}
				\hline
				\multicolumn{2}{|c|}{\multirow{2}*{\diagbox [width=16em,trim=l] {\textbf{Test}}{\textbf{Train}}}} & \multicolumn{2}{c|}{\textbf{animal}} & \multicolumn{2}{c|}{\textbf{public transport}} & \multicolumn{2}{c|}{\textbf{building}}  & \multicolumn{2}{c|}{\textbf{vessel}} & \multicolumn{2}{c|}{\textbf{vehicle}} & \multicolumn{2}{c|}{\textbf{person}} \\
				\cline{3-14}
				 &  &  sheep  &  horse & coach & single-decker bus &  office building & skyscraper &  bulk carrier & motorboat &  SUV & sedan &  biking & walking\\
			    \hline
			    
			   \multirow{2}*{\textbf{animal}} & sheep &  0.411 & 0.399  & 0.318  & 0.391  & 0.319  & 0.335  & 0.382  & 0.396   & 0.404  & 0.402  & \textbf{0.425}  & \underline{0.423} \\
				
			    & horse & 0.380 & 0.342  & 0.311  & 0.408  & 0.366  & 0.410  & 0.379  & 0.420   & 0.381  & \textbf{0.442}  & 0.408  & \underline{0.430} \\
				\hline
				
			   \multirow{2}*{\textbf{public transport}} & coach & 0.462 & 0.442  & 0.504  & \textbf{0.564}  & 0.518  & 0.379  & 0.487  & 0.468   & 0.514  & \underline{0.524}  & 0.487  & 0.483 \\
				
			    & single-decker bus &  0.457 & 0.418  & \textbf{0.521}  & \underline{0.494}  & 0.443  & 0.445  & 0.446  & 0.469   & 0.454  & 0.486  & 0.437  & 0.425 \\
				\hline
				
			   \multirow{2}*{\textbf{building}} &  office building  & 0.638 & 0.695 & 0.664  & 0.650  & 0.715  & \textbf{0.758}  & 0.722  & 0.688  & 0.668  & 0.687  & \underline{0.738}  & 0.723 \\
				
			    & skyscraper & 0.618 & 0.615  & 0.573  & 0.594  & 0.585  & \textbf{0.657}  & 0.587  & 0.621   & \underline{0.651}  & 0.603  & 0.603  & \underline{0.651} \\
				\hline
				
			   \multirow{2}*{\textbf{vessel}} &  bulk carrier &  0.542 & 0.481  & 0.516  & 0.510  & 0.444  & 0.461  & 0.522  & \textbf{0.561}   & 0.542  & \underline{0.560}  & \textbf{0.561}  & 0.546 \\
				
			    & motorboat &  0.529 & 0.591  & 0.595  & 0.593  & 0.511  & 0.519  & 0.622  & 0.623   & 0.603  & \underline{0.646}  & \textbf{0.660}  & \underline{0.646} \\
				\hline
			
			   \multirow{2}*{\textbf{vehicle}} & SUV & 0.523 & 0.505  & 0.512  & 0.529  & 0.509  & 0.497  & 0.494  & 0.519   & 0.548  & 0.536  & \underline{0.550}  & \textbf{0.560} \\
				
			    & sedan &  0.434 & 0.437  & 0.444  & 0.456  & 0.434  & 0.422  & 0.451  & 0.447   & \underline{0.503}  & \textbf{0.513}  & 0.480  & 0.488 \\
				\hline
				
				\multirow{2}*{\textbf{person}} & biking &  0.255 & 0.243  & 0.206  & 0.243  & 0.231  & 0.253  & 0.262  & 0.312   & 0.323  & 0.346  & \underline{0.356}  & \textbf{0.370} \\
				
			    & walking &  0.284 & 0.296  & 0.294  & 0.312  & 0.316  & 0.293  & 0.315  & 0.330   & 0.362  & 0.375  & \underline{0.379}  & \textbf{0.402}  \\
				\hline
			\end{tabular}
		}
	\end{center}
    }
\vspace{-0.1cm}
\end{table*}

\begin{table}[t]
	{
	\scriptsize
	\renewcommand\arraystretch{1.0}
	\caption{Cross-superclass transfer learning results (cAUC) of SiamFC on the six largest superclasses, \ie \emph{person}, \emph{vehicle}, \emph{vessel}, \emph{building}, \emph{public transport} and \emph{animal}. }
	\label{tab:cross_superclass_transfer_learning}
	\begin{center}
		\setlength{\tabcolsep}{1.2mm}{
		\begin{tabular}{|l|c|c|c|c|c|c|}
				\hline
				\diagbox [width=8em,trim=l] {\textbf{Test}}{\textbf{Train}} & \textbf{animal} & \textbf{public transport} & \textbf{building}  & \textbf{vessel} & \textbf{vehicle} & \textbf{person} \\
				\hline
			    \hline
			    
			    \textbf{animal} &  \textbf{0.474}  & 0.409 & 0.418  & 0.430  & 0.449  & \underline{0.454}  \\
				
		        \textbf{public transport} &  0.347  & \textbf{0.457} & 0.363  & 0.393  & \underline{0.441}  & 0.355 \\
				
                \textbf{building} &  0.559  & \underline{0.569} & \textbf{0.609}  & 0.566  & 0.555  & \underline{0.569} \\
				
			    \textbf{vessel} &  0.599  & 0.589 & 0.559  & \textbf{0.637}  & \underline{0.619}  & 0.615 \\
			
			    \textbf{vehicle} &   0.523  & 0.534 & 0.477  & 0.536  & \textbf{0.551}  & \underline{0.540} \\
				
				\textbf{person} &   0.380  & 0.358 & 0.362  & \underline{0.399}  & 0.398  & \textbf{0.446}  \\
				\hline
				
				\textbf{Average} &   0.480  & 0.486 & 0.465  & 0.494  & \textbf{0.502}  & \underline{0.497}\\
				\hline

			\end{tabular}
		}
	\end{center}
    }
\end{table}

\subsection{Intraclass and Interclass Domain Generalization}

\myPara{Intraclass domain generalization.}
We choose the two largest superclasses, \ie \emph{vehicle} and \emph{person}, to conduct intraclass domain generalization experiments. More specifically, we choose the five largest target/motion classes in each superclass (\ie sedan, SUV, minivan, hatchback, and box truck from \emph{vehicle}, walking, biking, standing, riding an electric bicycle and sitting from \emph{person}). For each target/motion class, we randomly select $80\%$ of its videos as the training set and use the remaining $20\%$ as the test set. We train a widely used deep model (SiamFC) and follow the same settings as those used in the retraining experiment. The results are summarized in Tables~\ref{tab:Intra_class_vehicle} and \ref{tab:Intra_class_person}. The average results obtained for each target/motion class are shown at the bottom of each column. We have some observations upon inspection. First, the models trained on diverse target classes (\ie SUV and sedan) have good domain generalization performance on other unseen target classes. This is reasonable since the SUV and sedan are the two most common target classes in vehicles, involving various scenes, lighting changes, scale variations, appearance variations, and motion patterns. Similar phenomena are also observed in Table~\ref{tab:Intra_class_person}, \ie the models trained on the biking and walking motion classes can obtain good performance on other unseen motion classes. Second, domain generalization indicates the test difficulty of the corresponding class; \eg sedan generalizes best to other target classes but has the lowest average test score of 0.434. The above observations imply that the inherent diversity in target/motion classes helps models learn better domain generalization abilities and provides challenging videos for UAV tracking in the wild, \ie confirming the value of the proposed WebUAV-3M dataset.

\myPara{Interclass domain generalization.} We select the two largest subclasses from each of the six largest superclasses, \ie \emph{person} (walking and biking), \emph{vehicle} (sedan and SUV), \emph{vessel} (motorboat and bulk carrier), \emph{building} (skyscraper and office building), \emph{public transport} (single-docker bus and coach) and \emph{animal} (horse and sheep). For each subclass, we randomly select $80\%$ of its videos for training and $20\%$ for testing. The interclass domain generalization results of SiamFC are presented in Table~\ref{tab:Inter_class}. As seen in the table, the models trained on the sedan, biking, and walking subclasses perform well on other subclasses belonging to the same superclass but also on subclasses belonging to the different superclasses. This indicates that the subclasses have vast appearance and motion variations and can be effectively generalized to other subclasses. In addition, except for the \emph{public transport}, \emph{building}, and \emph{person} superclasses, the two best models on other superclasses are not trained on the same superclass. This is reasonable, as the subclasses in \emph{public transport}, \emph{building} and \emph{person} have some essential and special appearance or motion patterns. Therefore, the model learns special knowledge and can perform well on the corresponding superclass. For instance, in \emph{building}, the appearances of office buildings and skyscrapers are usually cubes with lighting changes and viewpoint changes. The target is relatively stationary, but the surrounding objects change. Both the biking and walking subclasses in \emph{person} have dramatic appearance changes and complex types of motion, making them distinct from the subclasses in other superclasses.

\subsection{Cross-Superclass and Dataset Transfer Learning} 
\myPara{Cross-superclass transfer learning.} We further evaluate the cross-superclass transfer learning abilities of models on the six largest superclasses, \ie \emph{person}, \emph{vehicle}, \emph{vessel}, \emph{building}, \emph{public transport} and \emph{animal}. For each superclass, we randomly select $80\%$ of the videos as the training set and the other $20\%$ of the videos as the test set. The results of SiamFC are summarized in Table~\ref{tab:cross_superclass_transfer_learning}. The diagonal scores represent the results obtained when conducting training and testing on the same superclass. In addition, we also report the average evaluation results obtained on each training superclass at the bottom of each row. We make the following observations. First, the model has the best performance when the training and test sets are from the same superclass; \ie the smaller the domain gap is, the better the transferability of the model. Second, \emph{vehicle} provides the best cross superclass transfer learning ability (\ie an average cAUC of 0.502), followed by \emph{person} (\ie an average cAUC of 0.497). This is reasonable as \emph{vehicle} and \emph{person} have the most diverse target categories and motion classes, respectively, which can help the model learn better feature representations.

\begin{table}[t]
	{
	\scriptsize
	\renewcommand\arraystretch{1.0}
	\caption{Cross-dataset evaluation results of GOTURN, SiamFC and SiamRPN using AO/SR$_{0.5}$ (2-4 rows) or cAUC/mAcc (5-10 rows). Three deep trackers are trained on the GOT-10k, VisDrone, and WebUAV-3M training sets and evaluated on the GOT-10k, VisDrone, and WebUAV-3M test sets.}
	\label{tab:cross_dataset_evaluation}
	\begin{center}
		\setlength{\tabcolsep}{1.195mm}{
		\begin{tabular}{|l|c|c|c|c|}
			\hline
			\textbf{Tracker}  & \textbf{GOTURN} & \textbf{SiamFC} & \textbf{SiamRPN} & \textbf{Average}\\
			\hline
			\hline

		    \textbf{VisDrone}$\rightarrow$\textbf{GOT-10k} &   0.152/0.115  & {0.261}/{0.250}  & 0.169/0.123 & 0.194/0.163\\

			\textbf{GOT-10k}$\rightarrow$\textbf{GOT-10k} & 0.334/0.355   & 0.295/0.337  & {0.386}/{0.438} & \textbf{0.338}/\textbf{0.377} \\

            \textbf{WebUAV-3M}$\rightarrow$\textbf{GOT-10k} &  0.275/0.255   & 0.310/0.325  & {0.355}/{0.400} & \underline{0.313}/\underline{0.327}\\
			\hline
			
            \textbf{VisDrone}$\rightarrow$\textbf{VisDrone} &   0.356/0.375   & {0.435}/{0.474}  & 0.327/0.355 & 0.373/0.401 \\
				
			\textbf{GOT-10k}$\rightarrow$\textbf{VisDrone} & 0.350/0.389   & 0.501/0.548  & {0.532}/{0.584} & \underline{0.461}/\underline{0.507} \\
				
	      	\textbf{WebUAV-3M}$\rightarrow$\textbf{VisDrone} & 0.404/0.433   & 0.553/0.596  & {0.562}/{0.615} & \textbf{0.506}/\textbf{0.548} \\
			\hline

			\textbf{VisDrone}$\rightarrow$\textbf{WebUAV-3M} &  0.102/0.133   & {0.359}/{0.398}  & 0.155/0.174 & 0.205/0.235 \\
				
	      	\textbf{GOT-10k}$\rightarrow$\textbf{WebUAV-3M} &   0.203/0.238  & 0.355/0.388  & 0.386/0.431 & \underline{0.315}/\underline{0.352} \\

            \textbf{WebUAV-3M}$\rightarrow$\textbf{WebUAV-3M} &   0.280/0.318   & {0.417}/{0.458}  & 0.415/0.457  & \textbf{0.371}/\textbf{0.411}  \\
			\hline
			\end{tabular}
		}
	\end{center}
    }
\end{table}

\myPara{Cross-dataset transfer learning.} We evaluate the transfer learning abilities of models trained on different datasets, including two UAV tracking datasets (\ie VisDrone and WebUAV-3M) and a GOT dataset (\ie GOT-10k). We retrain three representative deep models (\ie GOTURN, SiamFC and SiamRPN) using the VisDrone, GOT-10k and WebUAV-3M training sets. The cross-dataset evaluation results obtained on the above three datasets are summarized in Table~\ref{tab:cross_dataset_evaluation}. On GOT-10k, we follow its submission policy and evaluation protocol~\cite{huang2019got} and submit the tracking results to the official evaluation server. We report the results (\ie average overlaps (AOs) and success rates (SRs)) obtained on the GOT-10k test set in Table~\ref{tab:cross_dataset_evaluation}.  The model trained on WebUAV-3M can effectively generalize to VisDrone and GOT-10k. Specifically, using the WebUAV-3M training set to train the models, the best evaluation results are obtained on the VisDrone and WebUAV-3M test sets, while the second-best evaluation result is achieved on the GOT-10k test set. This is because VisDrone only has 10 classes, which is far from sufficient for enabling a model to generalize well to other datasets. In contrast, WebUAV-3M has more than 220 target classes and can generalize well to VisDrone and GOT-10k. Because of the large number of classes (563) in GOT-10k and the large domain gap between the GOT-10k GOT dataset and our WebUAV-3M dataset, the model trained on our WebUAV-3M training set achieves suboptimal results on the GOT-10k test set. The above results imply that training on a diverse dataset can improve the generalization ability of deep models on other test data, again confirming the value of our million-scale and highly diverse WebUAV-3M dataset.

\subsection{Qualitative Evaluation}
\begin{figure}[t]   
	\centering\centerline{\includegraphics[width=1.0\linewidth]{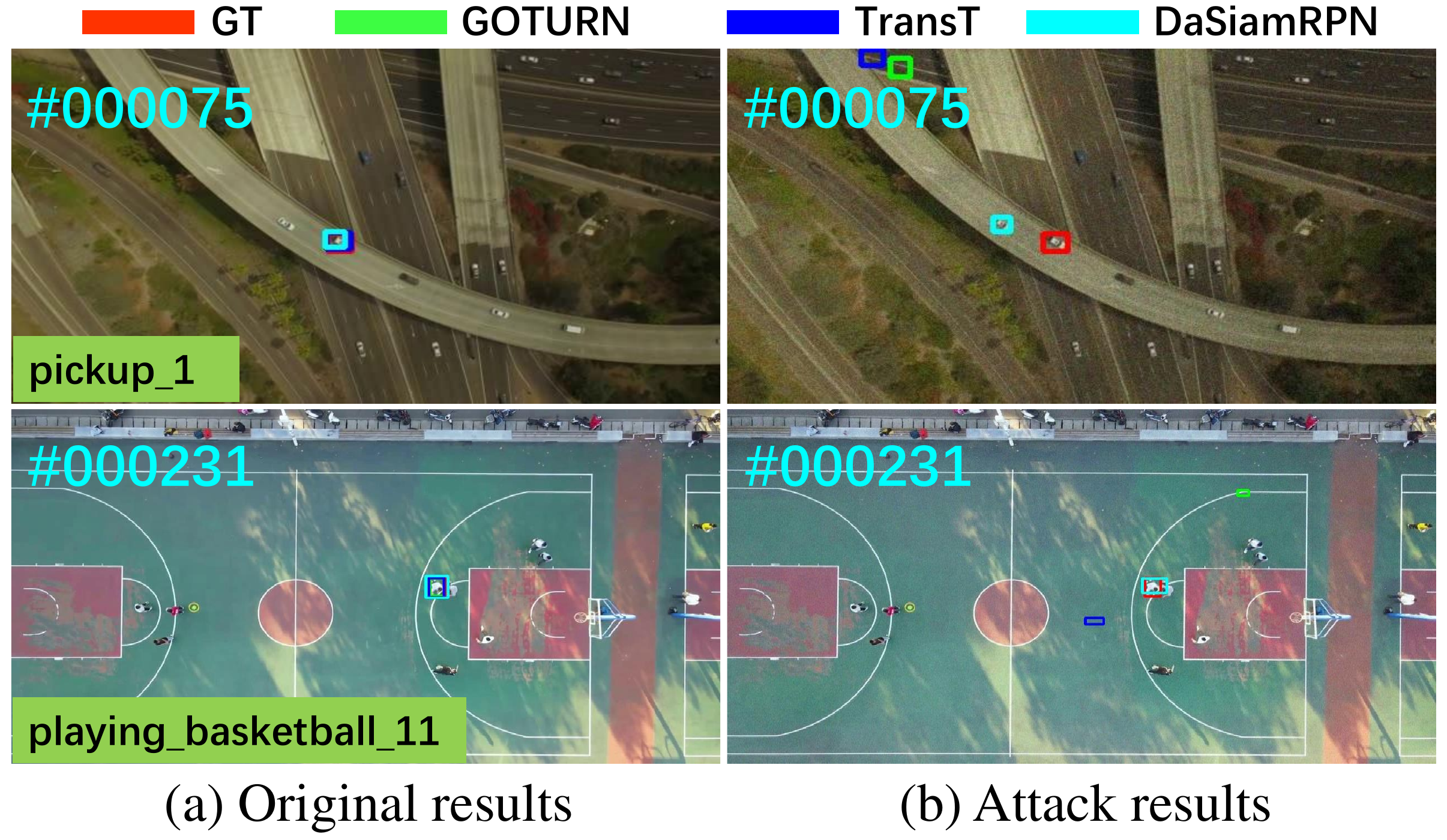}}
	\caption{Qualitative evaluation results obtained on two challenging sequences with adversarial examples ($\Lambda=6000$).}
	\label{fig:Visualization_results_Adversarial_examples}
\end{figure}

To qualitatively analyze different trackers, we demonstrate the visual tracking results obtained in the adversarial attack scenario~\cite{jia2021iou} by TransT, DaSiamRPN and GOTURN on two challenging sequences. The original tracking results and the attack results are shown in Figs.~\ref{fig:Visualization_results_Adversarial_examples} (a) and (b), respectively. Adversarial examples make the above trackers yield inaccurate target location predictions. We also present the qualitative results obtained for six other scenarios, \ie long-term occlusion, target distortions, dual-dynamic disturbances, small targets, high-speed motion, and low light, in the~\textbf{supplemental material}.

\section{Conclusion and Future Research}
\myPara{Conclusion.} In this paper, we introduce WebUAV-3M with visual box annotations, natural language specifications, and audio descriptions for the first time to enable comprehensive and rigorous evaluations of deep UAV tracking methods. To the best of our knowledge, WebUAV-3M is the most comprehensive and largest UAV tracking benchmark with multi-modal annotations to date. By releasing WebUAV-3M, we aim to offer a dedicated platform for the unified training and assessment of deep UAV tracking algorithms with million-scale dense annotations. To construct a sound and high-quality benchmark, we propose a general SATA pipeline to label the tremendous WebUAV-3M dataset. In addition, the UTUSC evaluation protocol, as well as seven subtest sets with fine-grained and challenging scenarios, are used to enable reliable evaluations. The experimental results obtained on WebUAV-3M imply that much room for improvement remains regarding high-performance deep UAV tracking. We hope that this benchmark will facilitate future research on large-scale multi-modal deep UAV tracking. Moreover, all the datasets, evaluation protocols, codes, and tracking results have been made public and researchers are welcome to jointly develop WebUAV-3M as an ecosystem by increasing its size, the number of target categories, attributes, evaluation metrics, and types of tasks covered.

\myPara{Future Directions for UAV Tracking.} Through the analysis of popular deep trackers, we find that the following widely existing challenges have not been well studied in tracking scenarios. 1) Nighttime tracking. The nighttime conditions, \ie low light, and even invisible targets, harm the performance of deep trackers designed primarily for high-visibility inputs. 2) Adversarial examples. We empirically identify that CNN- and transformer-based deep trackers are vulnerable to perceiving input samples injected with imperceptible perturbations. The design of more efficient adversarial attack algorithms and the removal of the threat of adversarial examples to deep trackers are valuable investigations. 3) Multi-modal tracking. Determining how to effectively integrate nonvisual features into deep tracking models poses a fundamental challenge in appearance-based visual tracking, which is far from being explored. 4) Data imbalances. Learning robust deep tracking models from data with imbalanced class distributions (\eg long tails) is a significant challenge for the tracking community. There is an urgent demand for studying robust deep trackers for imbalanced data not only because minority class instances often represent the target objects of interest in real-world applications but also because class distributions have significant impacts on existing deep trackers; with population disadvantages, minority class instances are essentially more vulnerable to being incorrectly located by tracking algorithms. 

In addition to deep UAV tracking, the rich visual, language, and audio annotations in our video dataset enable a wide range of potential studies and applications, such as video transformer pretraining on extra-large-scale datasets, vision-language-audio pretraining, UAV-based video understanding, wildlife conservation, crowd and vehicle counting, and human behavior understanding with UAVs.

\ifCLASSOPTIONcompsoc
  \section*{Acknowledgments}
\else
  \section*{Acknowledgment}
\fi

{This work is supported by the National Natural Science Foundation of China (No. 62101351), the Guangdong Basic and Applied Basic Research Foundation (No. 2020A1515110376), Shenzhen Outstanding Scientific and Technological Innovation Talents Ph.D. Startup Project (No. RCBS20210609104447108), the Key-Area Research and Development Program of Guangdong Province (2020B0101350001), and the Chinese University of Hong Kong, Shenzhen.}

\ifCLASSOPTIONcaptionsoff
  \newpage
\fi



%

\bibliographystyle{IEEEtran}
\bibliography{references}

%








\vspace{-1.5cm}
\begin{IEEEbiography}[{\includegraphics[width=1in,height=1.25in,clip,keepaspectratio]{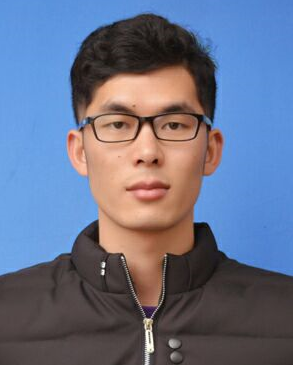}}]{Chunhui Zhang} is currently pursuing the Ph.D. degree in Shanghai Jiao Tong University, China. He received his B.S. and M.S. degrees from Hunan University of Science and Technology, and University of Chinese Academy of Sciences in 2016 and 2020, respectively. He also spent 2 years (2020-2022) at the Chinese University of Hong Kong, Shenzhen as a research associate. His major research interests are focused on machine learning and visual tracking.
\end{IEEEbiography}

\vspace{-1.5cm}
\begin{IEEEbiography}[{\includegraphics[width=1in,height=1.25in,clip,keepaspectratio]{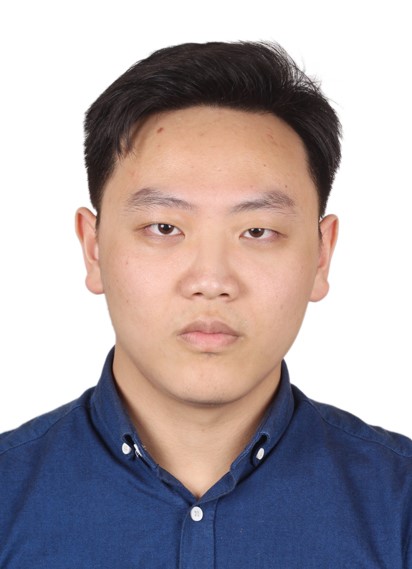}}]{Guanjie Huang} received his B.S. and M.S. degrees from University of Electronic Science and Technology of China, and Australian National University in 2018 and 2020, respectively. He was a research associate at Shenzhen Research Institute of Big Data from 2021 to 2022. His research interests include visual tracking, image generation, and learning to optimize. \end{IEEEbiography}

\vspace{-1.5cm}
\begin{IEEEbiography}[{\includegraphics[width=1in,height=1.25in,clip,keepaspectratio]{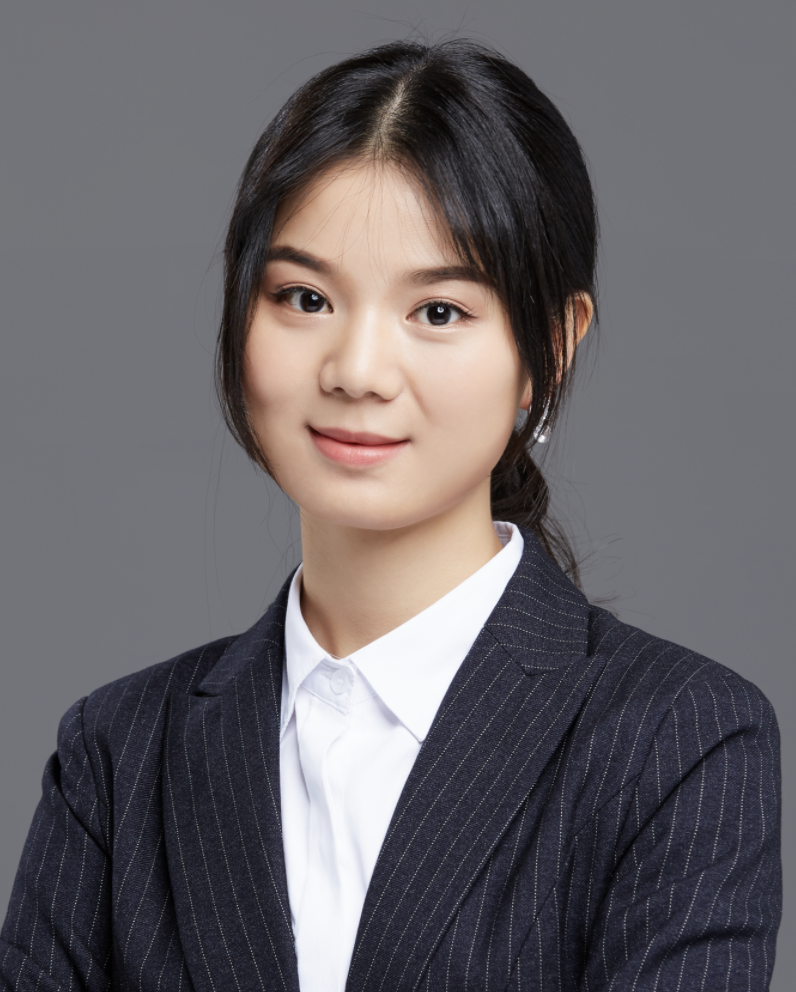}}]{Li Liu} works as a research scientist at Shenzhen Research Institute of Big Data, Shenzhen, China. She received Ph.D. degree in 2018 from Gipsa lab, University Grenoble Alpes, Grenoble, France. From September 2018 to September 2019, she was a postdoc researcher in the Department of Electrical, Computer, and Biomedical Engineering, Ryerson University, Toronto, Canada. Her current research interests include automatic audio-visual speech recognition, multi-modal fusion, Cued Speech development, lips/hand gesture recognition, and medical imaging. She has published in more than twenty top international peer-reviewed journals and conferences. She received the International Sephora Berribi Scholarship for Women Scientists and the French Phonetics Association (AFCP) Young researcher Scholarship in 2017.
\end{IEEEbiography}

\vspace{-1.0cm}
\begin{IEEEbiography}[{\includegraphics[width=1in,height=1.25in,clip,keepaspectratio]{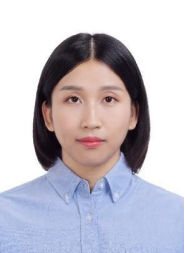}}]{Shan Huang} works as an engineer at Shenzhen Research Institute of Big Data, Shenzhen, China. She received her B.S. and M.S. degrees from Zhengzhou University, and Shenzhen University in 2017 and 2020, respectively. Her major research interests are focused on machine learning, medical image analysis, and visual tracking.
\end{IEEEbiography}

\vspace{-0.5cm}
\begin{IEEEbiography}[{\includegraphics[width=1in,height=1.25in,clip,keepaspectratio]{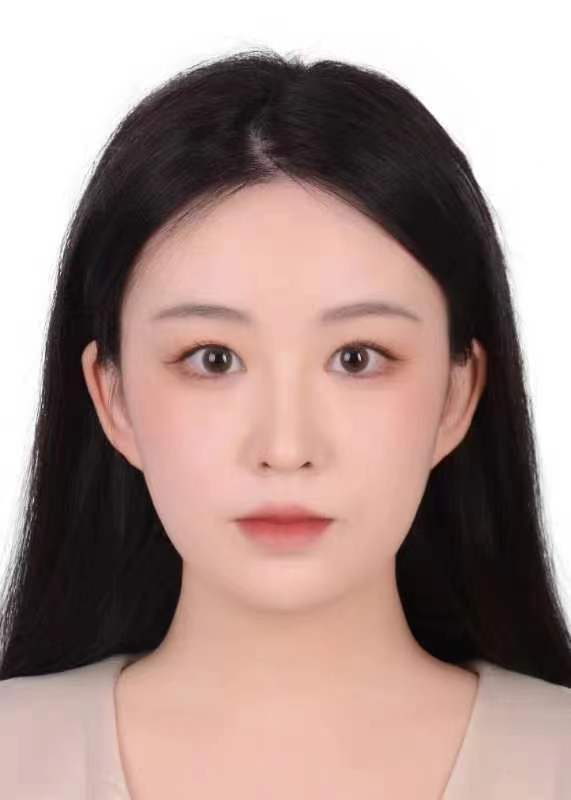}}]{Yinan Yang} received her B.S. degree from Xiamen University, Fujian, China, in 2020. She was a research assistant at the Chinese University of Hong Kong, Shenzhen from 2020 to 2021, and now she is going to finish her master's education at the University of Edinburgh. Her research interests are focused on machine learning and computer vision.
\end{IEEEbiography}

\vspace{-0.5cm}
\begin{IEEEbiography}[{\includegraphics[width=1in,height=1.25in,clip,keepaspectratio]{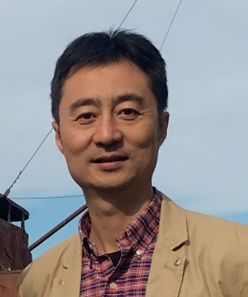}}]{Xiang Wan} received the B.S. from Renmin University, China, in 1994 and the M.S. and Ph.D. from University of Alberta in 2002 and 2006, respectively. He was a research assistant professor at Hong Kong Baptist University from 2012 to 2017. He is currently a senior research scientist at Shenzhen Research Institute of Big Data (SRIBD), China. His research interests include data mining, machine learning, bioinformatics, and medical data analysis. He has published more than 60 papers in many top-tier journals, including Nature Genetics, American Journal of Human Genetics, PLoS computational biology, BMC Genetics, Bioinformatics, BMC Bioinformatics, Neuroinformatics and IEEE Transactions on Neural Networks and Learning Systems.  
\end{IEEEbiography}

\vspace{-0.5cm}
\begin{IEEEbiography}[{\includegraphics[width=1in,height=1.25in,clip,keepaspectratio]{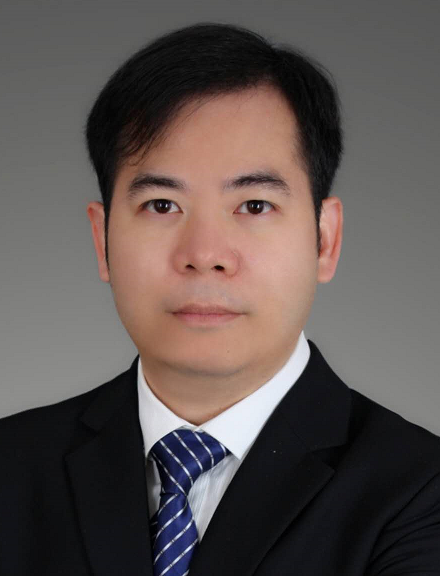}}]{Shiming Ge} (M'13-SM'15) is a professor with the Institute of Information Engineering, Chinese Academy of Sciences. Prior to that, he was a senior researcher and project manager at Shanda Innovations and a researcher at Samsung Electronics and Nokia Research Center. He received B.S. and Ph.D. degrees both in Electronic Engineering from the University of Science and Technology of China (USTC) in 2003 and 2008, respectively. His research mainly focuses on computer vision, data analysis, machine learning, and AI security, especially trustworthy learning solutions towards scalable applications. He is a senior member of IEEE, CSIG, and CCF.
\end{IEEEbiography}

\vspace{-0.5cm}

\begin{IEEEbiography}[{\includegraphics[width=1in,height=1.25in,clip,keepaspectratio]{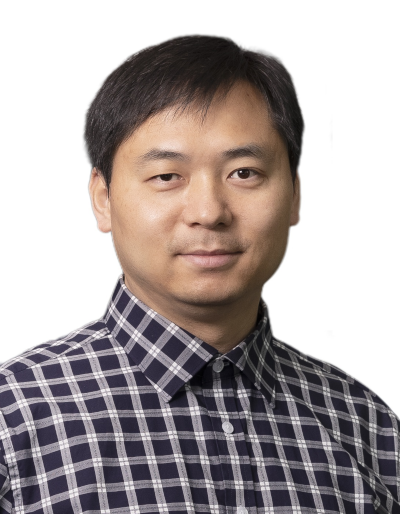}}]{Dacheng Tao} (Fellow, IEEE) is the Inaugural Director of the JD Explore Academy and a Senior Vice President of JD.com. He is also an advisor and chief scientist of the digital sciences initiative in the University of Sydney. He mainly applies statistics and mathematics to artificial intelligence and data science, and his research is detailed in one monograph and over 200 publications in prestigious journals and proceedings at leading conferences. He received the 2015 Australian Scopus-Eureka Prize, the 2018 IEEE ICDM Research Contributions Award, and the 2021 IEEE Computer Society McCluskey Technical Achievement Award. He is a fellow of the Australian Academy of Science, AAAS, ACM and IEEE.
\end{IEEEbiography}

\clearpage
\renewcommand\thetable{\Alph{table}}
\renewcommand\thefigure{\Alph{figure}}
\pagestyle{plain}
\setcounter{page}{1} 

\addtocounter{figure}{-14}
\addtocounter{table}{-11}
\addtocounter{footnote}{-7}

\noindent {\textbf{\Large Supplemental Material}}\\

This supplemental material contains three parts:

\renewcommand\thesection{\Alph{section}}
\setcounter{section}{0} 

\begin{itemize}
\item Section~\ref{sec:details_of_WebUAV3M} presents more details about the constructed WebUAV-3M dataset.

\item Section~\ref{sec:more_discussion_about_SATA} provides more discussion about the proposed SATA pipeline.

\item Section~\ref{sec:additional_results} demonstrates the results of tracking by joint language and bounding box, and more quantitative and qualitative evaluation results.

\end{itemize}

\section{Details about WebUAV-3M}
\label{sec:details_of_WebUAV3M}

\subsection{Annotation Rules}
\label{sec:Annotation Rules}
The details of our annotation rules are as follows.

\textbf{Rule 1:} All visible parts of the chosen object must be included, and bounding boxes must be drawn as tightly as possible. 

\textbf{Rule 2:} If the target is completely obscured or disappears in one frame, a bounding box is not provided to this frame.

\textbf{Rule 3:} A valid annotation bounding box must be located at the first frame of each video to show the target object.
    
\textbf{Rule 4:} When an object is occluded, disappears, or reappears, it is necessary to annotate the whole gradual process of the target.

\begin{figure}[h] 
    \centering
    \subfloat[Tight bounding box on the first frame]{\includegraphics[width = 0.48\columnwidth]{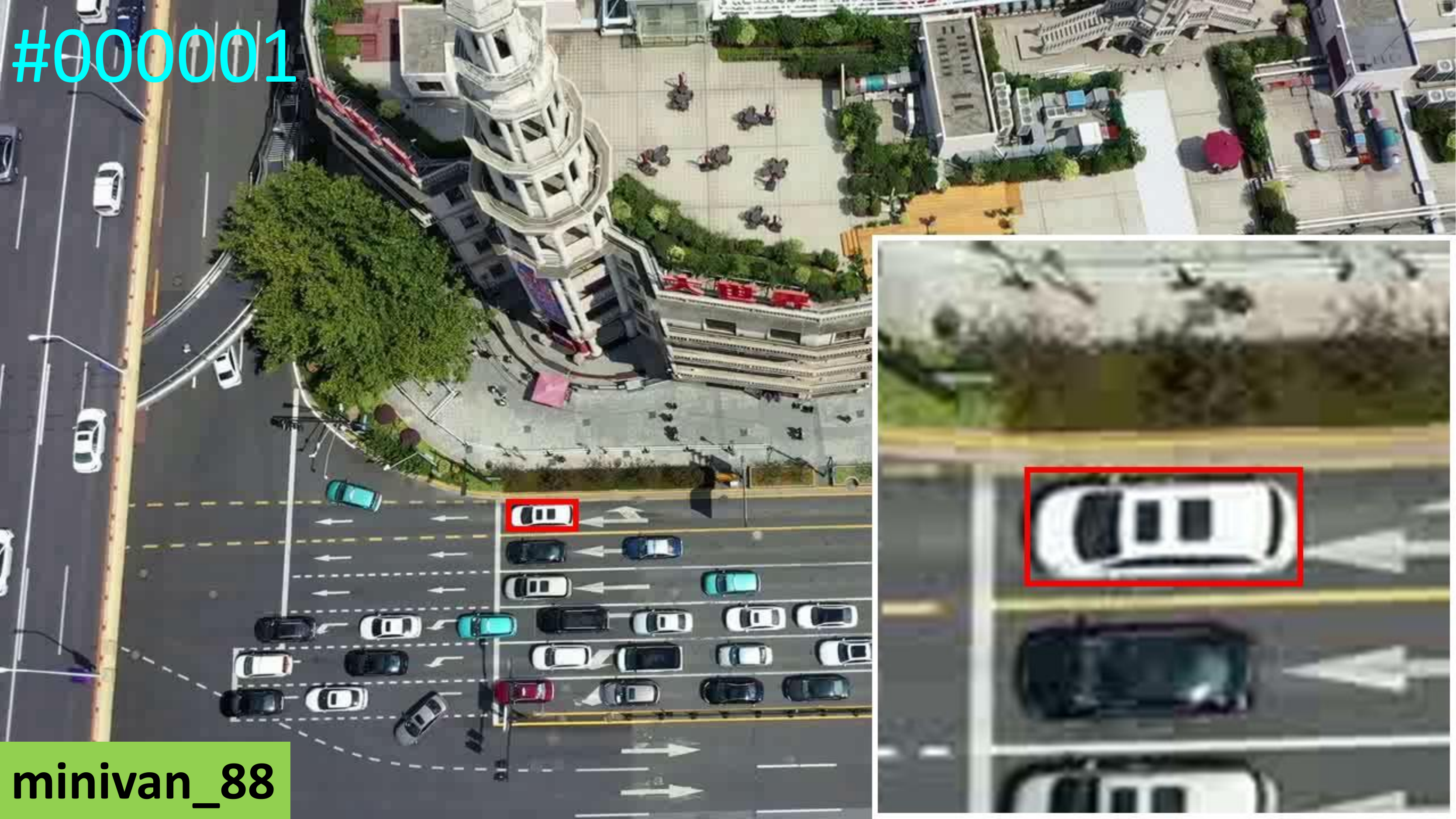}}
    ~
    \subfloat[The minivan is partially occluded by the overpass ]{\includegraphics[width = 0.48\columnwidth]{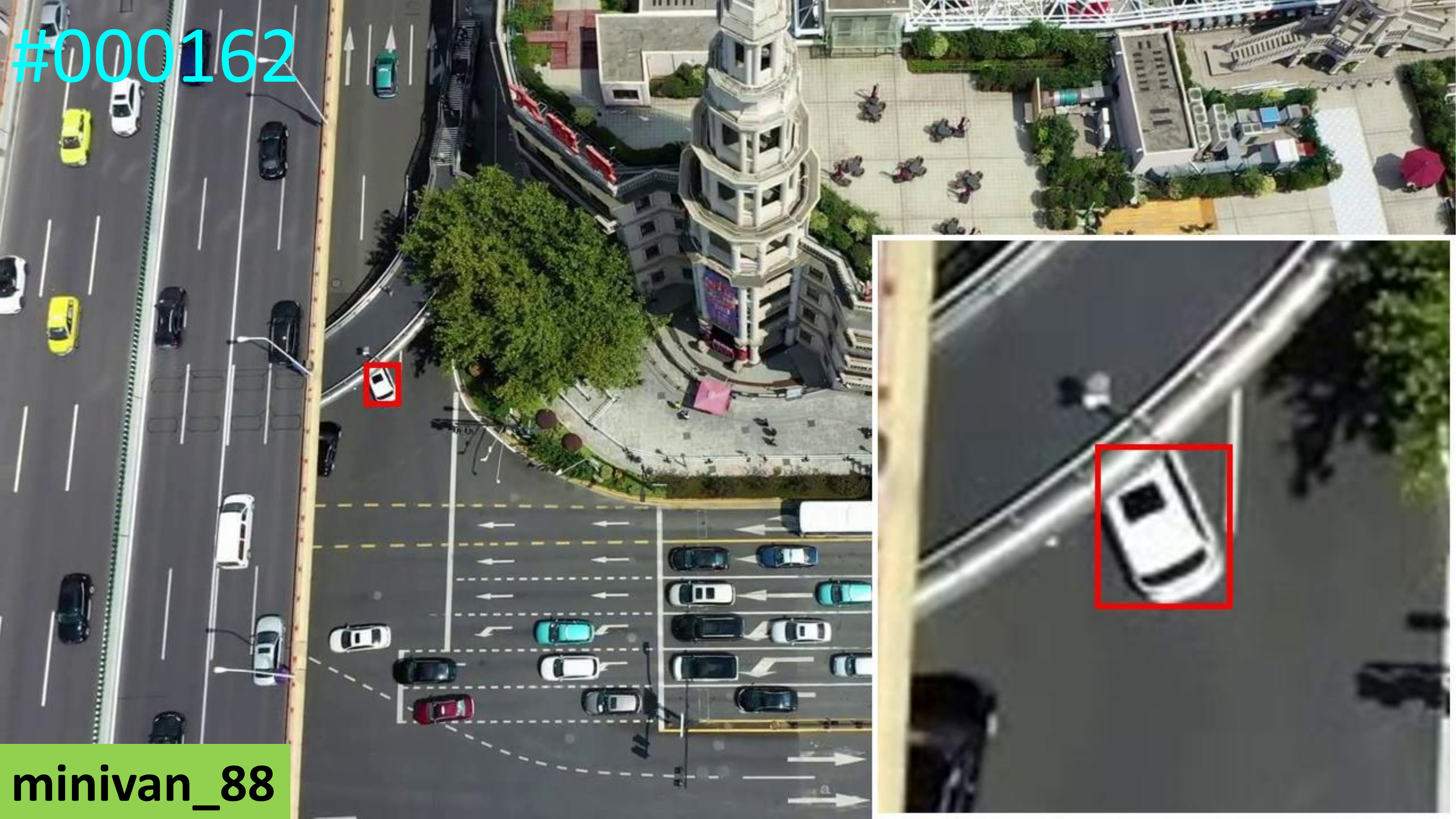}}\\
    
    \subfloat[The minivan is fully occluded by the overpass]{\includegraphics[width =0.48\columnwidth]{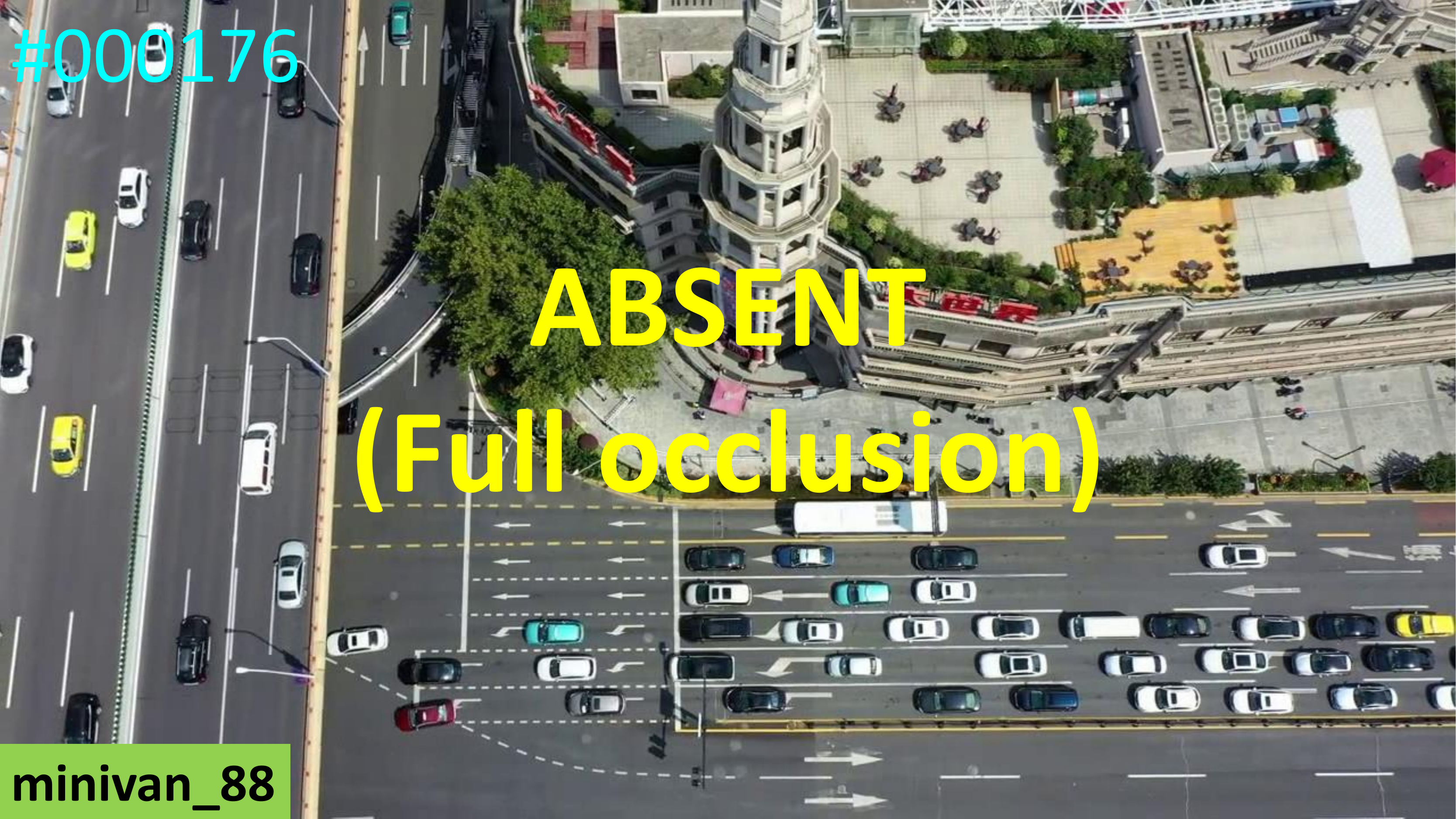}}
    ~
    \subfloat[The minivan partially passed the overpass]{\includegraphics[width =0.48\columnwidth]{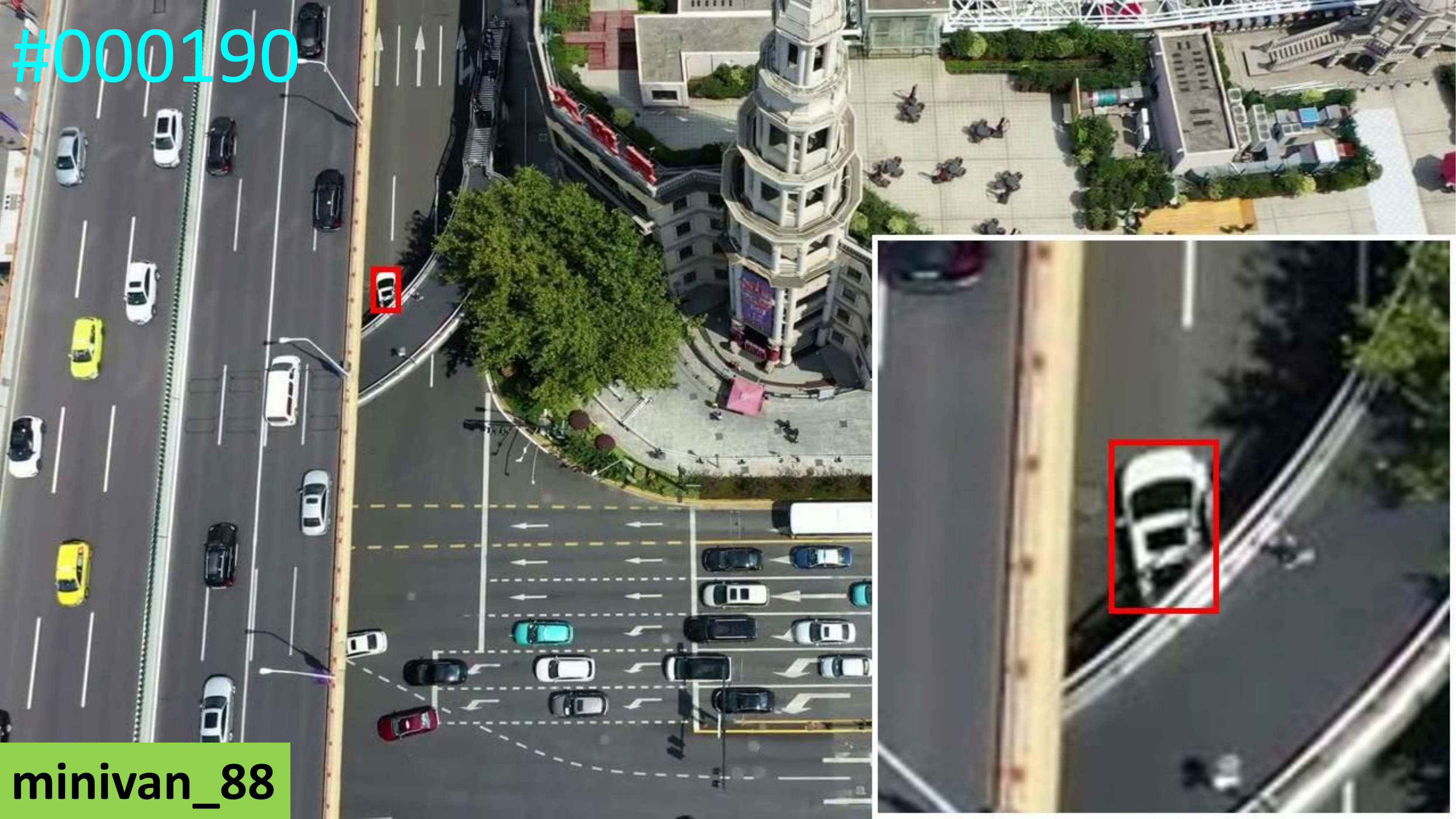}}
\caption{Illustration of our annotation rules using \textit{minivan$\_$88}, where a white minivan turns right and crosses under an overpass. In (a), the bounding box is kept as tight as possible. In (b) and (d), the objects are partially occluded. We provide tight bounding boxes that are close to the visible parts. In (c), no bounding box is present due to the FO scenario.}
\label{fig:rule}
\end{figure}

Fig.~\ref{fig:rule} shows an example of applying our rules. To achieve rule 1 when given a tremendous number of frames, we allow the bounding boxes to be slightly smaller than the object boundary rather than larger. Additionally, we retain the flexibility of not necessarily having a bounding box on every frame to simulate realistic tracking situations in which the target object can be occluded or OV. Due to this rule, we form a specific scenario (\ie long-term occlusion) to evaluate tracking algorithms’ recatching abilities, as demonstrated in Section~\ref{sec:UTUSC}. In rule 4, we believe that accurately annotating the vanishing and reappearing processes can result in more positive impacts on tracker training. In addition, if the target object disappears in a frame (either FO or OV), the annotators provide an absent label to the frame.

\subsection{Attribute Definitions}
\label{sec:attribute_definition}

Table~\ref{tab:Attribute_defination} lists the definition of each attribute in WebUAV-3M. {\color{black}These challenging global attributes can be divided into two categories: the target-level (\ie LR, PO, FO, OV, FM, CM, VC, ROT, DEF, BC, SV, ARV, IV, and MB) and the video-level (\ie COM, SIZ, and LEN).}

\begin{table}[h]
	{\scriptsize
	\renewcommand\arraystretch{1.2}
	\caption{Descriptions of the 17 attributes in WebUAV-3M.}
	\label{tab:Attribute_defination}
	\begin{center}
		\setlength{\tabcolsep}{1.6mm}{
			\begin{tabular}{|l|l|}
				\hline
				Attribute & Definition \\
				\hline
			    \hline	
			    \textbf{01. LR} & \tabincell{l}{The target box is smaller than 400 pixels in at least one frame.} \\
			    
			    \textbf{02. PO} & The target is partially occluded in the sequence. \\
			    
                \textbf{03. FO} &  The target is fully occluded in the sequence. \\
                
                \textbf{04. OV} &  The target completely leaves the video frame. \\
                
			    \textbf{05. FM} &  \tabincell{l}{The motion of the ground truth is larger than 20 pixels.} \\
			    
                \textbf{06. CM} &  Abrupt motion of the camera. \\
                
                \textbf{07. VC} &  Viewpoint affects target appearance significantly. \\
                
                \textbf{08. ROT} & The target rotates in the image. \\
                
                \textbf{09. DEF} &  The target is deformable during tracking. \\
                
                \textbf{10. BC} &  \tabincell{l}{The background has a similar appearance as the target.} \\
                
                \textbf{11. SV} &  The ratio of the bounding box is outside the range [0.5, 2]. \\
                
                \textbf{12. ARV} &  \tabincell{l}{The ratio of bounding box aspect ratio is outside the rage [0.5, 2].} \\
                
                \textbf{13. IV} &  The illumination in the target region changes. \\
                
                \textbf{14. MB} & \tabincell{l}{The target region is blurred due to target or camera motion.} \\
                
                \textbf{15. COM} & The complexity of the current video (easy or medium or hard).\\
                
                \textbf{16. SIZ} &  \tabincell{l}{The size $s\!=\!\sqrt{w \!\times \!h}$ of current video is small ($s\! <\! \sqrt{640 \! \times\! 480}$ pixels),\\ or medium ($ \sqrt{640 \! \times\! 480}$ pixels $\!\leq \! s \!\leq\! \sqrt{1280 \!\times\! 720}$ pixels), \\or large ($s\! >\! \sqrt{1280 \!\times\! 720}$ pixels).} \\
                
                \textbf{17. LEN} & \tabincell{l}{The length ($l$) of current video is short ($l \!\leq \!600$ frames, 20s for 30 fps),\\ or medium ($600 \!<\! l \!\leq\! 1800$ frames, 60s for 30 fps), \\or long($l \!>\!1800$ frames).}\\
				\hline	
			\end{tabular}
		}
	\end{center}
    }
\end{table}

\subsection{The Co-occurrence Distributions of the 17 Attributes in WebUAV-3M}
\label{sec:co-occurrence}
We provide the co-occurrence distributions of the 17 attributes described above in Table~\ref{tab:Attribute}. This demonstrates that one video sequence may be annotated with many attributes, and PO and BC co-occur more frequently than other pairs of attributes in the proposed WebUAV-3M dataset.

\begin{table*}[t]
	{\scriptsize
	\renewcommand\arraystretch{1.05}
	\caption{Co-occurrence distributions of the attributes in WebUAV-3M. The diagonal demonstrates the distribution of the entire dataset, and each row or column represents the distribution of the corresponding attribute subset. The top five co-occurrence attribute pairs are PO and BC, BC and SIZ-M, BC and LEN-S, ROT and BC, and PO and SIZ-M.
	}
	\label{tab:Attribute}
	\begin{center}
	\setlength{\tabcolsep}{0.9mm}{
    \begin{tabular}{|l|c|c|c|c|c|c|c|c|c|c|c|c|c|c|c|c|c|c|c|c|c|c|c|} 
        \hline
        \textbf{Attribute}  & \textbf{LR}	&\textbf{PO}	&\textbf{FO}	&\textbf{OV}	&\textbf{FM}	&\textbf{CM}	&\textbf{VC}	&\textbf{ROT}&	\textbf{DEF}&	\textbf{BC}&	\textbf{SV}&	\textbf{ARV}	&\textbf{IV}& \textbf{MB} &	\textbf{COM-E} &	\textbf{COM-M} &	\textbf{COM-H} &	\textbf{SIZ-S} &	\textbf{SIZ-M} & \textbf{SIZ-B} &	\textbf{LEN-S} &	\textbf{LEN-M}	&\textbf{LEN-L}\\
        \hline 
        \hline
        
        \textbf{LR}  & \textbf{1860}  & 1153  & 273  & 131  & 192  & 964  & 812  & 966  & 503  & 1522  & 788  & 602  & 334  & 1028  & 501  & 879  & 480  & 460  & 1255  & 145  & 1270  & 551  & 39 \\ 
       
        \textbf{PO}  & 1153  & \textbf{2674}  & 424  & 286  & 461  & 1198  & 1040  & 1370  & 877  & {\color{black}{2063}}  & 984  & 960  & 582  & 1050  & 737  & 1193  & 744  & 469  & {\color{black}{1593}}  & 612  & 1478  & 1069  & 127 \\ 
         
        \textbf{FO}  & 273  & 424  & \textbf{425}  & 32  & 57  & 146  & 126  & 257  & 150  & 372  & 178  & 172  & 104  & 206  & 8  & 132  & 285  & 30  & 235  & 160  & 233  & 178  & 14 \\ 
        
        \textbf{OV}  & 131  & 286  & 32  & \textbf{302}  & 89  & 153  & 123  & 183  & 127  & 211  & 134  & 167  & 59  & 133  & 39  & 119  & 144  & 63  & 128  & 111  & 144  & 122  & 36\\ 
        
        \textbf{FM}  & 192  & 461  & 57  & 89  & \textbf{587}  & 404  & 393  & 248  & 163  & 363  & 373  & 343  & 135  & 132  & 155  & 260  & 172  & 39  & 479  & 69  & 194  & 302  & 91 \\ 
        
        \textbf{CM}  & 964  & 1198  & 146  & 153  & 404  & \textbf{2037}  & 1516  & 816  & 521  & 1340  & 822  & 654  & 314  & 729  & 714  & 888  & 435  & 520  & 1320  & 197  & 1179  & 749  & 109 \\
        
        \textbf{VC}  & 812  & 1040  & 126  & 123  & 393  & 1516  & \textbf{1740}  & 731  & 432  & 1141  & 715  & 579  & 290  & 623  & 554  & 774  & 412  & 402  & 1176  & 162  & 958  & 674  & 108 \\ 
        
        \textbf{ROT}  & 966  & 1370  & 257  & 183  & 248  & 816  & 731  & \textbf{2088}  & 1017  & {\color{black}{1671}}  & 587  & 797  & 511  & 968  & 582  & 930  & 576  & 378  & 1165  & 545  & 1185  & 823  & 80 \\
        
        \textbf{DEF}  & 503  & 877  & 150  & 127  & 163  & 521  & 432  & 1017  & \textbf{1273}  & 1059  & 314  & 471  & 326  & 521  & 328  & 553  & 392  & 180  & 667  & 426  & 645  & 565  & 63 \\
        
        \textbf{BC}  & 1522  & {\color{black}{2063}}  & 372  & 211  & 363  & 1340  & 1141  & {\color{black}{1671}}  & 1059  & \textbf{3242}  & 940  & 917  & 728  & 1333  & 1172  & 1374  & 696  & 556  & {\color{black}{1982}}  & 704  & {\color{black}{1870}}  & 1242  & 130 \\ 
        
        \textbf{SV}  & 788  & 984  & 178  & 134  & 373  & 822  & 715  & 587  & 314  & 940  & \textbf{1391}  & 639  & 240  & 544  & 427  & 611  & 353  & 308  & 923  & 160  & 760  & 532  & 99 \\ 
        
        \textbf{ARV}  & 602  & 960  & 172  & 167  & 343  & 654  & 579  & 797  & 471  & 917  & 639  & \textbf{1241}  & 258  & 457  & 325  & 520  & 396  & 198  & 826  & 217  & 549  & 599  & 93 \\ 
        
        \textbf{IV}  & 334  & 582  & 104  & 59  & 135  & 314  & 290  & 511  & 326  & 728  & 240  & 258  & \textbf{915}  & 406  & 258  & 406  & 251  & 105  & 514  & 296  & 478  & 399  & 38 \\
        
        \textbf{MB}  & 1028  & 1050  & 206  & 133  & 132  & 729  & 623  & 968  & 521  & 1333  & 544  & 457  & 406  & \textbf{1730}  & 427  & 805  & 498  & 518  & 856  & 356  & 1218  & 486  & 26 \\ 
        
        \textbf{COM-E}  & 501  & 737  & 8  & 39  & 155  & 714  & 554  & 582  & 328  & 1172  & 427  & 325  & 258  & 427  & \textbf{1848}  & 0  & 0  & 651  & 861  & 336  & 1125  & 633  & 90 \\
        
        \textbf{COM-M}  & 879  & 1193  & 132  & 119  & 260  & 888  & 774  & 930  & 553  & 1374  & 611  & 520  & 406  & 805  & 0  & \textbf{1796}  & 0  & 333  & 1119  & 344  & 1123  & 613  & 60 \\
        
        \textbf{COM-H}  & 480  & 744  & 285  & 144  & 172  & 435  & 412  & 576  & 392  & 696  & 353  & 396  & 251  & 498  & 0  & 0  & \textbf{856}  & 129  & 479  & 248  & 436  & 366  & 54 \\
        
        \textbf{SIZ-S}  & 460  & 469  & 30  & 63  & 39  & 520  & 402  & 378  & 180  & 556  & 308  & 198  & 105  & 518  & 651  & 333  & 129  & \textbf{1113}  & 0  & 0  & 874  & 238  & 1 \\
        
        \textbf{SIZ-M}  & 1255  & {\color{black}{1593}}  & 235  & 128  & 479  & 1320  & 1176  & 1165  & 667  & {\color{black}{1982}}  & 923  & 826  & 514  & 856  & 861  & 1119  & 479  & 0  & \textbf{2459}  & 0  & 1338  & 990  & 131 \\
        
        \textbf{SIZ-B}  & 145  & 612  & 160  & 111  & 69  & 197  & 162  & 545  & 426  & 704  & 160  & 217  & 296  & 356  & 336  & 344  & 248  & 0  & 0  & \textbf{928}  & 472  & 384  & 72 \\ 
        
        \textbf{LEN-S}  & 1270  & 1478  & 233  & 144  & 194  & 1179  & 958  & 1185  & 645  & {\color{black}{1870}}  & 760  & 549  & 478  & 1218  & 1125  & 1123  & 436  & 874  & 1338  & 472  & \textbf{2684}  & 0  & 0 \\ 
        
        \textbf{LEN-M}  & 551  & 1069  & 178  & 122  & 302  & 749  & 674  & 823  & 565  & 1242  & 532  & 599  & 399  & 486  & 633  & 613  & 366  & 238  & 990  & 384  & 0  & \textbf{1612}  & 0 \\ 
        
        \textbf{LEN-L}  & 39  & 127  & 14  & 36  & 91  & 109  & 108  & 80  & 63  & 130  & 99  & 93  & 38  & 26  & 90  & 60  & 54  & 1  & 131  & 72  & 0  & 0  & \textbf{204} \\
        \hline 
        
		\end{tabular}
		}
	\end{center}
    }
\end{table*}

\subsection{\color{black}Language Annotation for WebUAV-3M}
\label{sec:Word cloud}

\begin{figure}[h]   	
        \centering\centerline{\includegraphics[width=1.01\linewidth]{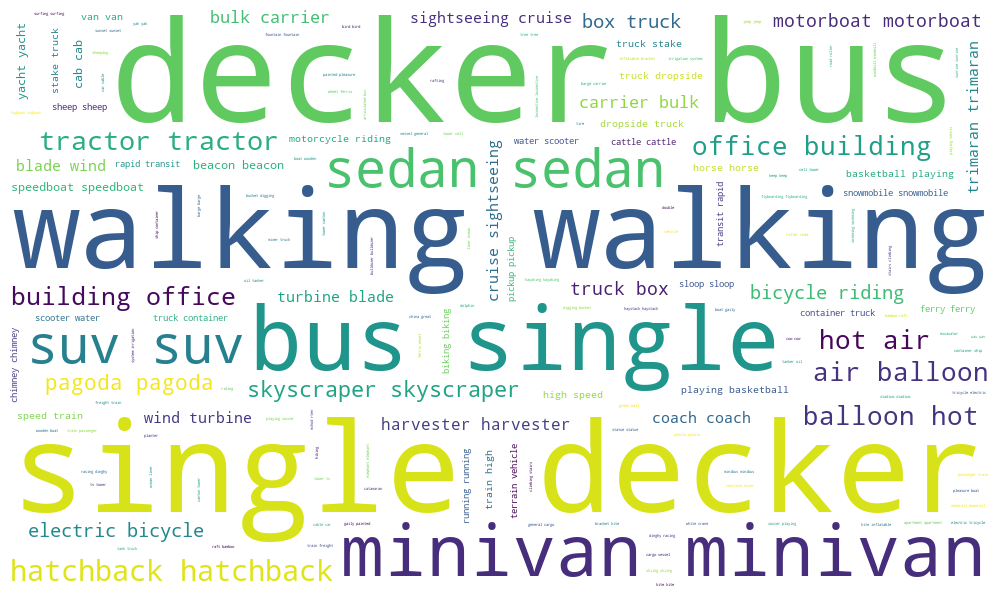}}
	\caption{Word cloud of the natural language specifications in our WebUAV-3M dataset.}
	\label{fig:nlp}
\end{figure}

As shown in Fig.~\ref{fig:nlp}, the word cloud demonstrates the English words that occur frequently in the natural language specifications in the proposed WebUAV-3M dataset. 

{\color{black} We assembled a professional data annotation team (approximately ten people) from a qualified data company for the language annotation task. For each video sequence, we ask the data annotation team to provide an English sentence that describes the class name (target class or motion class), position (relative location), attribute, behavior, and surroundings of the target. Considering the complexity (\eg tiny targets, similar distractors and low light) of UAV tracking videos, it is hard to accurately describe the object in each video sequence using one sentence (only plain text) in our WebUAV-3M dataset. Fortunately, we noticed a recent groundbreaking work~\cite{chen2021pix2seq} proposed to use a sentence containing bounding box coordinates and class labels to describe objects for the object detection (localization) task. Following~\cite{chen2021pix2seq}, we allow the annotator to add the center position $(x_{1}, y_{1})$ of the target bounding box in the sentence as a weak supervision signal when it is hard to describe the position (relative location) of the target in plain text. The proportion of these sentences is approximately $5\%$ in our WebUAV-3M dataset. Some examples are shown in Fig.~\ref{fig:Example_sequences} (\eg row 3 column 2, row 4 column 1, row 5 column 2, and row 6 column 2). To verify the correctness of the natural language specifications, the authors performed three times validations for language annotations (see Table~\ref{tab:quality_control}).}

\subsection{\color{black}Audio Annotation for WebUAV-3M}
\label{sec:Balabolka}

{\color{black}We use a free and open-source text-to-speech software (Balabolka v2.15.0.818) for audio annotation. Balabolka is a powerful software based on various versions of Microsoft Speech API, which can convert the input text into an audio file. The output audio formats supported by Balabolka include WAV, MP3, OGG, WMA, \etc. Users can easily adjust the rate, pitch, and volume of the output audio file. In this work, we apply the Microsoft Speech API 5 and the Microsoft Speech Platform text-to-speech engines. Specifically, we use the voices of \emph{Microsoft Zira Desktop} and \emph{Microsoft David Desktop} to generate female and male audio descriptions, respectively. The whole audio annotation is done on Microsoft Windows 10 operating system by the data annotation team. After obtaining the audio files (MP3 format), the authors carefully verify and revise each audio file.}

\begin{figure}[t]   
\centering\centerline{\includegraphics[width=1.0\linewidth]{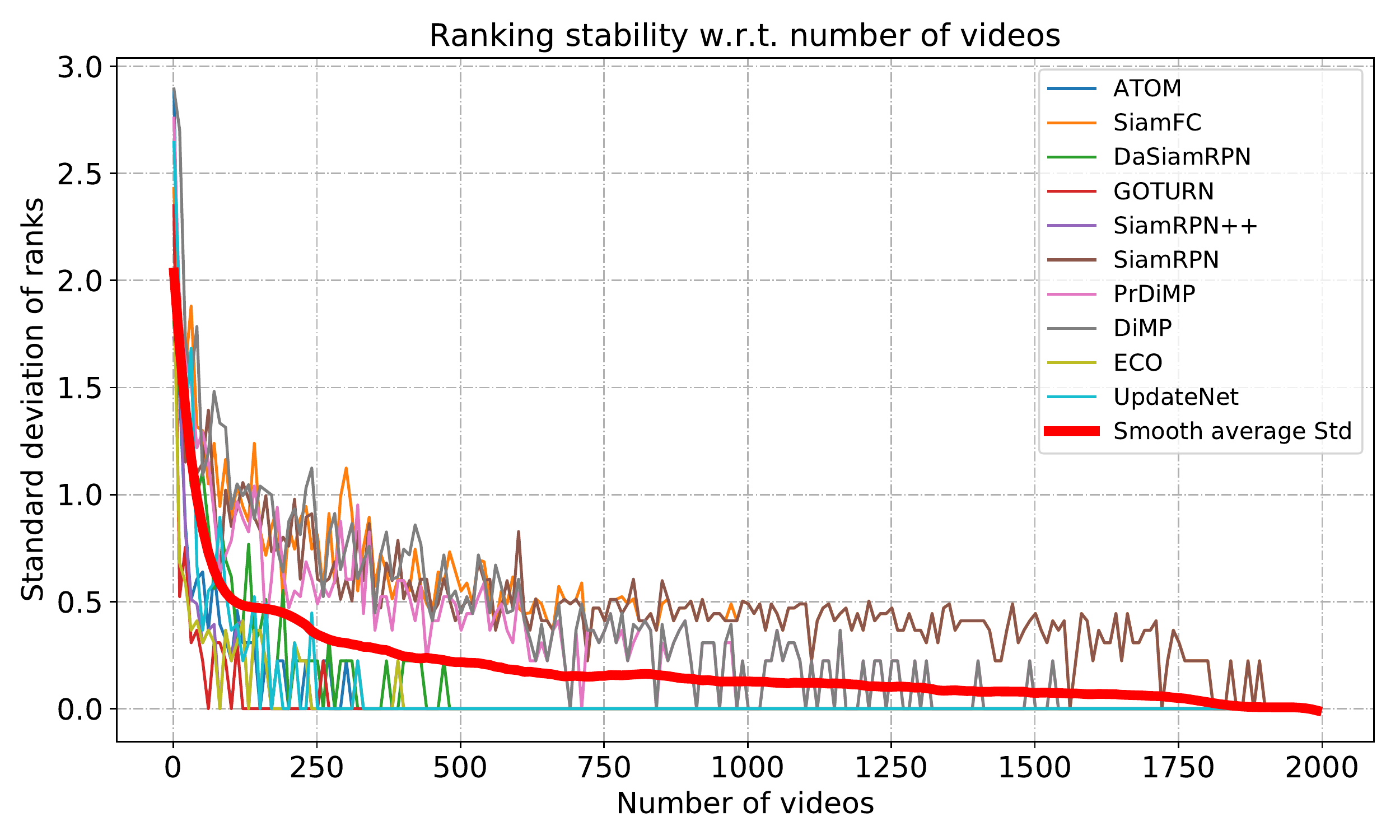}}
	\caption{Ranking stability with regard to the number of videos. The smooth average Std of the ranks significantly decreases as the number of videos increases, indicating improved evaluation stability.}
	\label{fig:ranking_stability}
\end{figure}

\subsection{Dataset Splitting Strategy}
\label{sec:splitting strategy}
The details of our dataset splitting strategy are as follows.

\textbf{Step 1:} We run ten deep tracking algorithms on the complete WebUAV-3M dataset, sort each video in ascending order by their accuracy scores, and select the top 2000 videos as candidates for the test set. This processing approach ensures that the test set can critically evaluate tracking performance by mainly comprising moderately complex and complicated videos.

\textbf{Step 2:} We conduct a stability experiment to find the best number of test sets for achieving reliable and stable evaluation results. Specifically, we randomly select a certain number of videos from the candidate test set as samples and repeat the sampling process twenty times with the same number of videos. Then, we run ten trackers on each sample and rank them according to their accuracy scores. The ranking stability, \ie the standard deviation (Std) of the algorithms’ ranks, is used as the evaluation stability indicator. We change the number of test videos from 1 to 2000, with a step size of 10.

As shown in Fig.~\ref{fig:ranking_stability}, as the number of videos increases, the standard deviation of the algorithm rankings gradually decreases, indicating that the evaluation process tends to stabilize. To balance evaluation stability and efficiency, we take $780$ as the number of videos in the final test set.

\textbf{Step 3:} We randomly select 1500 videos from the 2000 total videos (acquired from step 1) as candidates for the test set in this step. First, to evaluate the generalizability/transferability of different tracking algorithms to various seen and unseen target classes and motion classes, we ensure that the categories (object and motion classes) of the test set and training set have as little overlap as possible. Then, we prefer to select videos from the categories with small numbers of videos to make the test set contain many classes. Finally, we sample evenly from each category to ensure balance regarding the number of videos to obtain 1500 candidate videos for the test set.

\textbf{Step 4:} We randomly select 780 videos from the above 1500 candidate videos as the final test set. Then, we randomly select 200 videos from the remaining videos for the validation set and the remaining 520 videos as the training set.

\section{More discussion about SATA}
\label{sec:more_discussion_about_SATA}

{\color{black}SATA is an efficient and interactive semi-automatic annotation tool. We propose to verify the effectiveness of SATA from two aspects of \emph{annotation time} and \emph{annotation accuracy}. 

For annotation time, we compare the times spent per bounding box using our SATA with the manual annotations and other semi-automatic tools on video sequences of three difficulty levels: easy, medium, and hard (see Table~\ref{tab:time}). Specifically, we randomly select the same number of video sequences (\ie 10) from each difficulty level to construct three video groups. Then, we ask several skilled annotators to label the three groups of video sequences independently and report their average time spent on each video group. Results demonstrate that our SATA approach has a significant advantage regarding average time consumption for each bounding box annotation (see Table~\ref{tab:time}). We find that using Labelme~\cite{russell2008labelme} and VoTT\footnote{https://github.com/microsoft/VoTT} for manual annotation, each bounding box takes 15.81 seconds and 8.05 seconds, respectively. The semi-automatic labeling of each bounding box using ViTBAT~\cite{biresaw2016vitbat} and CVAT\footnote{https://github.com/openvinotoolkit/cvat} takes 4.68 seconds and 3.86 seconds, respectively. Encouragingly, the proposed SATA pipeline further reduces the time spent per bounding box to only 2.99 seconds. 

For annotation accuracy, the most accurate comparison is to annotate the entire dataset using both semi-automatical and manual methods, then compare results. However, it is impractical due to the massive amount of data. We believe that accurate data annotation still requires human annotators' intervention to ensure the annotation's correctness. SATA achieves an efficient combination of tools and human supervision to achieve a double improvement in speed and quality. Using a prediction model to replace the operation of drawing frames, humans only need to supervise and correct the quality of the annotations. The change of human roles not only increases the overall labeling efficiency but also makes the human focus more on accurate corrections. With the confidence of our SATA pipeline, we provide some \emph{indirect results} demonstrating that SATA can be even more precise than manual annotations. \textbf{First}, in Fig.~\ref{fig:Pipeline}, our interactive annotation pipeline can use advanced deep tracking models (we employ SiamRPN++~\cite{LiWWZXY19} in this work) to obtain tight bounding boxes containing objects of interest in \emph{short segments} (\eg a few to dozens of frames) of a video sequence. Then, annotators perform real-time manual checking and error fixing in an interactive manner, which can greatly correct and reduce annotation errors. In this way, SATA can generate accurate bounding boxes; some examples are shown in Fig.~\ref{fig:accurate_annotations}. \textbf{Second}, in subsection~\ref{sec:data_quality_validation}, we leverage data quality verification to prove the effectiveness of SATA quantitatively. We retrain several deep trackers on the training set annotated with SATA and achieve consistent performance gains on the test set, indicating the high quality of the annotations. We further explore the impact of the number of data annotations. Excluding the influence of the domain gap, the above consistent experimental results can verify the accuracy of our annotations, \ie indirectly proving the effectiveness of SATA.

In contrast, if we only perform manual annotation, there may be a lot of human errors. With the increase in the number of annotated videos and the time of annotation, the physical fatigue of the annotator increases, and the annotation quality will drop sharply. More seriously, annotators often cannot discover annotation errors in real-time. Once the labeling error occurs, it is hard to find the labeling error of a few frames hidden in the middle of the video sequence by manual checking. In this case, more manual checking costs are usually required to reduce labeling errors.

In general, the advantages of our interactive semi-automatic SATA pipeline compared to full manual annotations are: generating accurate bounding boxes in short segments by an advanced tracking model, real-time manual checking, and simultaneous error fixing. By automatically generating high-quality bounding boxes and real-time human checking, we will significantly reduce the total annotation time and human labor. Finally, multiple rounds of manual verification (see Table~\ref{tab:quality_control}) can further reduce the labeling error and ensure the effectiveness of SATA.

Next, we discuss the potential annotation bias regarding SATA due to using an off-the-shelf deep tracking model (\ie SiamRPN++~\cite{LiWWZXY19}). The assumption for semi-automatic labeling is that within a short segment, the changes in the target itself (\eg appearance, size, scale, and motion) and the surrounding environment (\eg illumination and background) are relatively small; current advanced trackers, such as~\cite{LiWWZXY19}, can achieve accurate target tracking. Based on the above assumption, we make some efforts to reduce the potential bias of automatic tracking. 

\textbf{First}, the proposed interactive semi-automatic annotation pipeline SATA allows the annotator to divide the video sequence into multiple short segments in a human-computer interaction manner. Therefore, the annotator can use the tracker to generate accurate bounding boxes in each short segment automatically. When low-quality tracking results appear, the annotator can conveniently re-select the starting point for semi-automatic labeling. \textbf{Second}, \cite{LiWWZXY19} is a powerful deep tracking model with good adaptability to target size, appearance changes, \etc. In short segments, it can achieve stable and accurate tracking. Since the tracking speed of~\cite{LiWWZXY19} is fast, it helps us improve the annotation speed. However, in some complex situations (\eg long-term occlusion, fast motion, and similar distractors), we still need to perform a manual intervention to reduce the labeling bias. \textbf{Third}, to further reduce the labeling bias, we design a strict quality control process (see Table~\ref{tab:quality_control}) for the whole data construction, including data annotation and verification.}

\begin{table*}[t]
\color{black}
\renewcommand\arraystretch{1.0}
\caption{
\color{black}Performance comparison on WebUAV-3M test set between tracking via joint natural language and bounding box (NL+BBox), and tracking via bounding box only (BBox). We report the results of the latest state-of-the-art vision-language trackers (VLT$_{\rm SCAR}$ and VLT$_{\rm TT}$) and their corresponding baseline visual trackers (SiamCAR and TransT).}
\label{table:vision_language_tracking}
\setlength{\tabcolsep}{4.92mm}{
\begin{tabular}{|c|c|c|c|c|c|c|c|}

\hline
\textbf{Type} &  \textbf{Tracker} & \textbf{Publication} &  \textbf{Pre} & \textbf{nPre}
  &  \textbf{AUC} &  \textbf{cAUC} & \textbf{mAcc} \\		 
\hline

\multirow{2}{*}{\textbf{Tracking via BBox}} & SiamCAR~\cite{guo2020siamcar}  & CVPR-2020 & 0.642 &  0.512 &  0.412 & 0.378  & 0.415 \\

 & TransT~\cite{chen2021transformer} & CVPR-2021  &  0.618 &  0.509 &  0.448 & 0.422  & 0.453 \\
\hline

\multirow{8}{*}{\textbf{Tracking via NL+BBox}} 
& VLT$_{\rm SCAR}$-\uppercase\expandafter{\romannumeral4}-got10k~\cite{guo2022divert} & NeurIPS-2022 &  0.501 &  0.386 &  0.358 & 0.345  & 0.361 \\

& VLT$_{\rm SCAR}$-\uppercase\expandafter{\romannumeral3}-lasotext~\cite{guo2022divert} & NeurIPS-2022 &  0.584 &  0.475 &  0.431 & 0.420  & 0.435  \\

& VLT$_{\rm SCAR}$-\uppercase\expandafter{\romannumeral2}-lasot-tnl2k~\cite{guo2022divert} & NeurIPS-2022 &  0.594 &  0.480 &  0.440 & 0.428  & 0.444  \\

& 
VLT$_{\rm SCAR}$-\uppercase\expandafter{\romannumeral1}-otb99~\cite{guo2022divert}  & NeurIPS-2022  &  0.600 &  0.490 &  0.449 & 0.438  & 0.453  \\

& VLT$_{\rm TT}$-\uppercase\expandafter{\romannumeral4}-lasotext~\cite{guo2022divert} & NeurIPS-2022  &  0.520 &  0.406 &  0.384 & 0.371  & 0.385   \\

& VLT$_{\rm TT}$-\uppercase\expandafter{\romannumeral3}-got10k~\cite{guo2022divert} & NeurIPS-2022  &  0.531 &  0.414 &  0.388 & 0.375  & 0.390  \\

& VLT$_{\rm TT}$-\uppercase\expandafter{\romannumeral2}-lasot-otb99~\cite{guo2022divert} & NeurIPS-2022  &  0.620 &  0.503 &  0.466 & 0.457  & 0.472  \\

& VLT$_{\rm TT}$-\uppercase\expandafter{\romannumeral1}-tnl2k~\cite{guo2022divert} & NeurIPS-2022  &  0.638 &  0.511 &  0.470 & 0.460  & 0.475  \\
\hline
\end{tabular}
}
\end{table*}

\begin{figure*}[h]
    \centering
    \subfloat{\includegraphics[width =1.025\columnwidth]{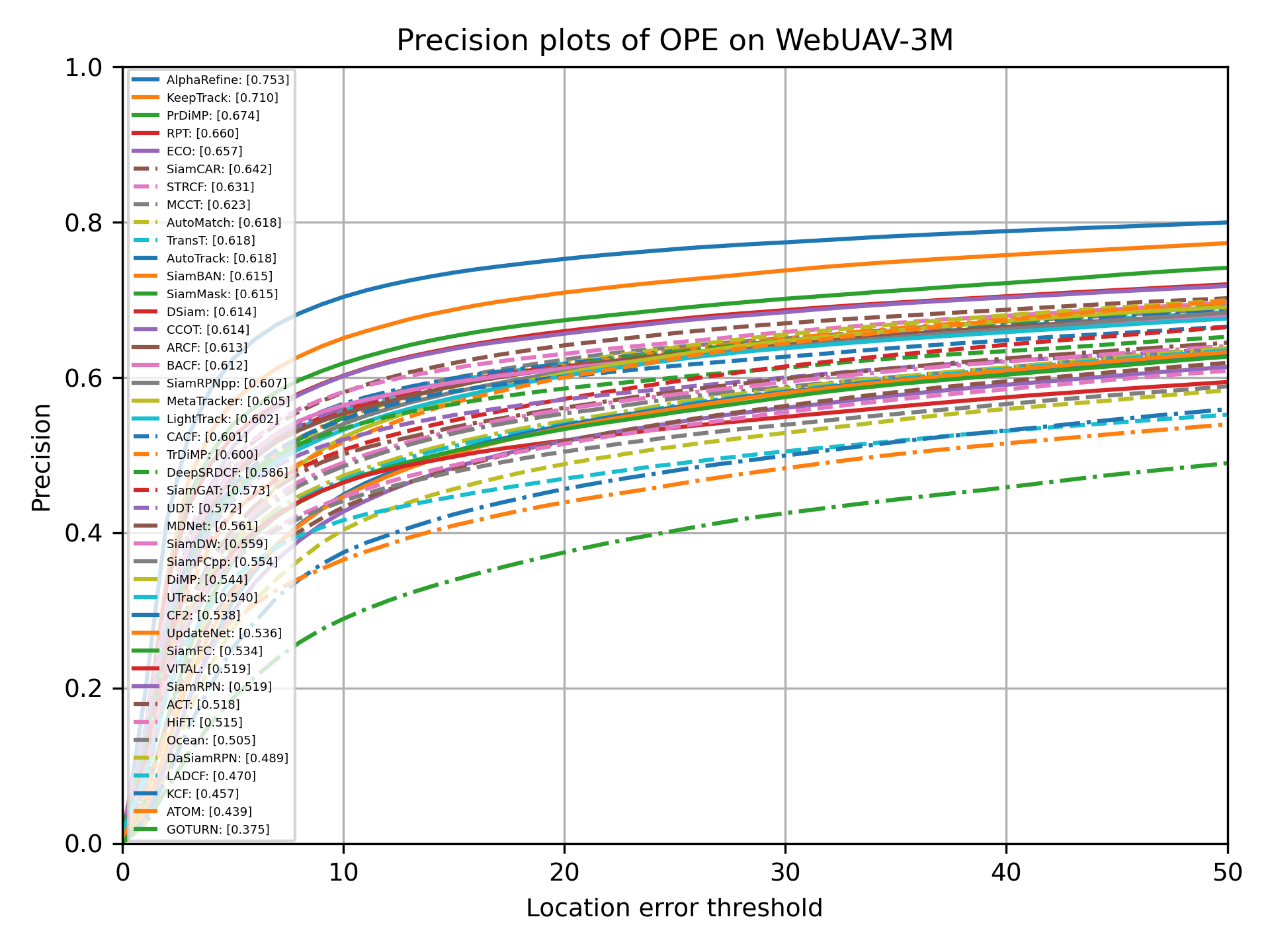}}
    ~
    \subfloat{\includegraphics[width =1.025\columnwidth]{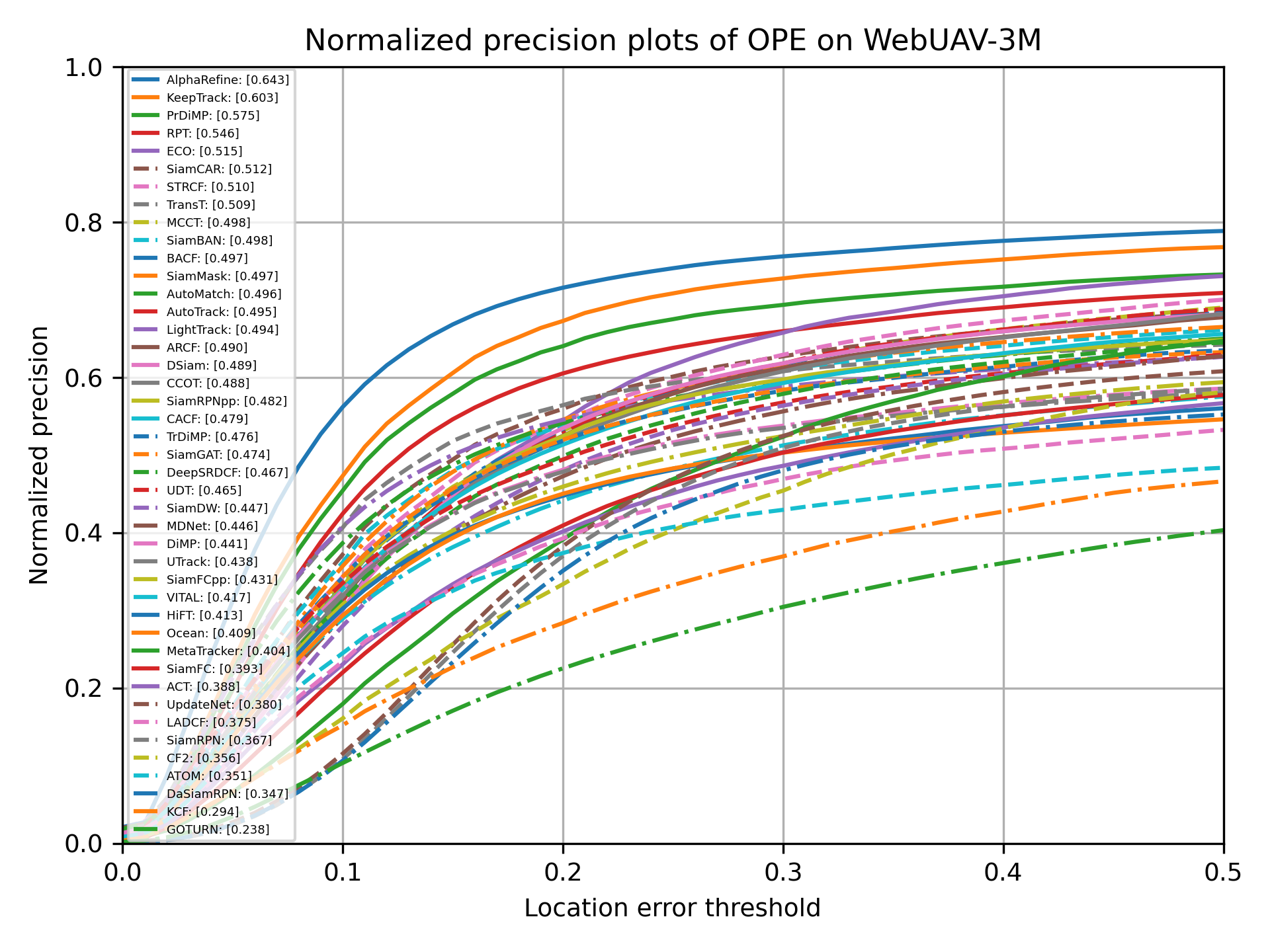}}
    \\
    \subfloat{\includegraphics[width =1.025\columnwidth]{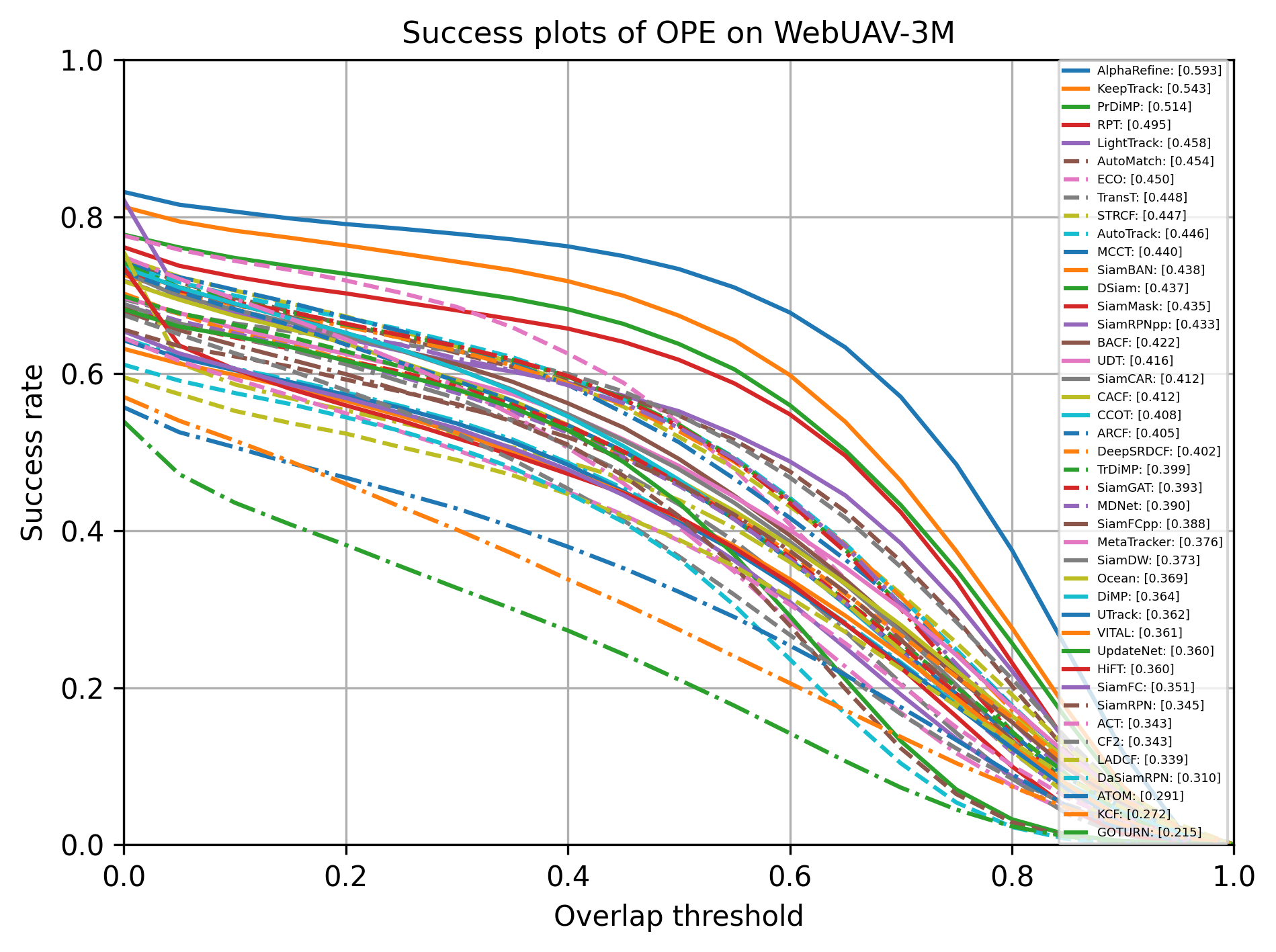}}
    ~
    \subfloat{\includegraphics[width =1.025\columnwidth]{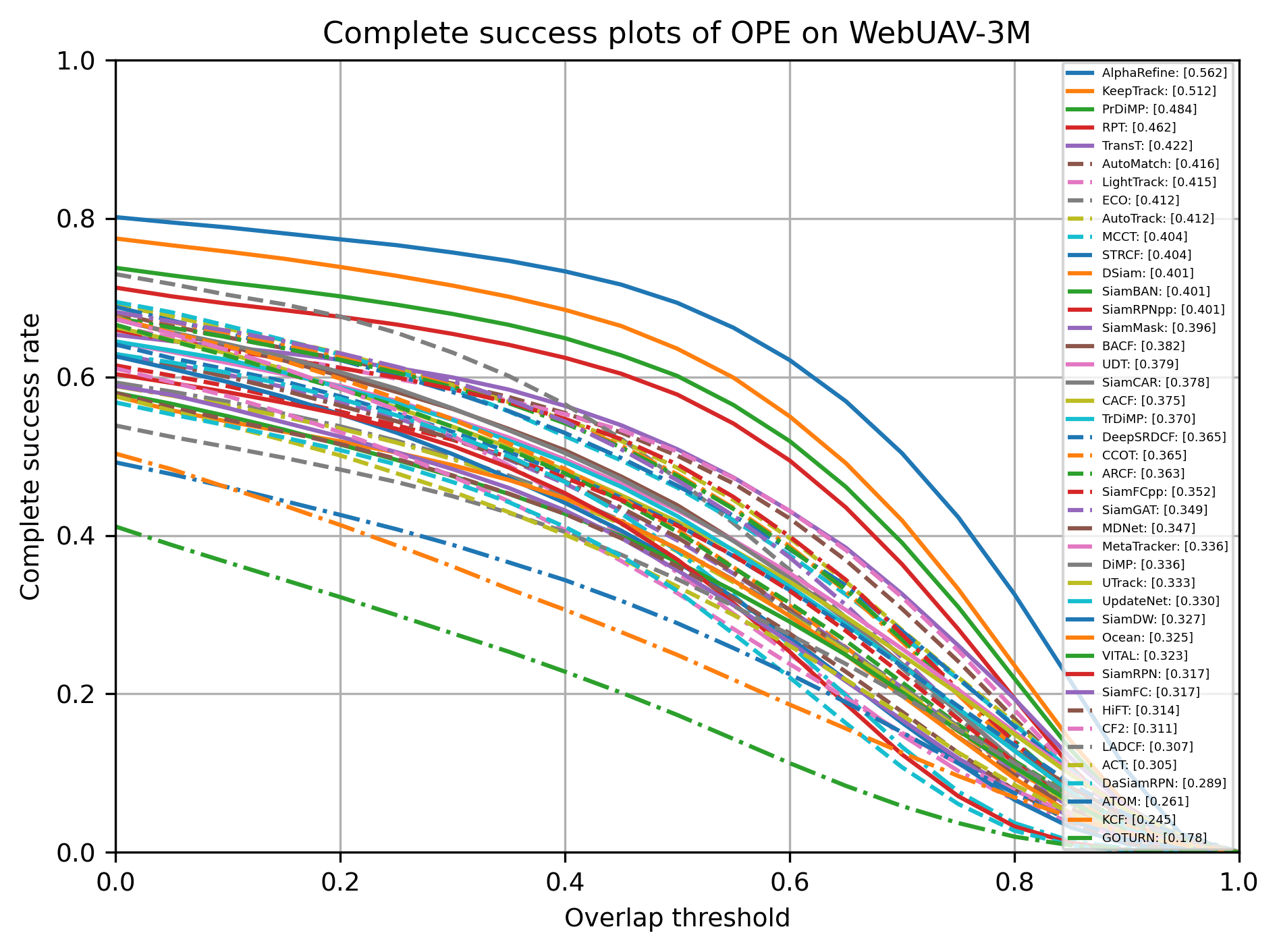}}
\caption{Benchmark results of 43 baseline trackers on WebUAV-3M test set using a precision plot, normalized precision plot, success plot, and complete success plot. Best viewed by zooming in.}
\label{fig:benchmark_results}
\end{figure*}

\section{Additional Results}
\label{sec:additional_results}

\subsection{\color{black}Results of Tracking by Joint Language and Bounding Box}
\label{sec:vision_language_tracking}
{\color{black}To advance research on multi-modal tracking, we evaluate two recent vision-language trackers (VLT$_{\rm SCAR}$~\cite{guo2022divert} and VLT$_{\rm TT}$~\cite{guo2022divert}) and their baseline visual trackers (SiamCAR and TransT) on the WebUAV-3M test set. In~\cite{guo2022divert}, Guo \etal~propose to learn adaptive vision-language representations using modality mixer and asymmetrical ConvNet search. They first use the training splits of COCO, ImageNet VID, ImageNet DET, Youtube-BB, GOT-10k, LaSOT, and TNL2K to search a modeling structure for vision-language tracking with a supernet~\cite{guo2022divert}. Then, they follow the training strategy of the baselines SiamCAR and TransT to optimize VLT$_{\rm SCAR}$ and VLT$_{\rm TT}$, respectively. In the process of optimization, the language embeddings of the datasets (\ie COCO, ImageNet VID, ImageNet DET, and Youtube-BB) without language annotations are replaced by 0-tensors or pooled visual features. For more details about VLT$_{\rm SCAR}$ and VLT$_{\rm TT}$, we kindly refer readers to~\cite{guo2022divert} and their project page\footnote{https://github.com/JudasDie/SOTS}. In Table~\ref{table:vision_language_tracking}, we comprehensively evaluate 8 tracking models (\ie VLT$_{\rm SCAR}$-\uppercase\expandafter{\romannumeral1}-otb99, VLT$_{\rm SCAR}$-\uppercase\expandafter{\romannumeral2}-lasot-tnl2k, VLT$_{\rm SCAR}$-\uppercase\expandafter{\romannumeral3}-lasotext, VLT$_{\rm SCAR}$-\uppercase\expandafter{\romannumeral4}-got10k, VLT$_{\rm TT}$-\uppercase\expandafter{\romannumeral1}-tnl2k, VLT$_{\rm TT}$-\uppercase\expandafter{\romannumeral2}-lasot-otb99, VLT$_{\rm TT}$-\uppercase\expandafter{\romannumeral3}-got10k, and VLT$_{\rm TT}$-\uppercase\expandafter{\romannumeral4}-lasotext) on the WebUAV-3M test set using the pretrained network weights provided by the original authors.

We have the following three observations. \textbf{First}, comparing SiamCAR with TransT, the Siamese network-based tracker (SiamCAR) is more accurate for the target center prediction (higher Pre and nPre scores), but the CNN-Transformer-based tracker (TransT) can achieve better target state estimation (higher AUC, cAUC, and mAcc scores). \textbf{Second}, adding natural language annotations does not always improve tracking performance. For example, the improved version VLT$_{\rm TT}$-\uppercase\expandafter{\romannumeral1}-tnl2k outperformers the baseline TransT by $2.0\%$, $0.2\%$, $2.2\%$, $3.8\%$, and $2.2\%$ in terms of Pre, nPre, AUC, cAUC, and mAcc scores, respectively. While the tracking performance of the advanced VLT$_{\rm SCAR}$-\uppercase\expandafter{\romannumeral1}-otb99 degrades (\eg $0.642\!\rightarrow \!0.600$ and $0.512\!\rightarrow\!0.490$ in terms of Pre and nPre scores, respectively). We argue that the high-level semantics in natural language can provide auxiliary help for tracking, but vision-language trackers can hardly benefit from limited multi-modal data. This motivates us to construct the large-scale dataset WebUAV-3M with visual box annotations, natural language specifications, and audio descriptions. \textbf{Third}, the performance of recent vision-language trackers falls far behind the state-of-the-art visual trackers (see Table~\ref{tab:Overall_results} and Table~\ref{table:vision_language_tracking}). For example on our WebUAV-3M test set, the gaps between the most advanced vision-language tracker (VLT$_{\rm TT}$-\uppercase\expandafter{\romannumeral1}-tnl2k) and the visual tracker (AlphaRefine) are absolute $11.5\%$, $13.2\%$, $12.3\%$, $10.2\%$, and $12.7\%$ in terms of Pre, nPre, {AUC}, {cAUC}, and {mAcc} scores, respectively. Therefore, we propose the  large-scale multi-modal dataset WebUAV-3M and expect that the community will pay more attention to vision-language tracking.
}

\subsection{Overall Performance}
\label{sec:overall_performance_appendix}
The precision, normalized precision, success, and complete success curves of all baseline trackers are shown in Fig.~\ref{fig:benchmark_results}, ranked by the Pre, nPre, AUC, and cAUC scores, respectively. Note that SiamRPNpp and SiamFCpp refer to SiamRPN++ and SiamFC++, respectively.

\subsection{Attribute-Based Performance}
\label{sec:Attribute-based Performance}

\begin{figure*}[t]
    \centering
    \subfloat{\includegraphics[width =0.5\columnwidth]{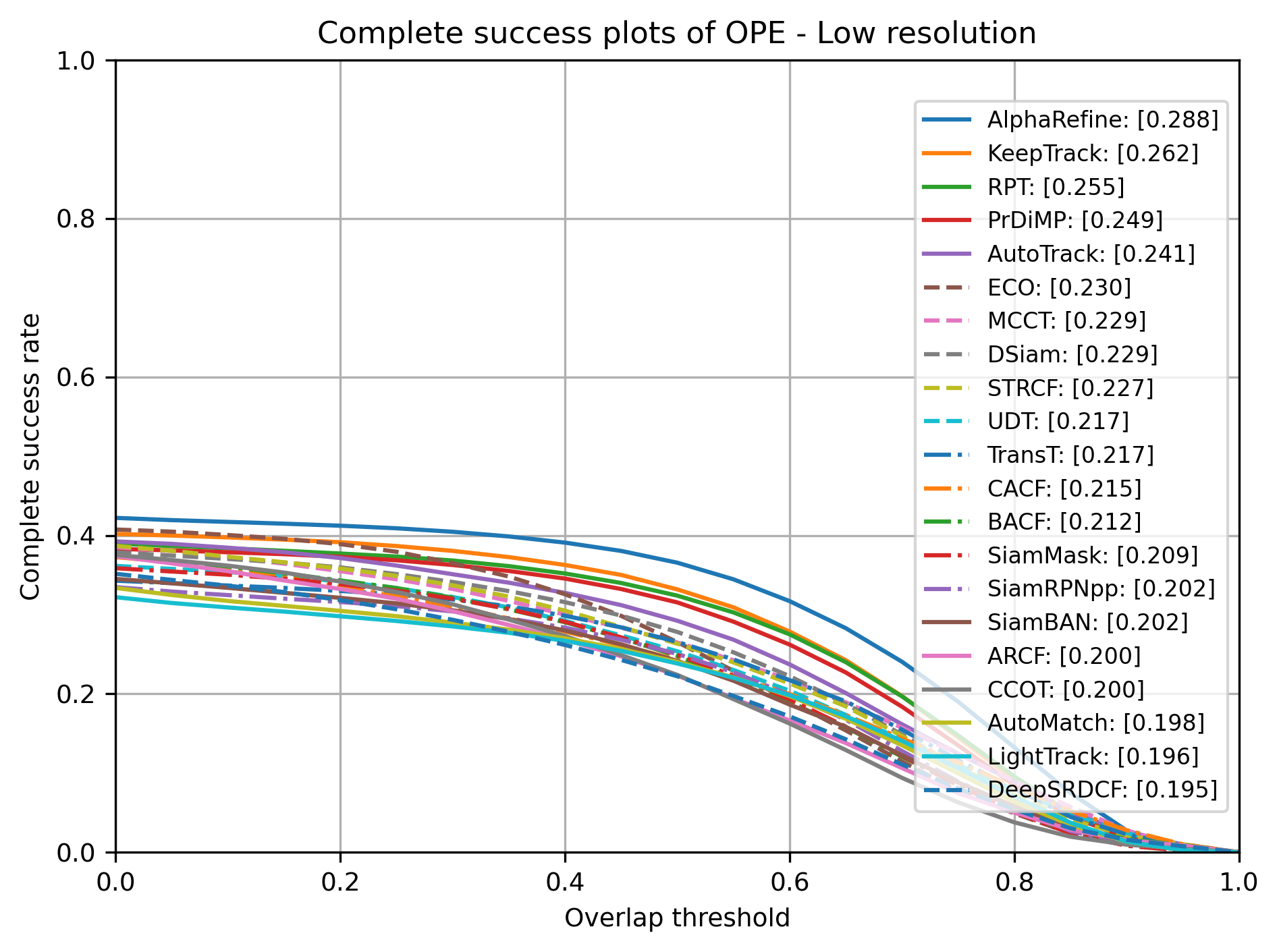}}
    ~
    \subfloat{\includegraphics[width =0.5\columnwidth]{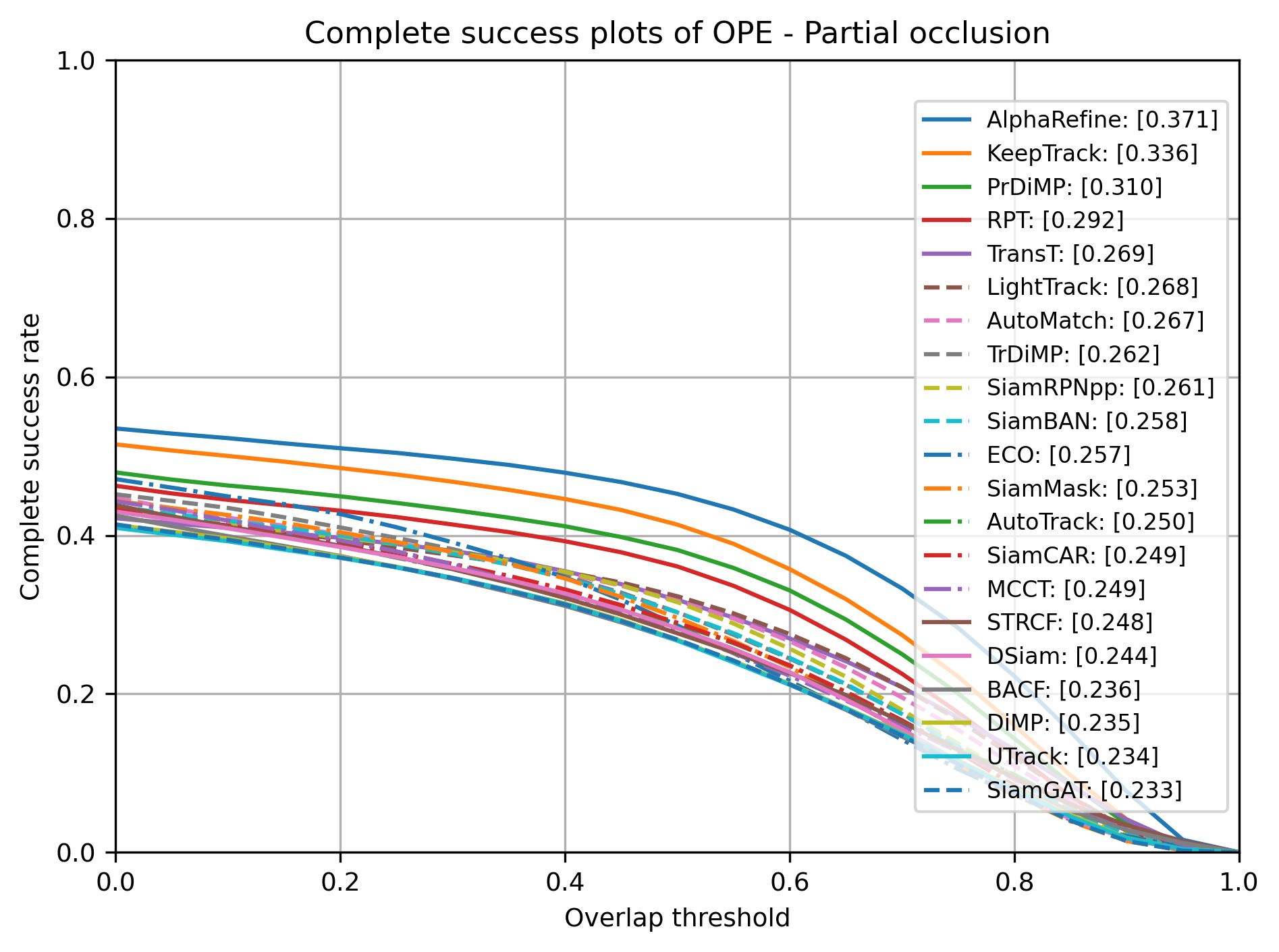}}
    ~
    \subfloat{\includegraphics[width =0.5\columnwidth]{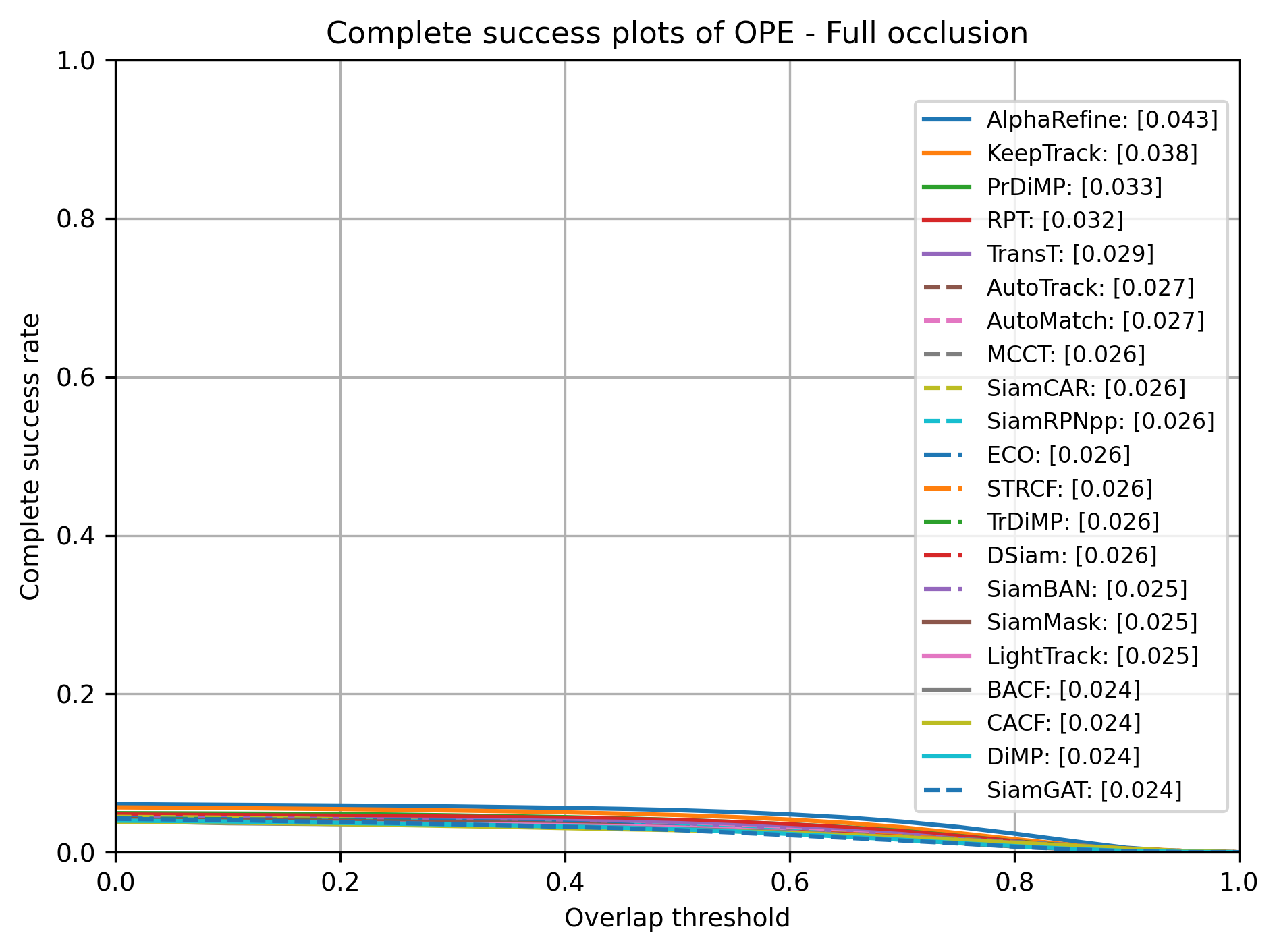}}
    ~
    \subfloat{\includegraphics[width =0.5\columnwidth]{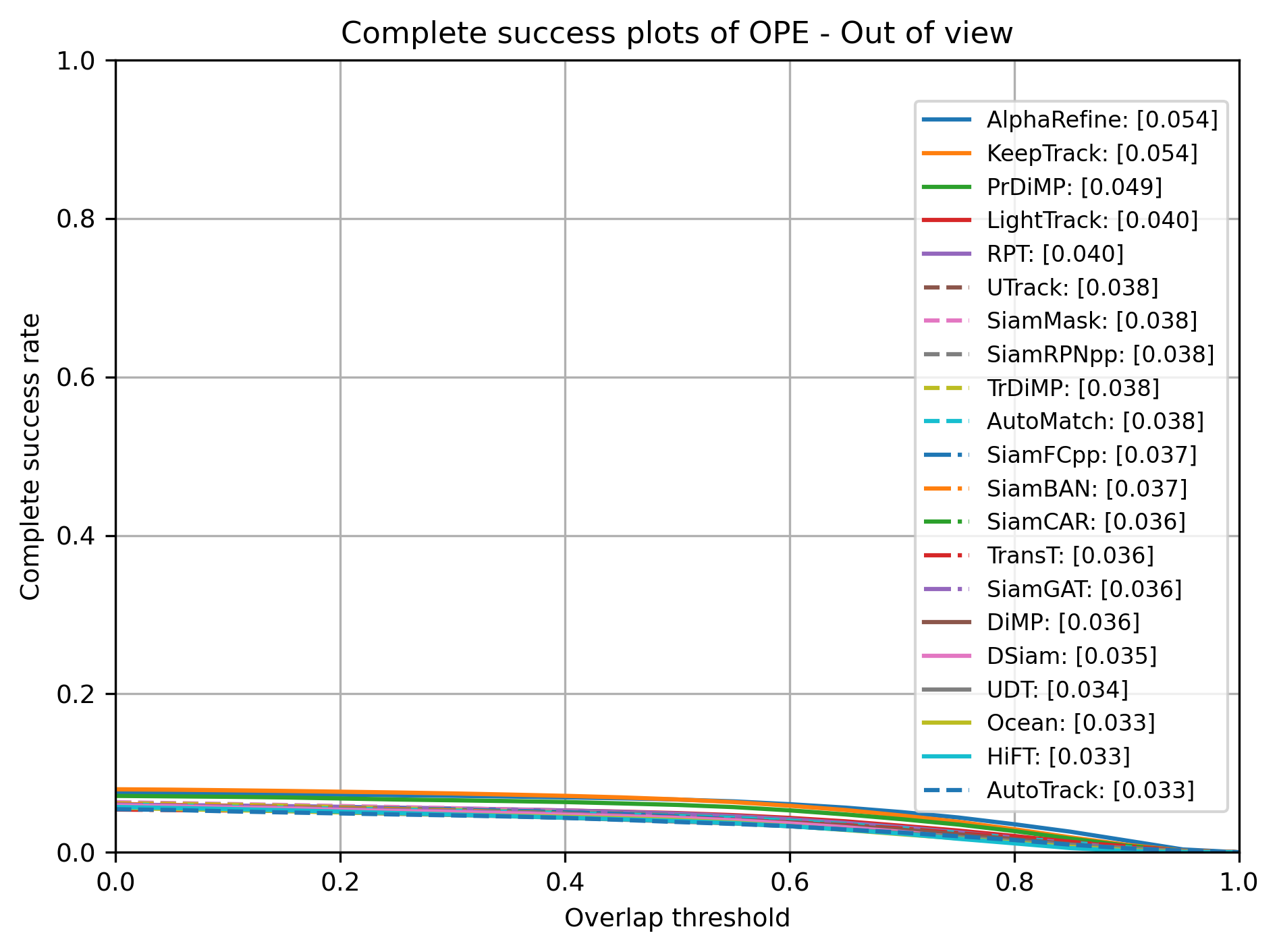}}\\
    
    \subfloat{\includegraphics[width =0.5\columnwidth]{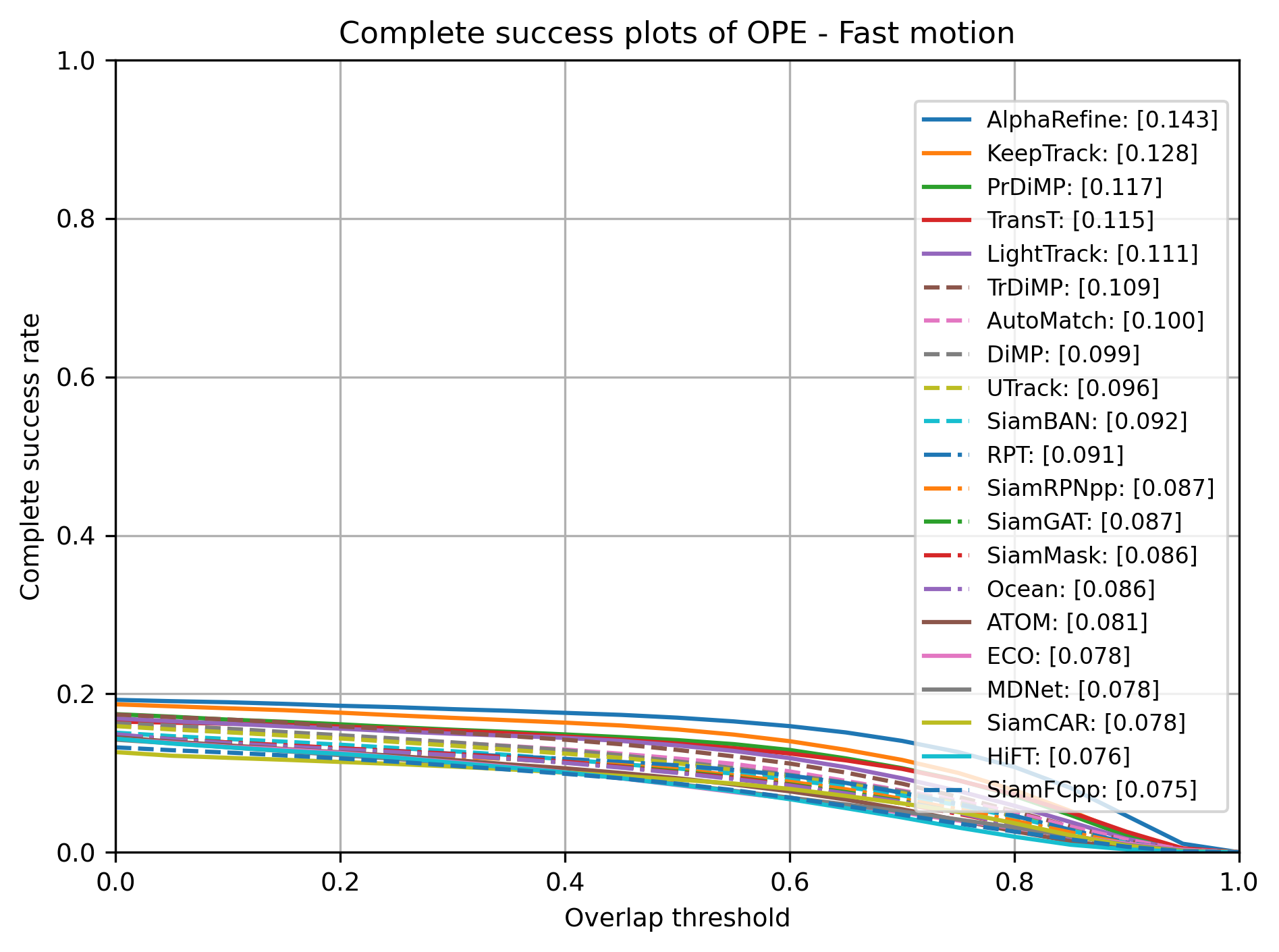}}
    ~
    \subfloat{\includegraphics[width =0.5\columnwidth]{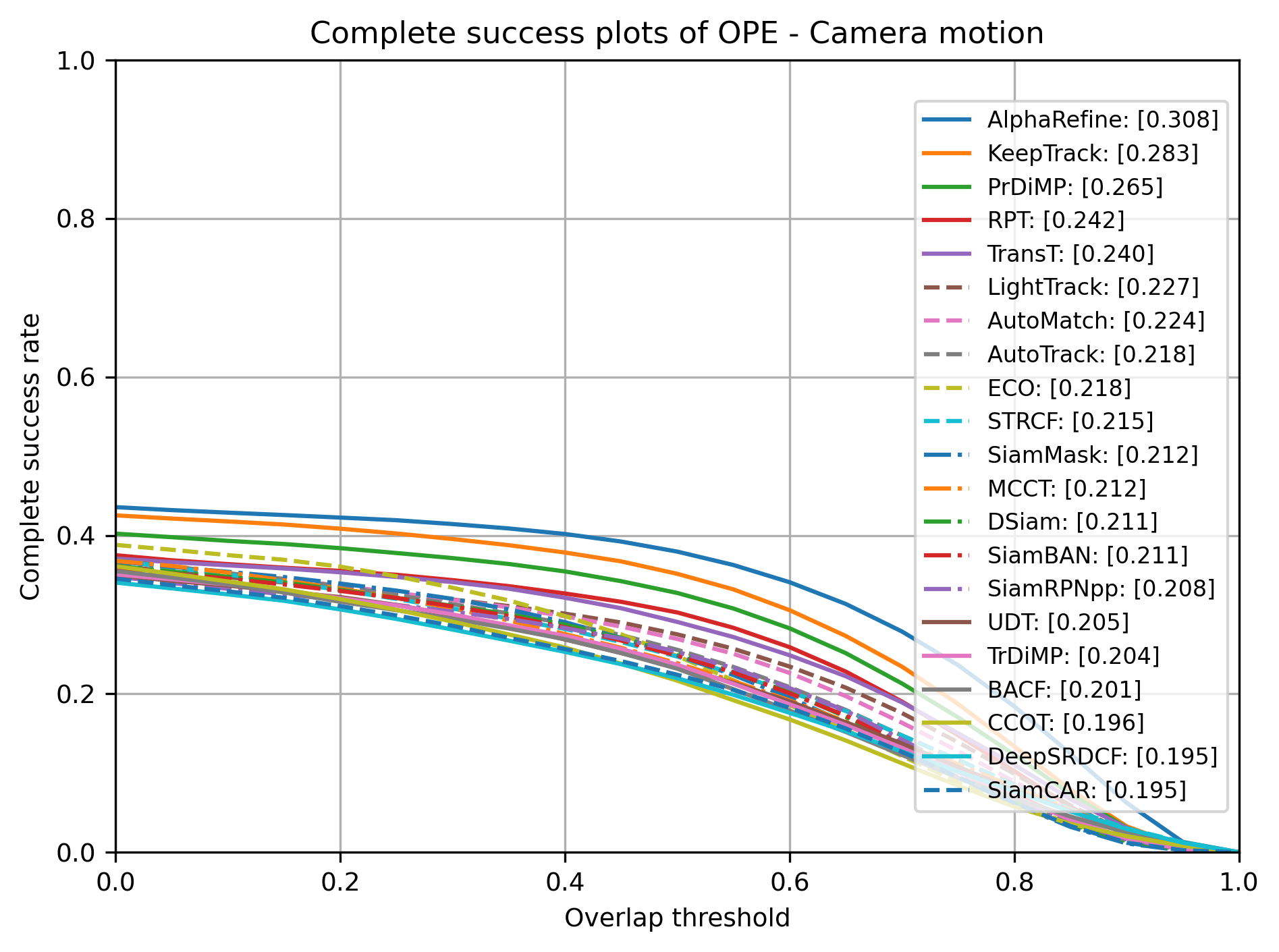}}
    ~
    \subfloat{\includegraphics[width =0.5\columnwidth]{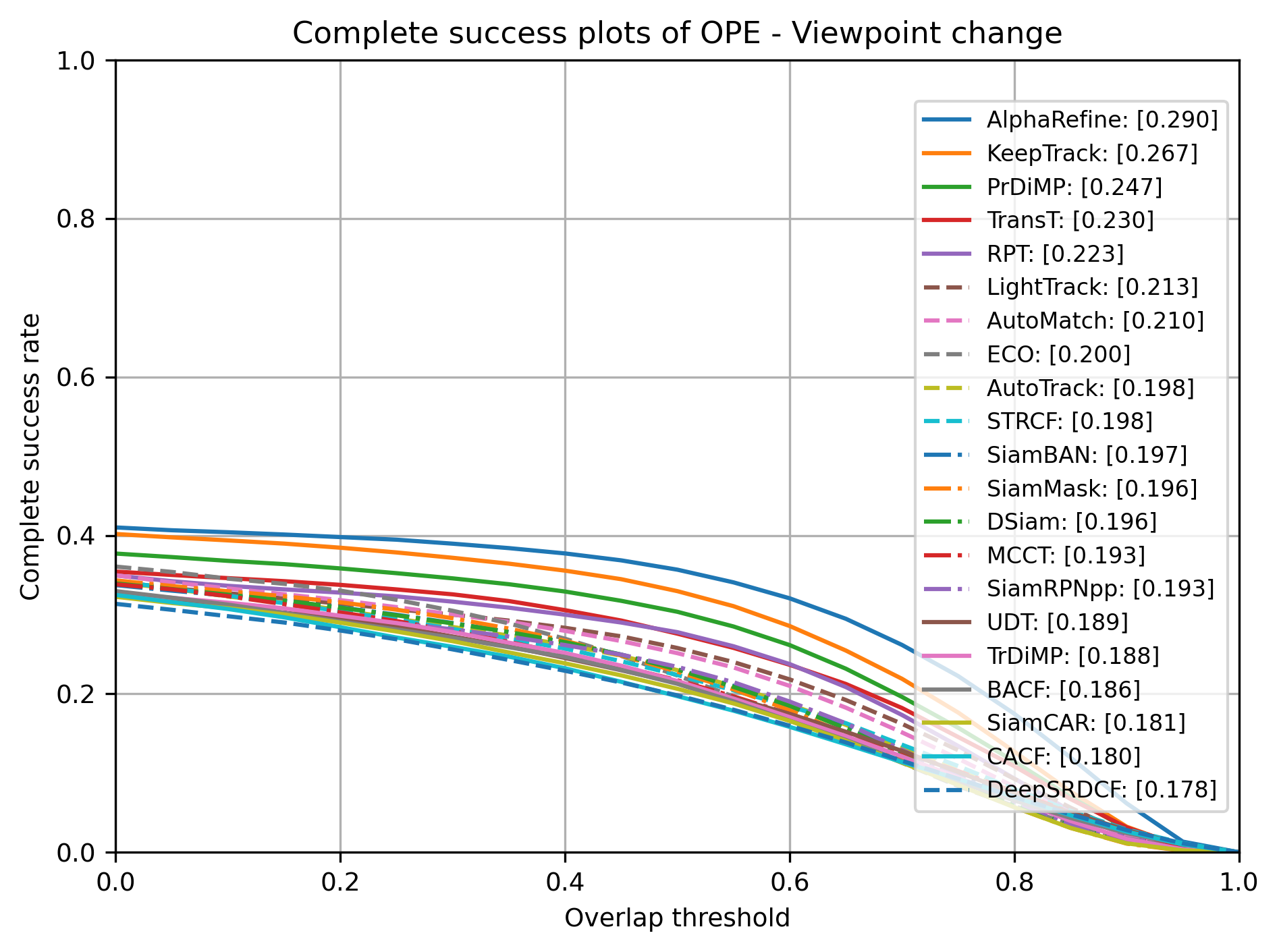}}
    ~
    \subfloat{\includegraphics[width =0.5\columnwidth]{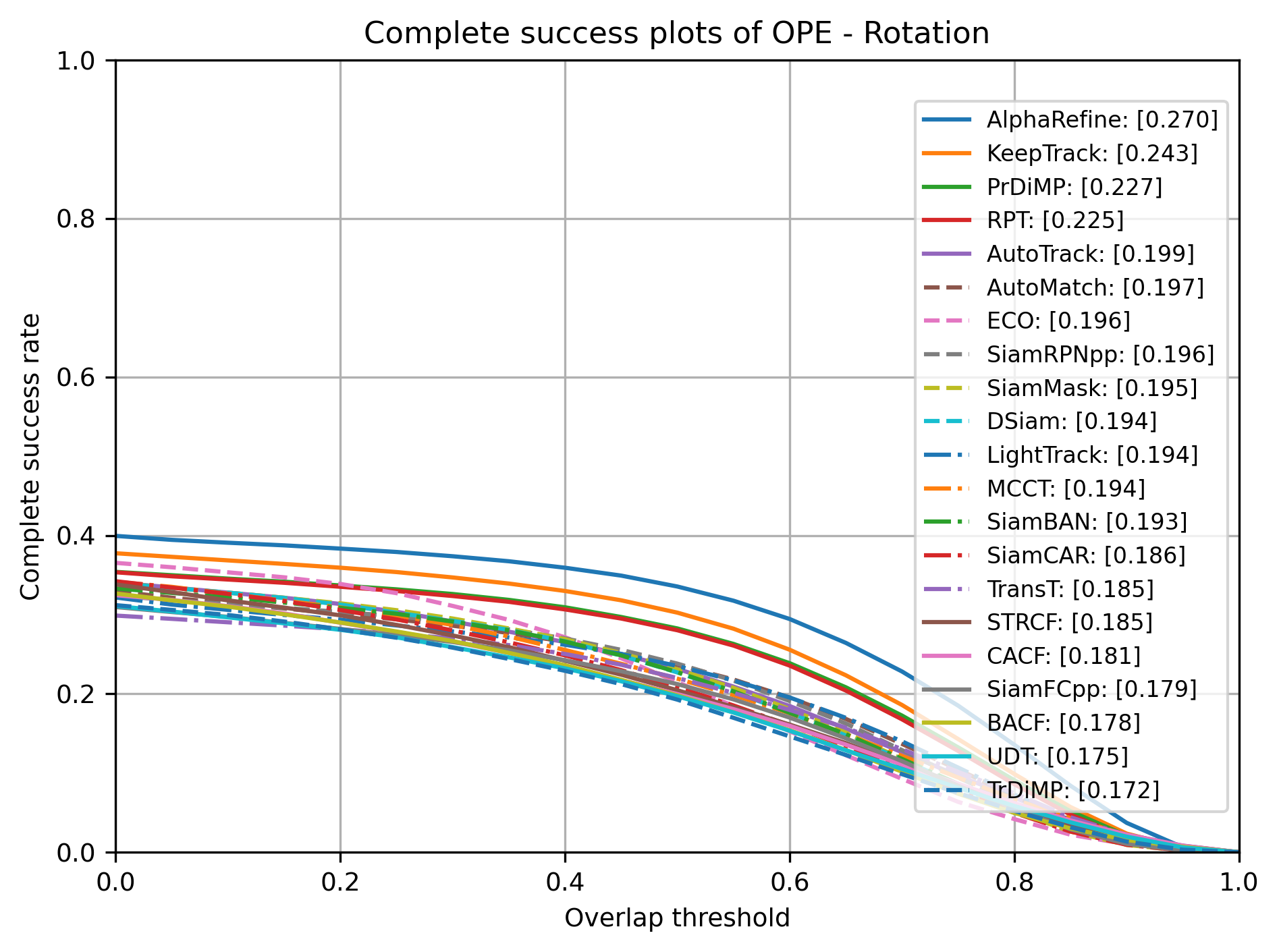}}\\

    \subfloat{\includegraphics[width =0.5\columnwidth]{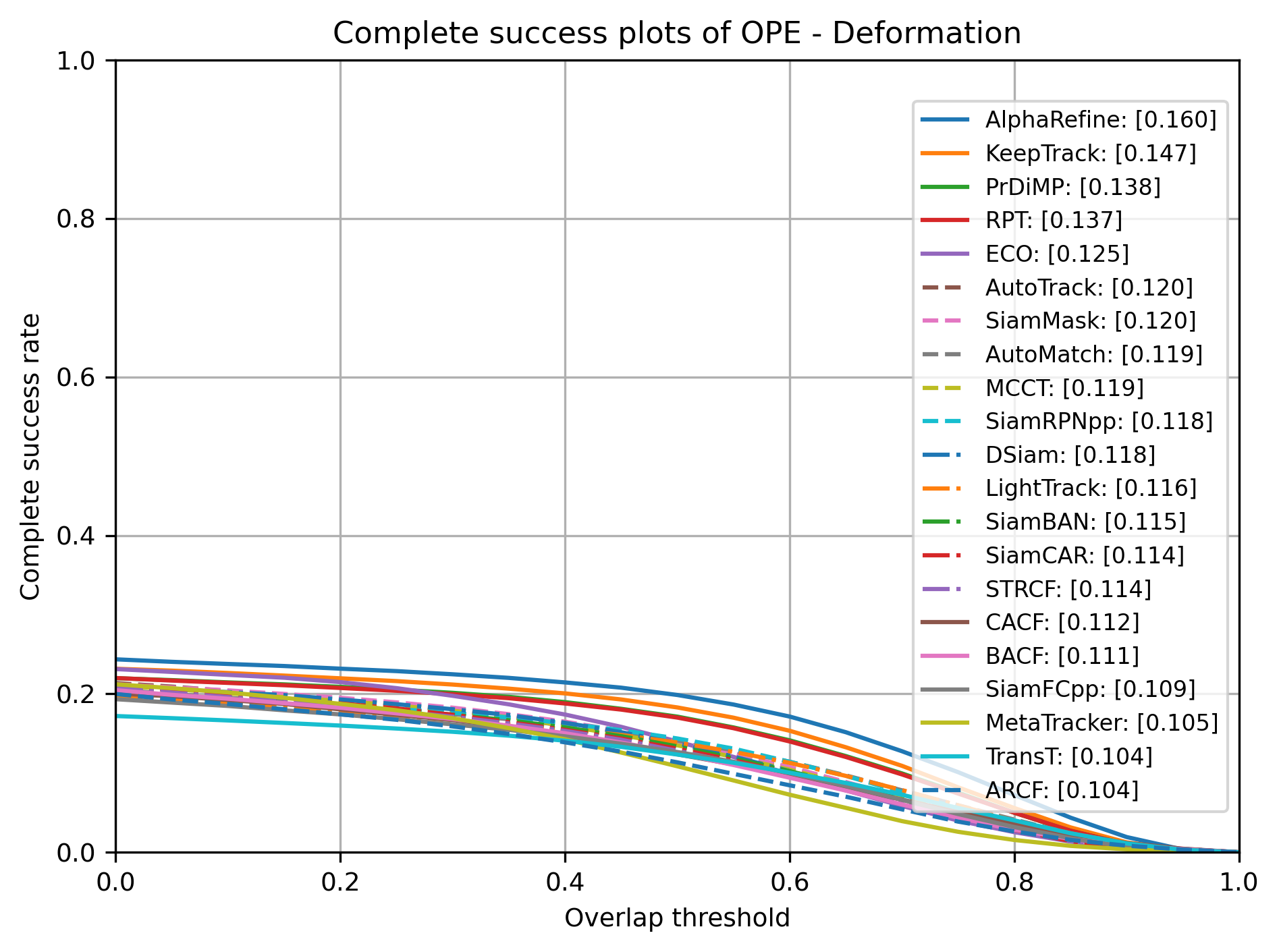}}
    ~
    \subfloat{\includegraphics[width =0.5\columnwidth]{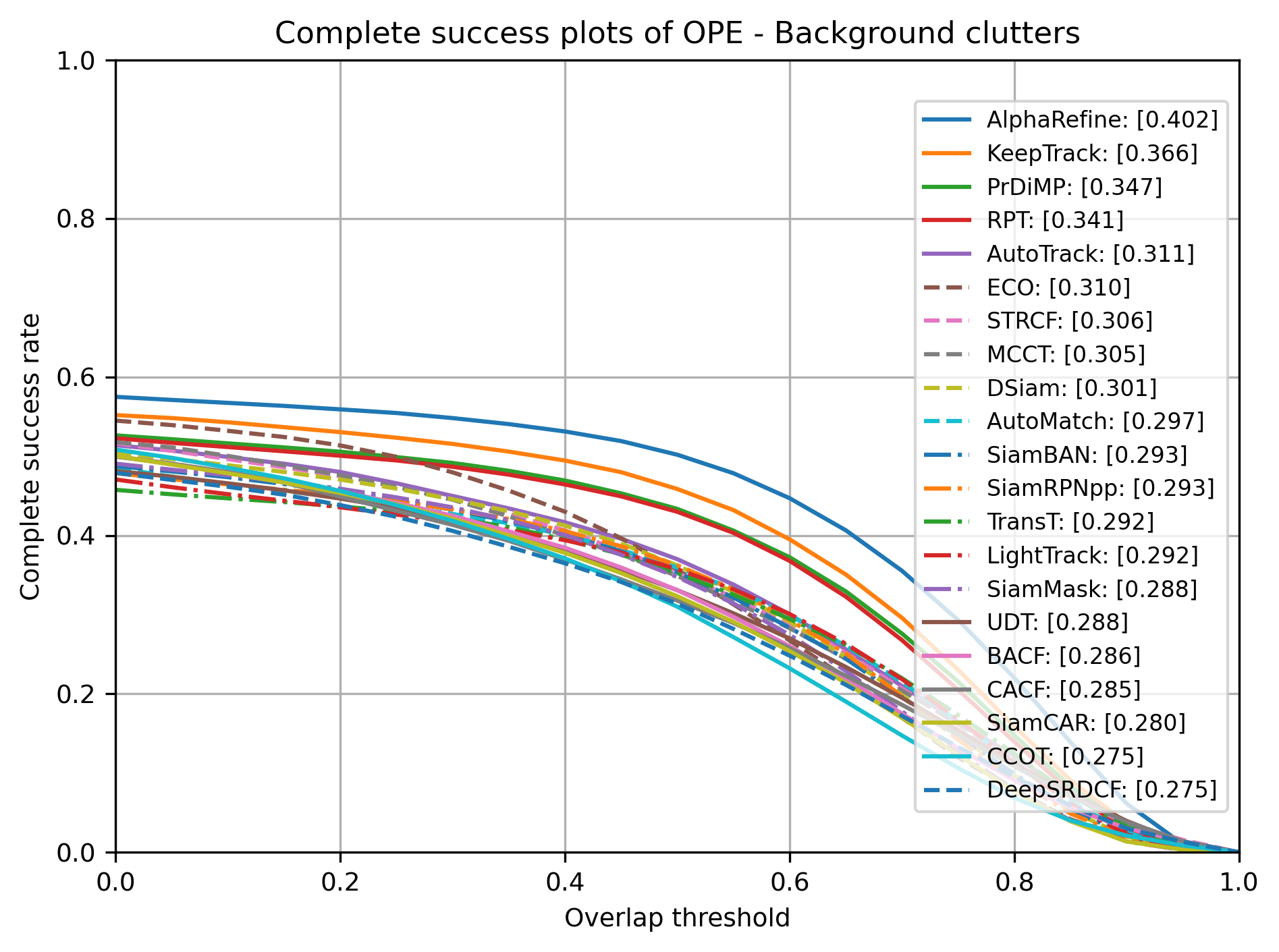}}
    ~
    \subfloat{\includegraphics[width =0.5\columnwidth]{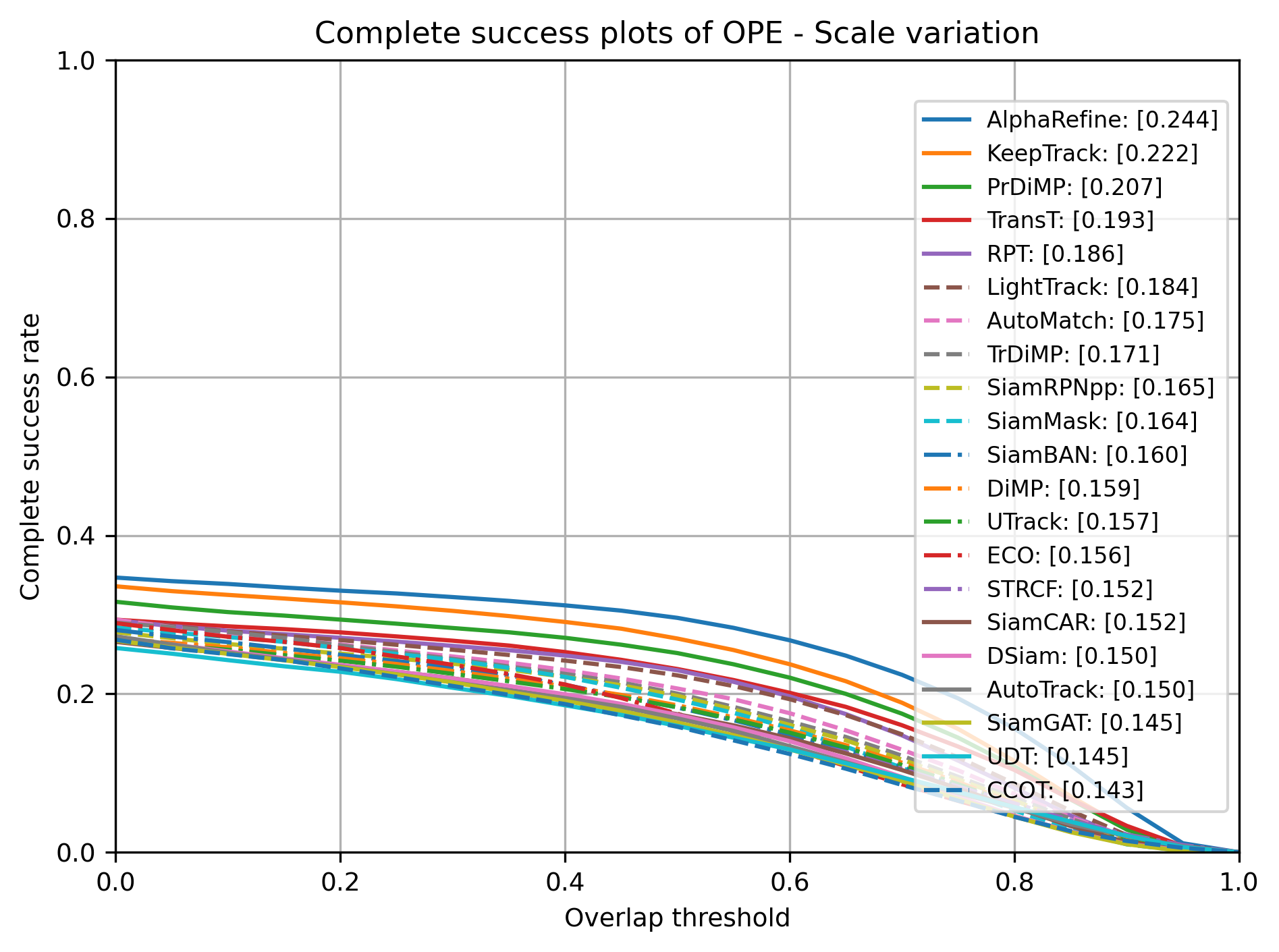}}
    ~
    \subfloat{\includegraphics[width =0.5\columnwidth]{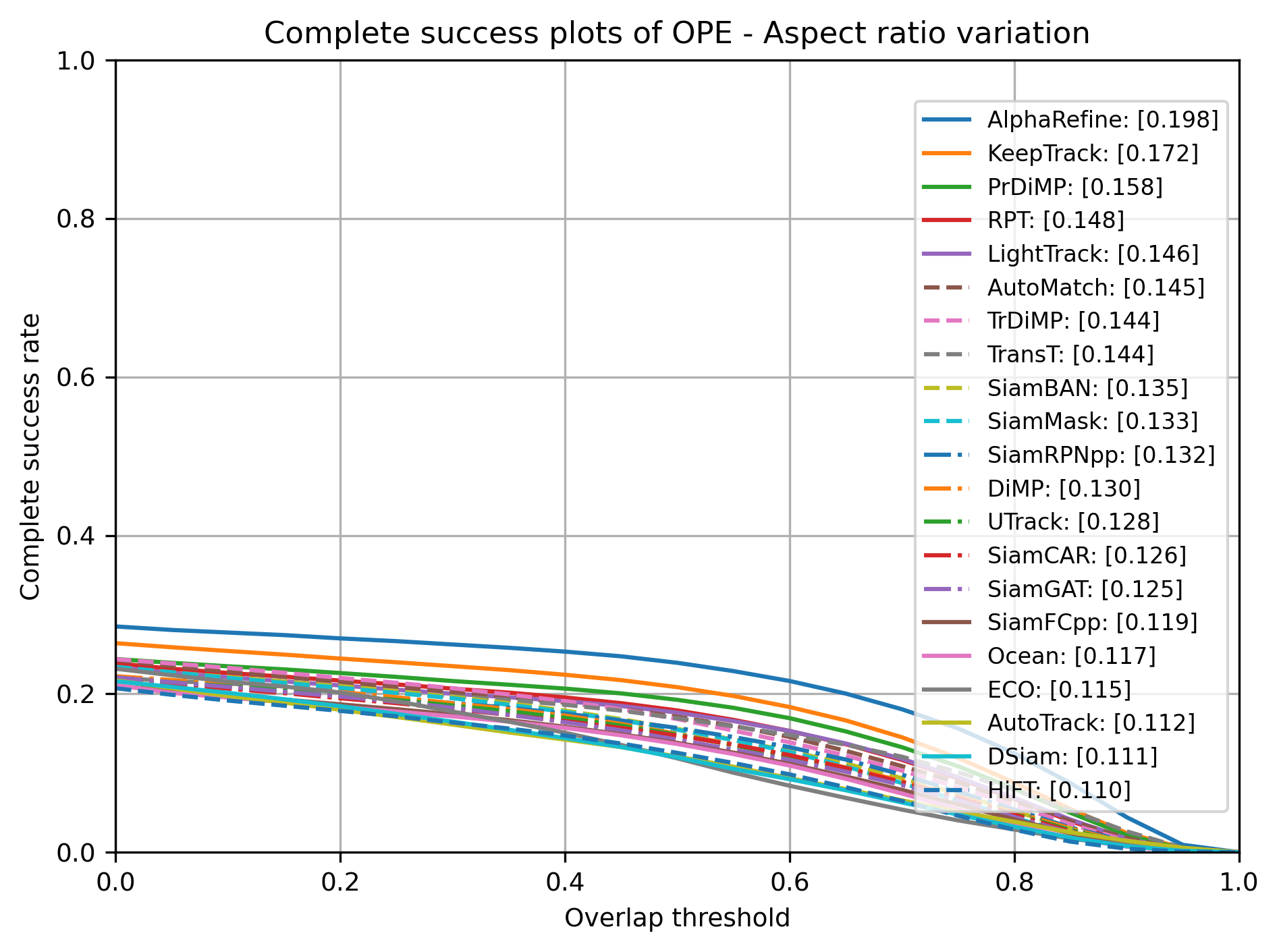}}\\
    
    \subfloat{\includegraphics[width =0.5\columnwidth]{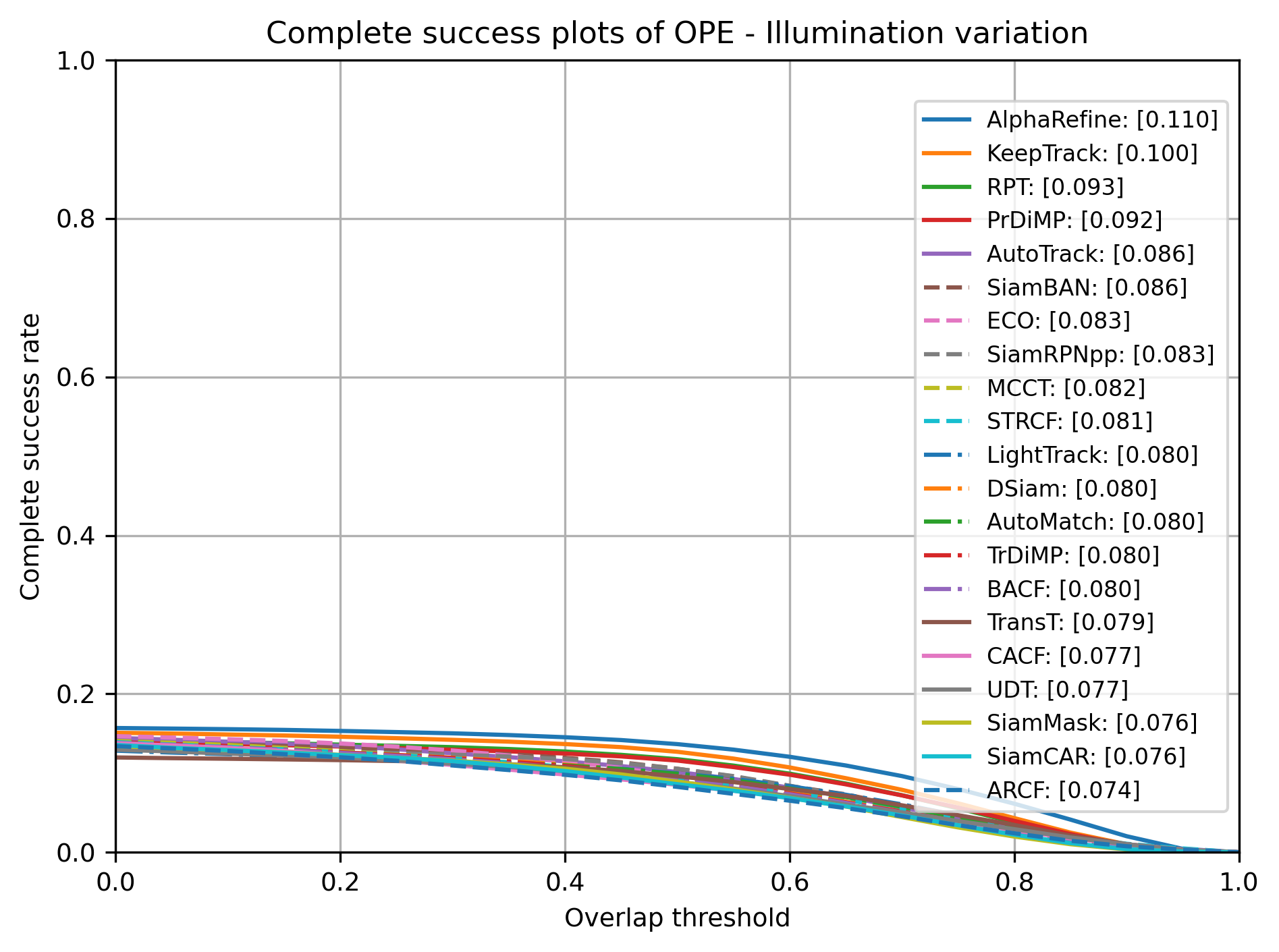}}
    ~
    \subfloat{\includegraphics[width =0.5\columnwidth]{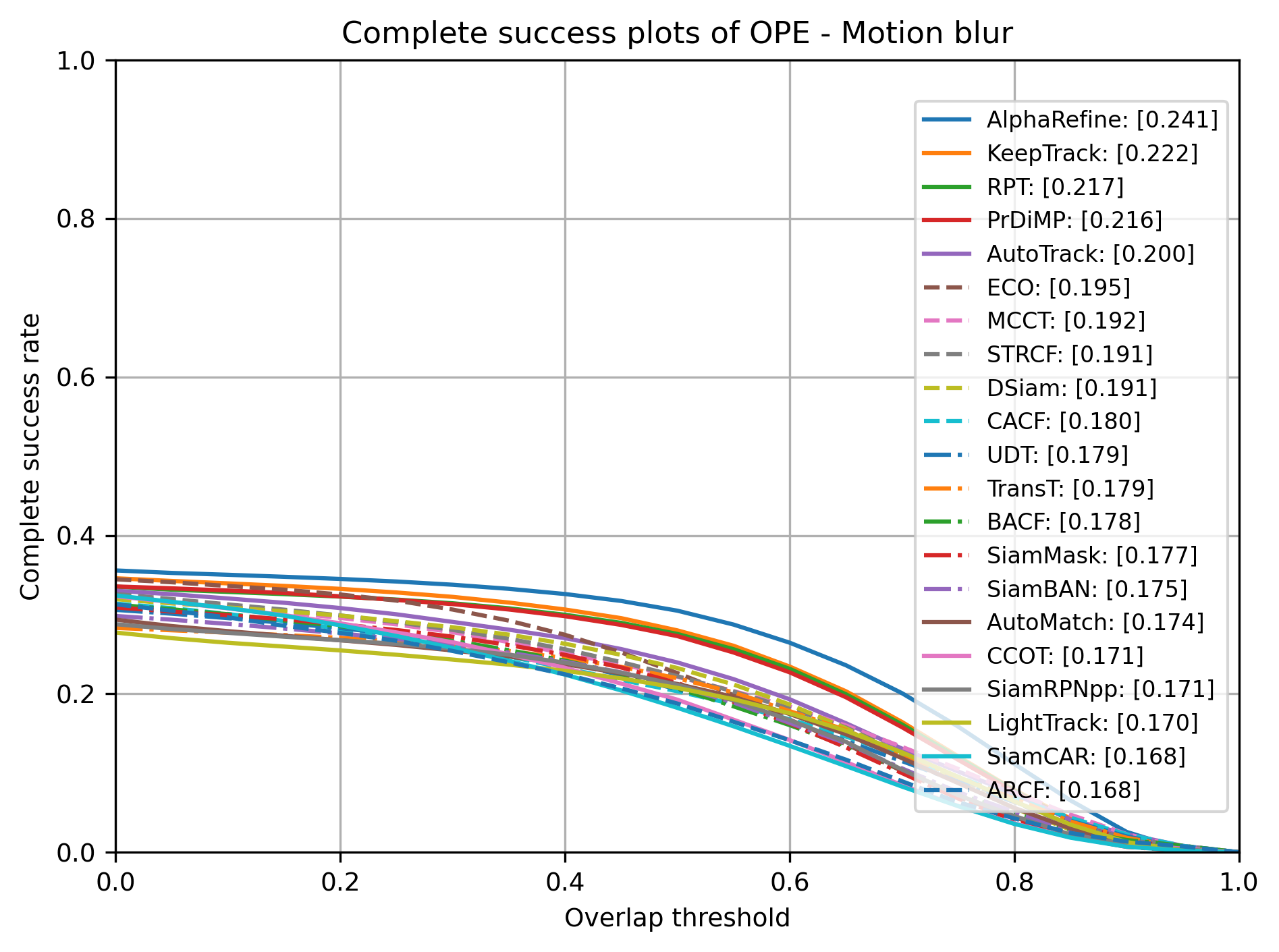}}
    ~
    \subfloat{\includegraphics[width =0.5\columnwidth]{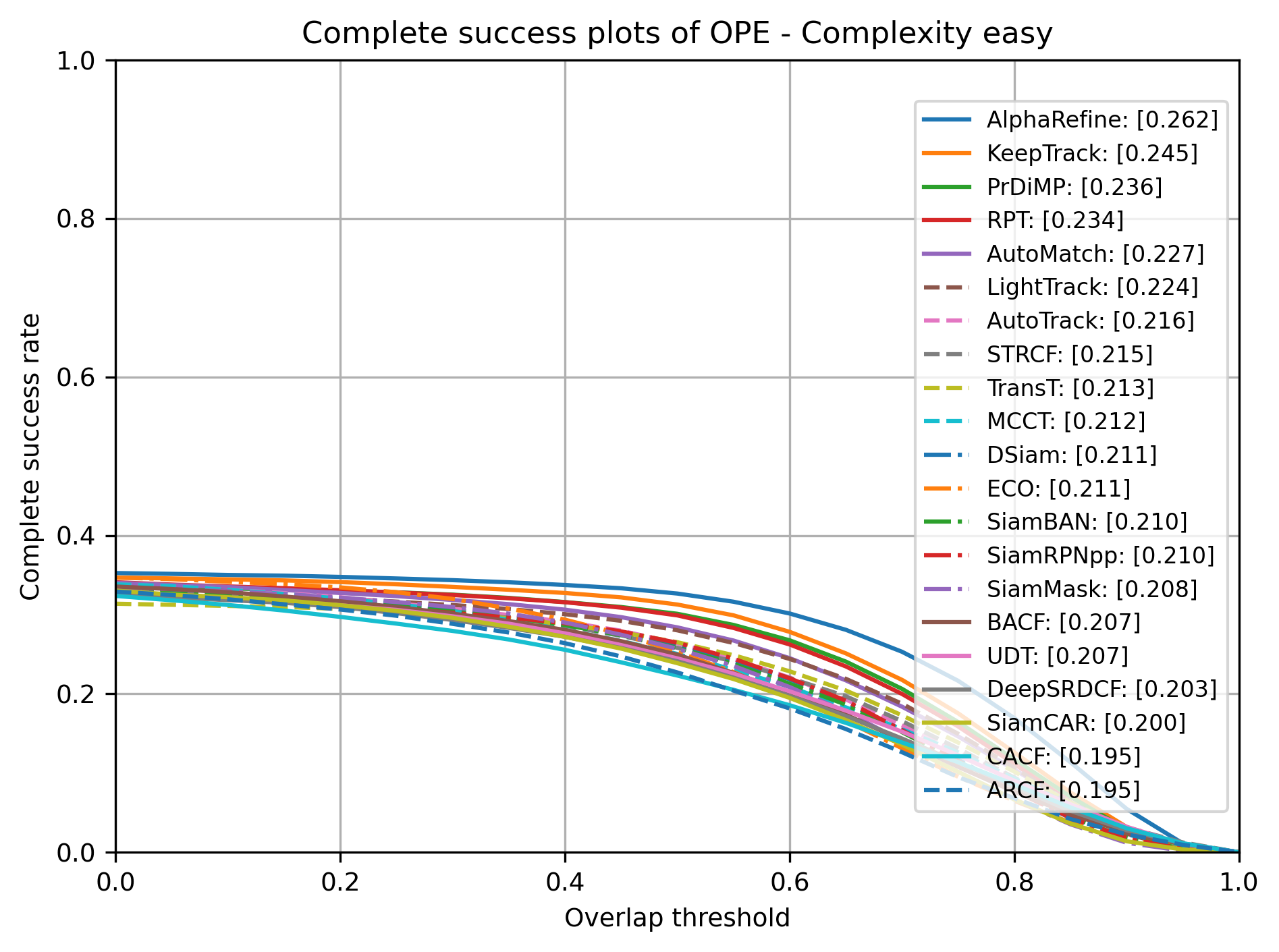}}
    ~
    \subfloat{\includegraphics[width =0.5\columnwidth]{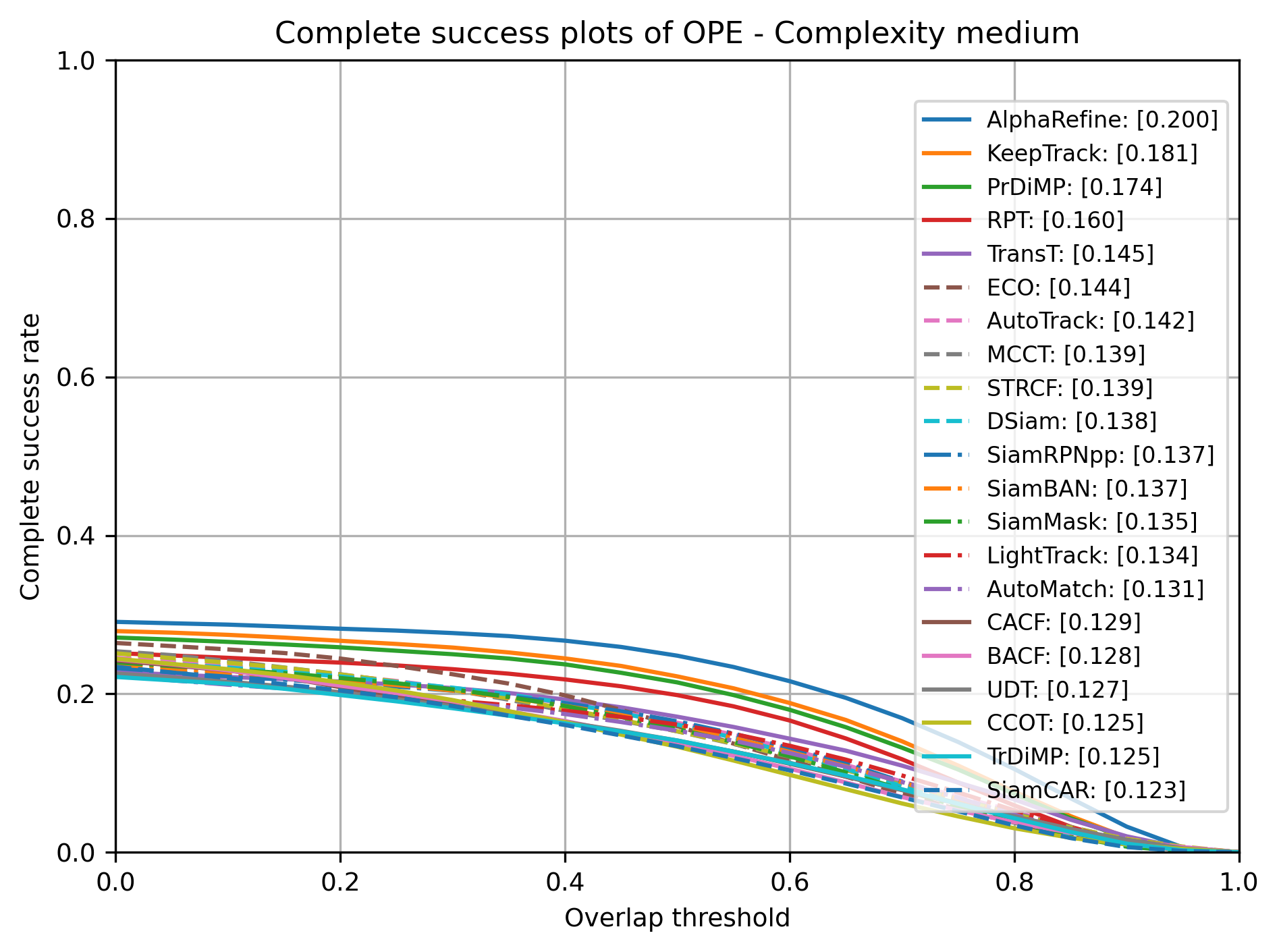}}
    \\
    \subfloat{\includegraphics[width =0.5\columnwidth]{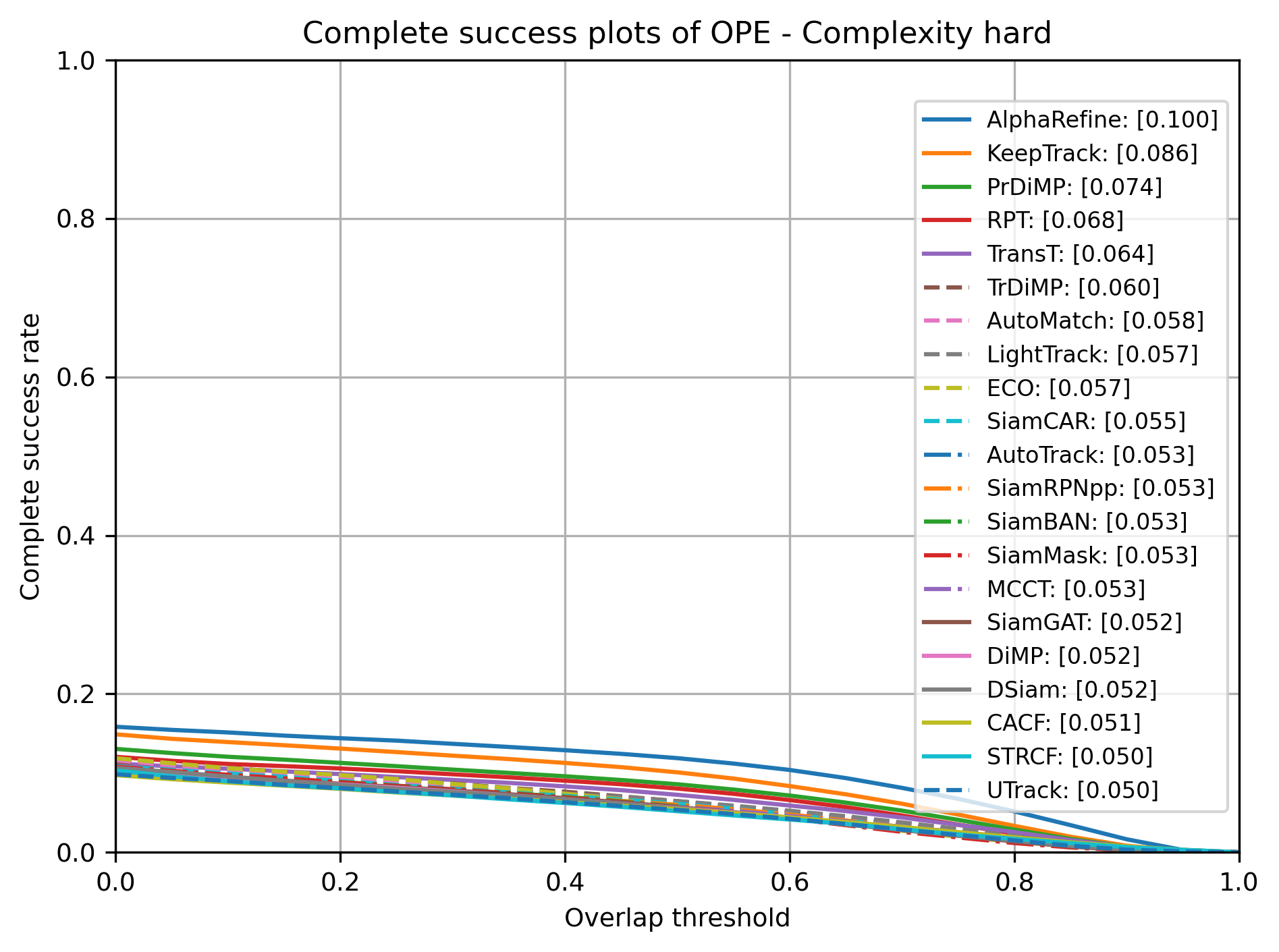}}
    ~
    \subfloat{\includegraphics[width =0.5\columnwidth]{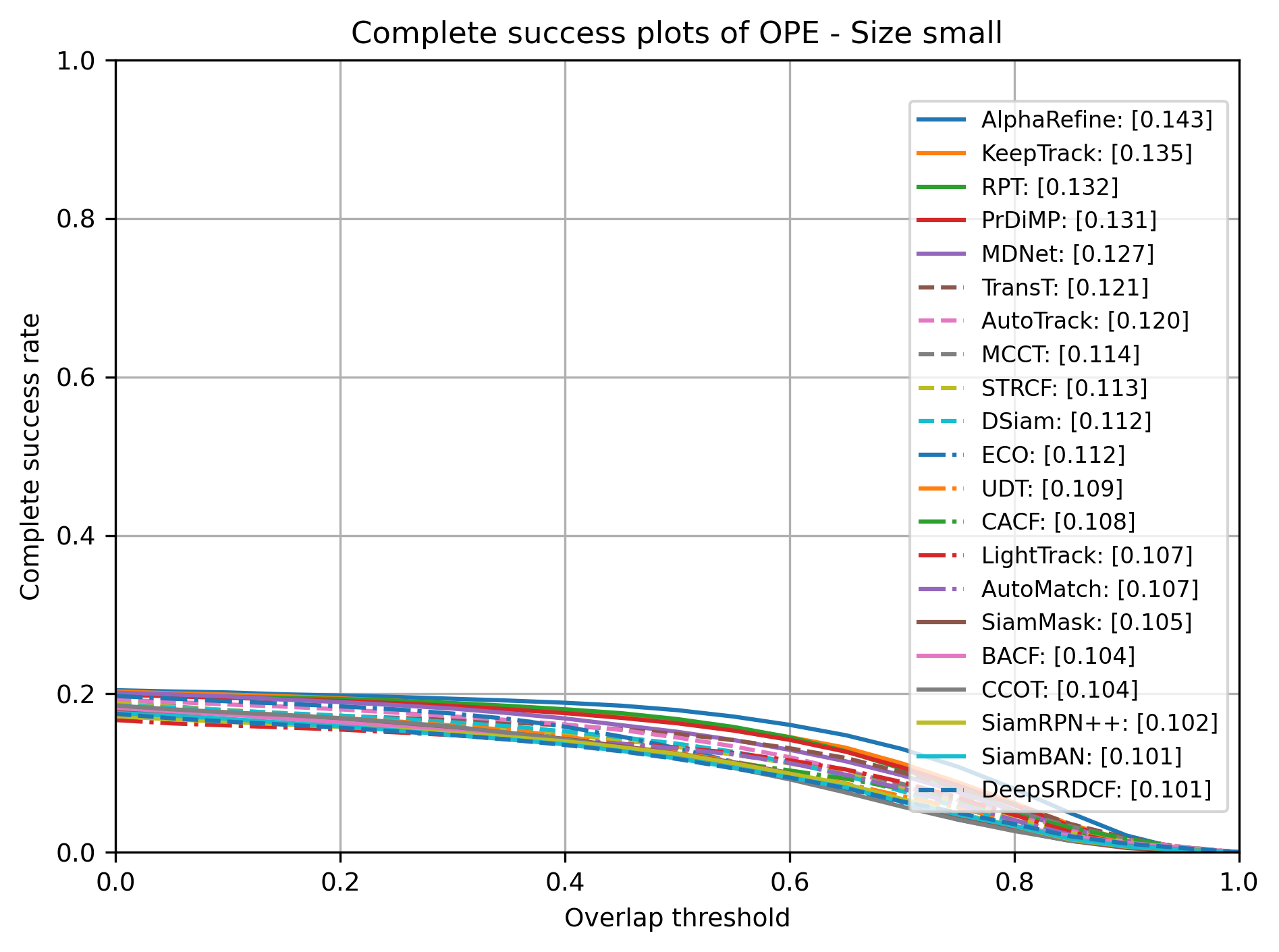}}
    ~
    \subfloat{\includegraphics[width =0.5\columnwidth]{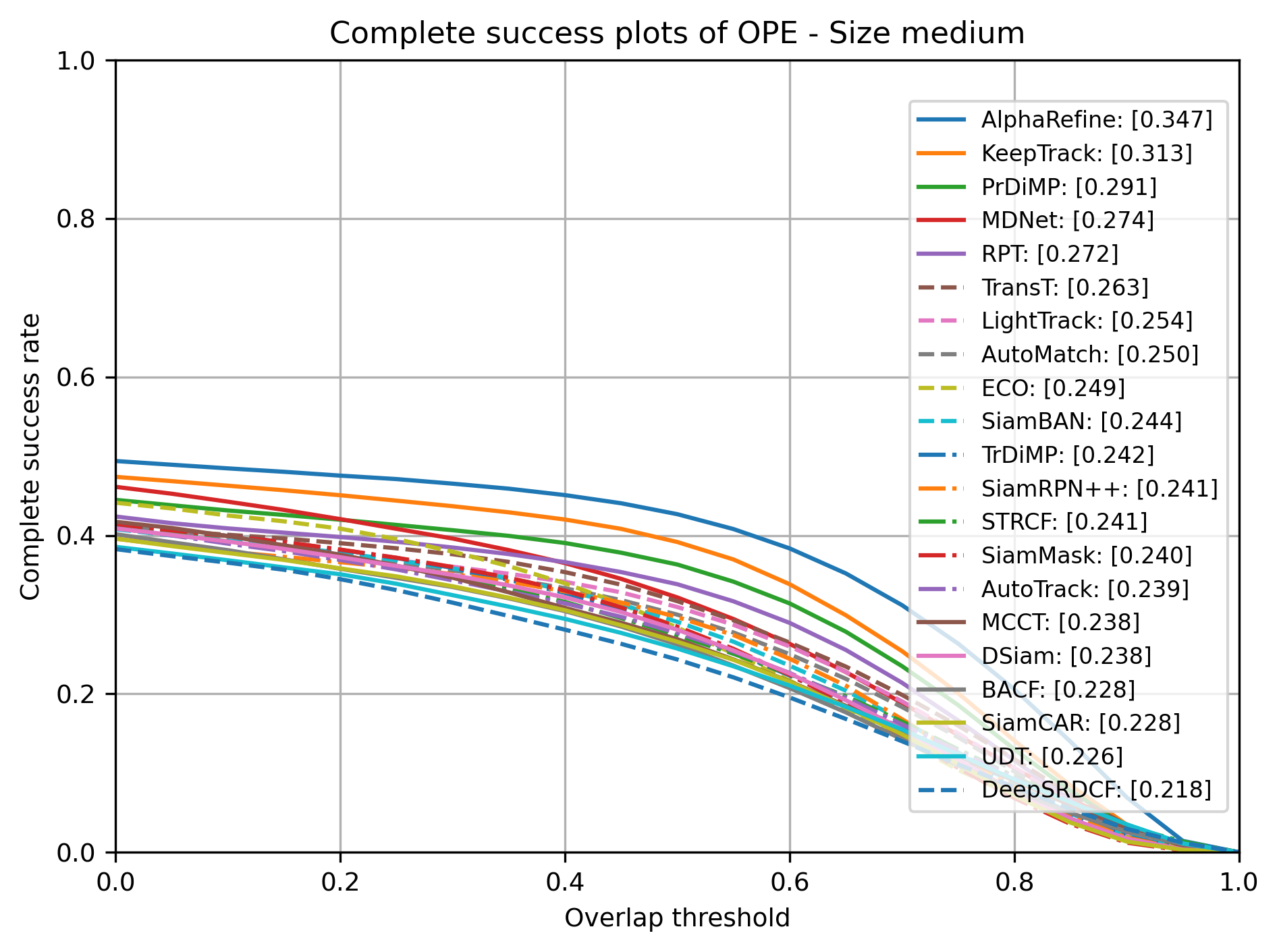}}
    ~
    \subfloat{\includegraphics[width =0.5\columnwidth]{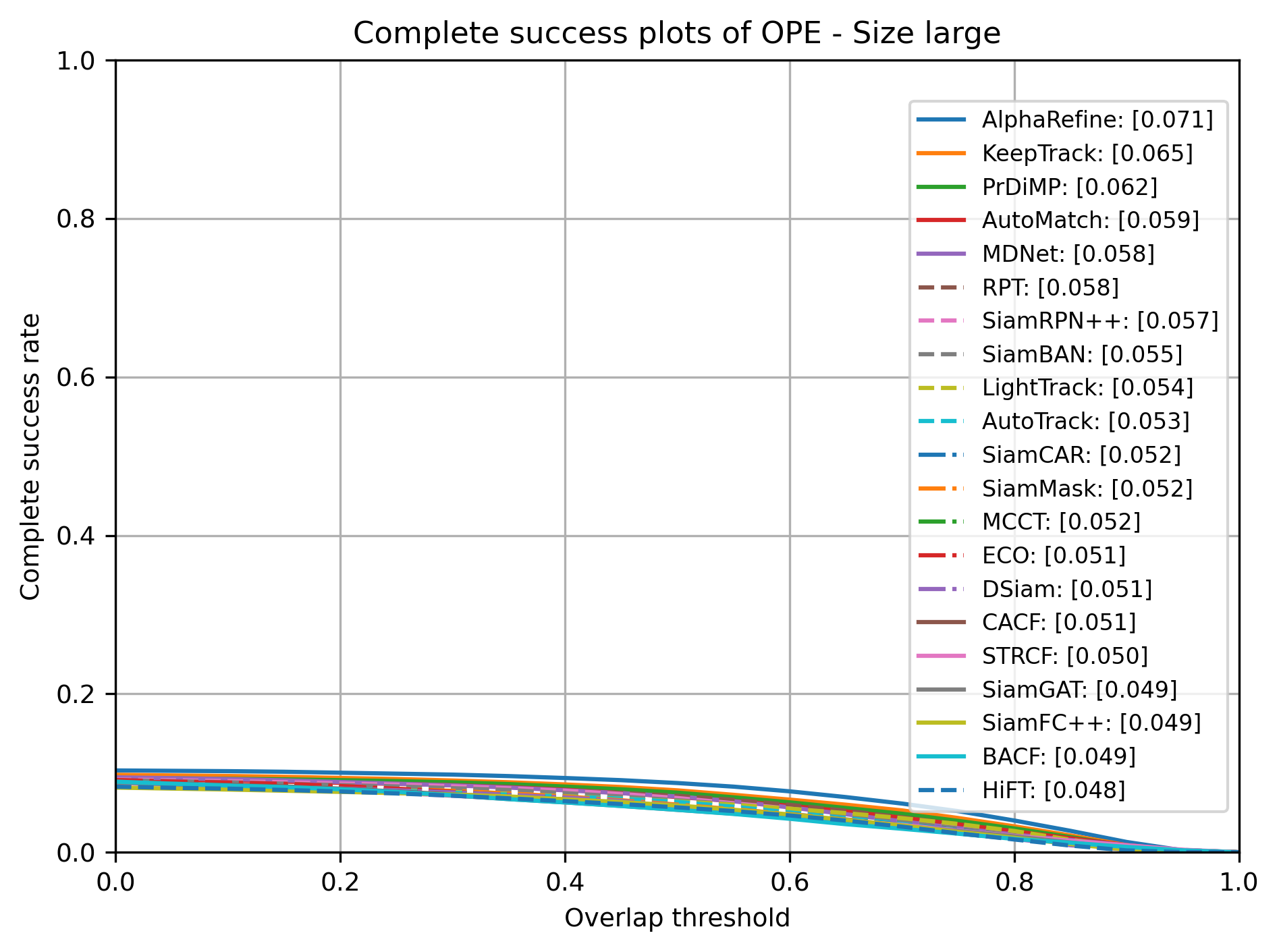}}\\
    \subfloat{\includegraphics[width =0.5\columnwidth]{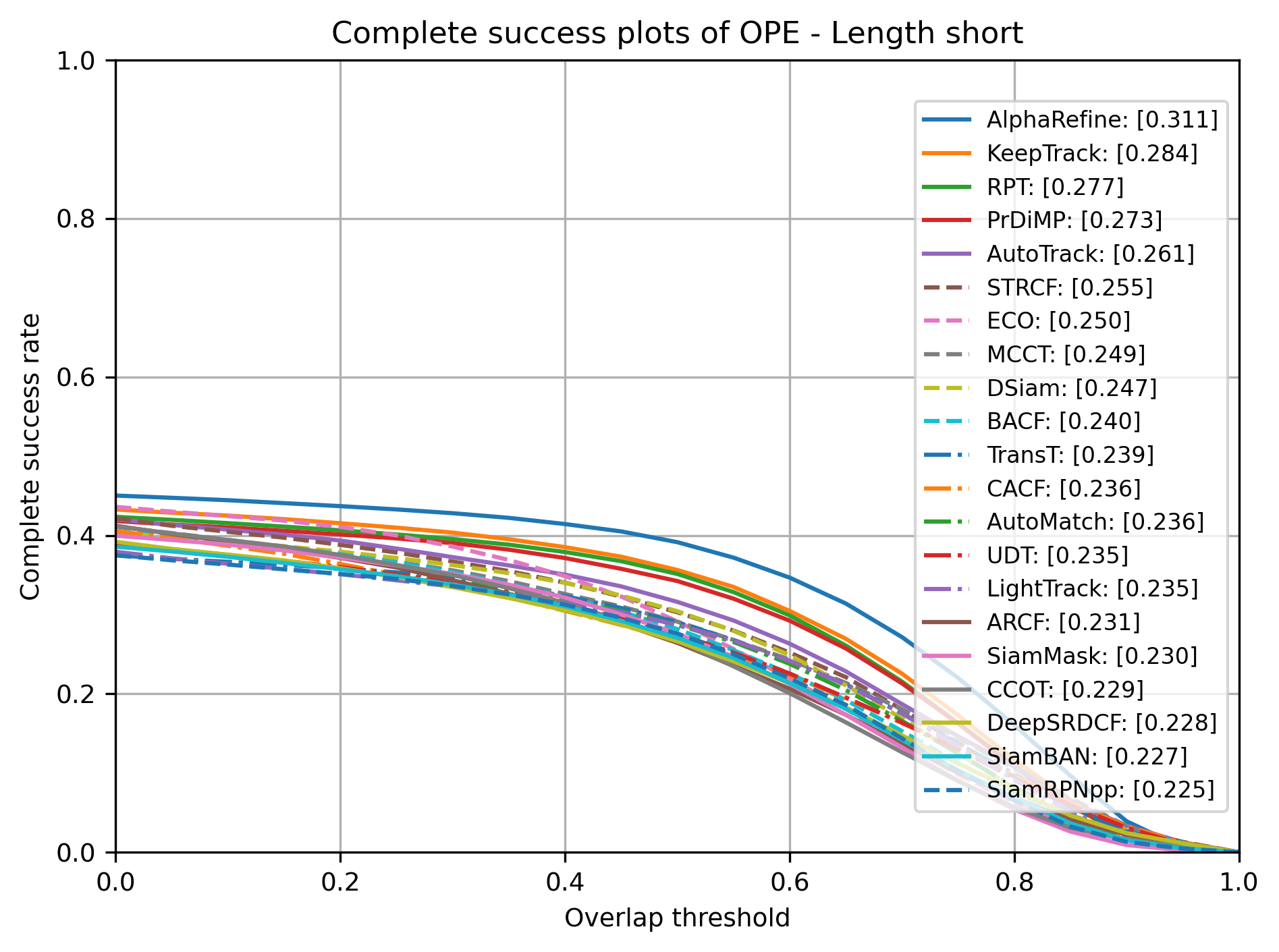}}
    ~
    \subfloat{\includegraphics[width =0.5\columnwidth]{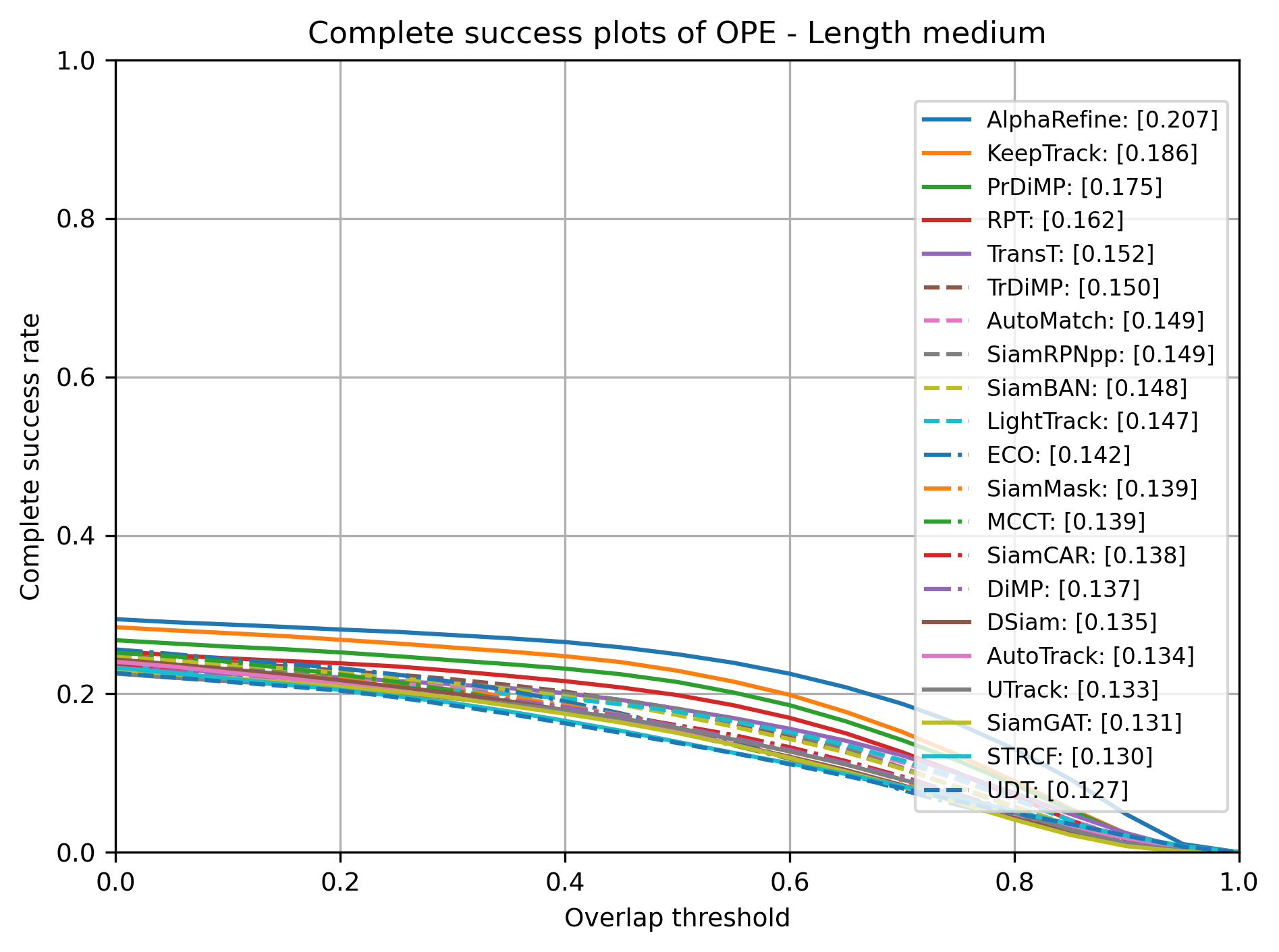}}
    ~
    \subfloat{\includegraphics[width =0.5\columnwidth]{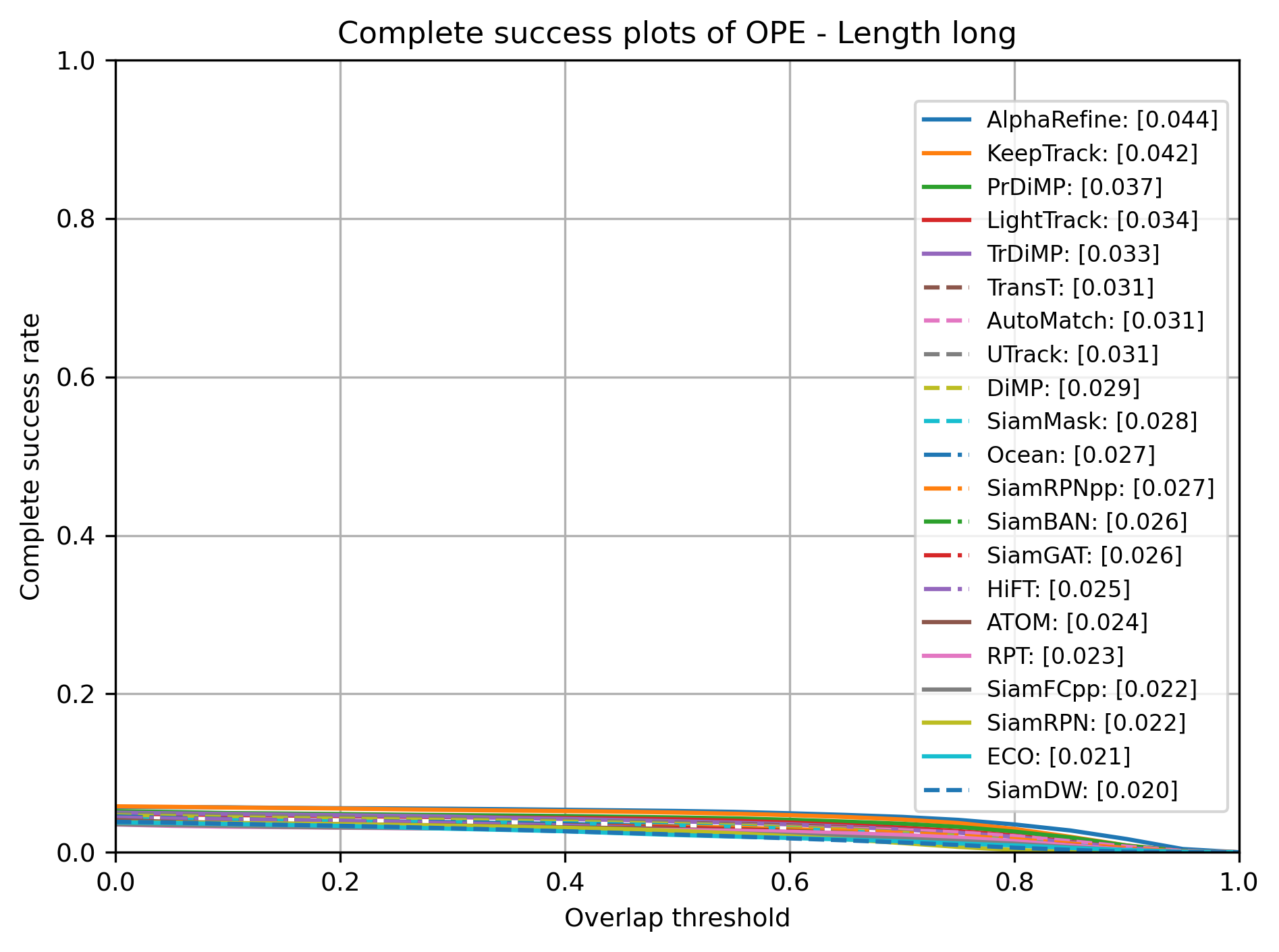}}\\
\caption{Performance of 43 baseline trackers in terms of 17 attributes (shown using a complete success plot). For clarity, only the top 21 trackers are shown. Best viewed by zooming in.}
\label{fig:Attribute_based_performance}
\end{figure*}

From Fig.~\ref{fig:Attribute_based_performance}, we find that AlphaRefine and KeepTrack rank in the top two for all 17 attributes, which is consistent with the results obtained on the entire test set. The performance of RPT improves on the LR subset, where its ranking increases by one place (from 4th to 3rd) compared to its ranking on the whole test set. In fact, the tracker can easily drift in LR videos due to its ineffective representations of small targets. We argue that the feature degradation caused by low resolution may be enhanced by a more refined representation with a set of learned representative points, leading to better performance. For videos with PO, FO, OV, FM, CM, and VC, the trackers are prone to losing the target because most of the existing trackers usually perform localization based on a small local region. To address these challenging factors, state-of-the-art deep trackers have introduced some practical solutions, such as extracting and maintaining as much detailed spatial information as possible (\eg AlphaRefine), keeping track of distractor objects to continue tracking the target of interest (\eg KeepTrack), predicting the conditional probability density of the target state given an input image (\eg PrDiMP), combining templates and searching region features using attention (\eg TransT). On the subsets with ROT, DEF, SV, and ARV, AlphaRefine, KeepTrack, and PrDiMP have significant advantages over STRCF, ECO, and AutoTrack. This result demonstrates that the target-specific features learned by the intersection-over-union network (IoU-Net) increase the accuracy of target estimation over that of classic multiscale search methods. Trackers are more likely to drift on subsets with BC, IV, and MB due to the presence of more minor discriminative representation features between the target and background distractors. A practical solution for alleviating this issue is to exploit both target and background appearance information to achieve enhanced discriminability (\eg PrDiMP). In addition, we observe that the brutal, small-sized, big-sized, and long videos are challenging for current deep trackers, causing all the baseline trackers to exhibit significant performance degradations.

\subsection{More Qualitative Results}
\label{sec:more_qualitative_results}
\begin{figure*}[t]   
    \centering\centerline{\includegraphics[width=1.0\linewidth]{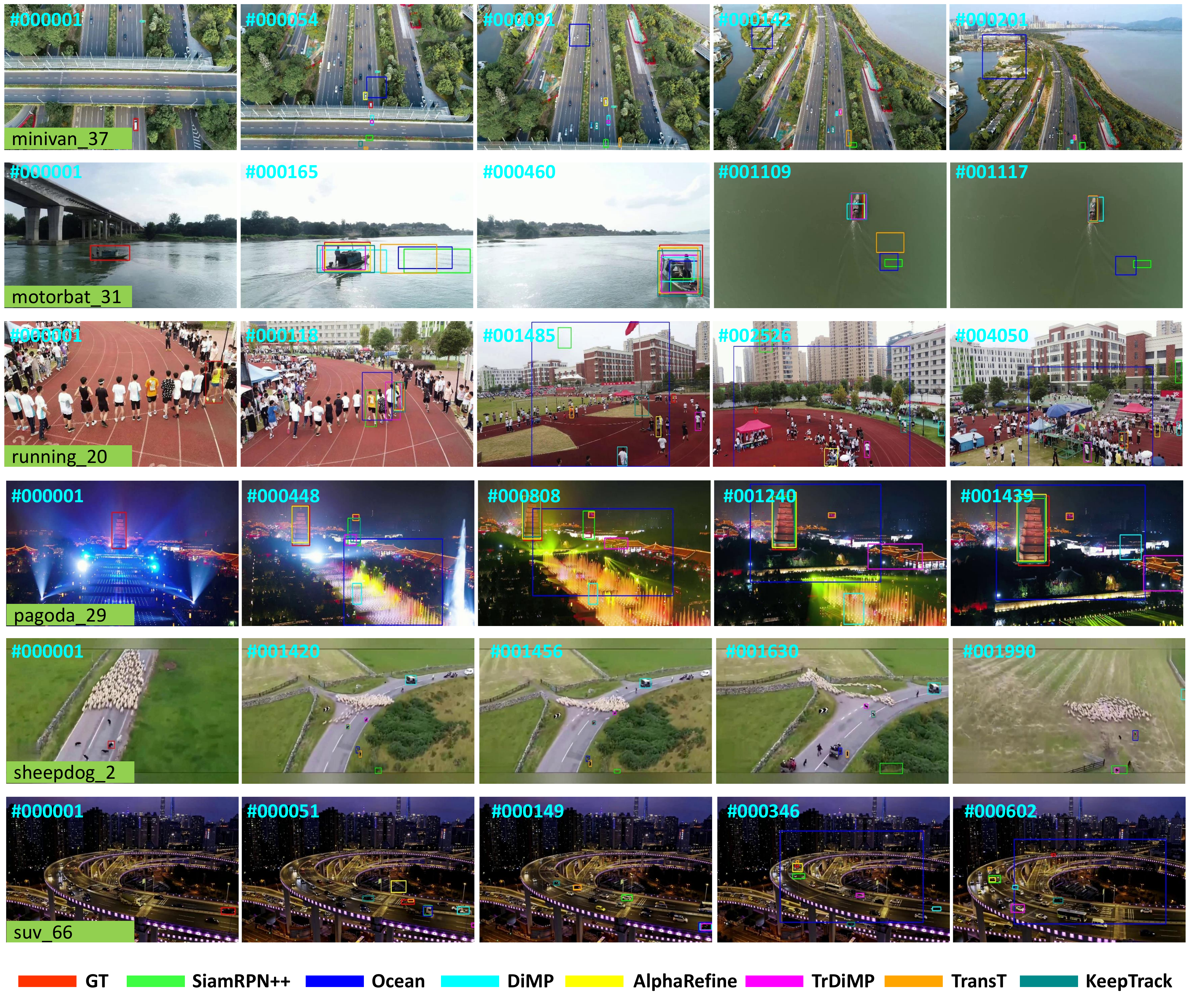}}
    \caption{Illustration of the qualitative evaluation results obtained in six hard scenario challenges: \textit{minivan$\_$37} with long-term occlusion, \textit{motorboat$\_$31} with dual-dynamic disturbances, \textit{running$\_$20} with high-speed motion, \textit{pagoda$\_$29} with target distortions, \textit{sheepdog$\_$2} with small targets and \textit{suv$\_$66} with low light. Best viewed by zooming in.}
    \label{fig:Visualization_results}
\end{figure*}

To qualitatively analyze different methods, we demonstrate the visual tracking results of seven state-of-the-art deep trackers (\ie SiamRPN++, Ocean, DiMP, AlphaRefine, TrDiMP, TransT, and KeepTrack) in six complex scenario challenges (\ie long-term occlusion, target distortions, dual-dynamic disturbances, small targets, high-speed motion, and low light). From Fig.~\ref{fig:Visualization_results}, we observe that none of these deep trackers can track the target in each frame in all scenarios, suggesting that tracking in real-world application scenarios is challenging and has still not been completely solved.

\end{document}